%% file: main.tex
\documentclass[11pt]{article}

\PassOptionsToPackage{hyperfootnotes=false}{hyperref}
\usepackage[preprint]{acl}

\usepackage{times}
\usepackage{latexsym}
\usepackage[T1]{fontenc}
\usepackage[utf8]{inputenc}
\usepackage{microtype}
\IfFileExists{inconsolata.sty}{\usepackage{inconsolata}}{}
\usepackage{graphicx}

\usepackage{amsmath}
\usepackage{amssymb}
\usepackage{booktabs}
\usepackage{placeins}
\usepackage{tikz}
\usetikzlibrary{arrows.meta,positioning,calc}
\usepackage{xcolor}
\input{figures/results_plot_data}
\input{figures/strength_control_data}
\input{figures/strength_comparison_plot_data}
\input{figures/overview_data}

\newcommand{\codeset}{\mathcal{Z}}        

\ifdefined\showrevisions
  \newcommand{\frank}[1]{{\color{red}#1}}
  \newcommand{\commentedit}[1]{{\color{red}#1}}
  
\else
  \newcommand{\frank}[1]{#1}
  \newcommand{\commentedit}[1]{#1}
  
\fi
\newif\ifarxivrevisions
\arxivrevisionsfalse
\newcommand{\arxivrev}[1]{\ifarxivrevisions
  {\color{red}\hypersetup{citecolor=red,linkcolor=red,urlcolor=red}#1}%
  \else#1\fi}
\title{Strength Controls Change How Interventions\\
Compare Across Sparse Autoencoder Codes}

\input{authors}
\hypersetup{
  pdftitle={Strength Controls Change How Interventions Compare Across Sparse Autoencoder Codes},
  pdfauthor={Jinchen Zhang, Zheyuan Liu, Yihan Zhu, Meng Jiang}
}

\begin{document}
\maketitle

\input{sections/00_abstract}

\input{figures/overview}
\input{figures/comparison_design}
\input{sections/01_introduction}
\input{sections/01_motivation}
\input{sections/02_related_work}
\input{sections/02_background}
\input{sections/03_comparison_protocol}
\input{sections/03_experiments}
\input{sections/03_audit_design}

\input{sections/04_lexical_calibration}
\input{sections/05_answer_excluded}
\input{figures/core_calibration}
\input{sections/04_candidate_stress}
\input{sections/06_global_transfer}
\input{sections/05_multibudget}
\input{sections/07_interpretation}
\input{sections/10_conclusion}

\phantomsection\label{main-text-end}

\input{sections/11_limitations}

\input{sections/12_ethics}

\bibliography{custom}

\clearpage
\appendix
\input{sections/b_experimental_details}

\input{sections/i_strength_controls}
\FloatBarrier
\input{sections/c_additional_results}
\FloatBarrier
\input{sections/g_two_setting_details}
\FloatBarrier
\input{sections/h_llama_details}
\FloatBarrier
\input{sections/d_registered_stops}
\FloatBarrier
\input{sections/f_related_tests}
\input{sections/e_artifact_guide}

\end{document}

%% file: figures/results_plot_data.tex
\newcommand{\BudgetNinePoints}{%
\budgetpoint{0.25}{0.074309467}{0.079821210}{339}{0}
\budgetpoint{0.5}{0.055536766}{0.059068638}{343}{0}
\budgetpoint{0.75}{0.042852493}{0.045805298}{343}{0}
\budgetpoint{1.0}{0.036864939}{0.039342957}{334}{0}
\budgetpoint{1.25}{0.037928589}{0.038097896}{303}{0}
\budgetpoint{1.5}{0.042288157}{0.042047135}{209}{0}
}
\newcommand{\BudgetNineCommon}{\draw[black!65,densely dashed,line width=.7pt] (0.000000,4.994636294) -- (1.000000,4.372059138) -- (2.000000,3.974158952) -- (3.000000,3.780288710) -- (4.000000,3.742936869) -- (5.000000,3.861414063);}
\newcommand{\BudgetTwoPoints}{%
\budgetpoint{0.25}{0.019268451}{0.022858266}{64}{0}
\budgetpoint{0.5}{0.016785661}{0.019075452}{64}{0}
\budgetpoint{0.75}{0.019015876}{0.022066113}{64}{0}
\budgetpoint{1.0}{0.019771460}{0.021687072}{64}{0}
\budgetpoint{1.25}{0.027497757}{0.024849292}{56}{1}
\budgetpoint{1.5}{0.028113721}{0.028113721}{35}{1}
}
\newcommand{\BudgetTwoCommon}{\draw[black!65,densely dashed,line width=.7pt] (0.000000,3.285747975) -- (1.000000,3.172263570) -- (2.000000,3.261983376) -- (3.000000,3.250612173) -- (4.000000,3.345478755) -- (5.000000,3.443411627);}

%% file: figures/strength_control_data.tex
\def\CandidatePrimaryReduction{0.724}
\def\CandidatePrecisionGoal{0.25}

\def\ConfirmationInputs{304}
\def\ConfirmationPairs{152}
\def\ConfirmationBlocks{113}
\def\ConfirmationEffectiveBlocks{98.3}
\def\ConfirmationSourcePairs{512}
\def\ConfirmationSourceBlocks{256}
\def\ConfirmationAlignedPairs{382}
\def\ConfirmationNativePairs{158}
\def\ConfirmationPrimary{0.160}

\def\ConfirmationQoneGap{0.021}

\def\ConfirmationEncoderTransfers{175}
\def\ConfirmationTotalLost{11}
\def\ConfirmationTotalGained{1}
\def\ConfirmationDraws{10,000}
\def\ConfirmationDeletionMin{0.152}
\def\ConfirmationDeletionMax{0.163}
\def\FollowupQtwoInputs{304}
\def\FollowupQtwoGap{0.029}

\def\FollowupQtwoLost{2}

\def\GenderCommonPosition{0.012}
\def\GenderCommonTotal{0.085}
\def\GenderCommonQone{0.047}
\def\StrengthCandidateNine{{504}{0.8140073494305686}{0.07555460267596775}{478}{0.7502463322803067}{0.028212980246444128}}
\def\StrengthCandidateTwo{{64}{0.5572201907634735}{0.013765215873718262}{64}{0.5572201907634735}{0.013765215873718262}}
\def\StrengthCandidateLlamaWide{{124}{1.5588496654264388}{0.27040390814504317}{92}{1.3192739849505217}{0.1298379431600156}}
\def\StrengthCandidateLlamaNarrow{{106}{1.4939820361587237}{0.1974245107398843}{75}{1.2336782582600911}{0.041013755798339845}}
\def\StrengthCandidateGender{{40}{0.8510470390319824}{0.3064012050628662}{16}{0.8160080909729004}{0.1612398624420166}}
\expandafter\def\csname UnmatchedQ1\endcsname{113}
\expandafter\def\csname UnmatchedQ2\endcsname{90}
\def\GlobalNineInputs{346}
\def\GlobalNineWorsened{216}
\def\GlobalTwoInputs{64}
\def\GlobalTwoWorsened{24}
\def\RangeSearches{203}
\def\RangeRescued{139}

\def\RangeRisingAtCap{155}
\def\RangeAlphaMin{2.3}
\def\RangeAlphaMax{3.0}

\def\RangeGenderQtwoBalancedBefore{0.401}
\def\RangeGenderQtwoBalancedAfter{0.361}
\def\StrengthRangeNine{{0.07555460267596775}{0.046513155339256165}}
\def\StrengthRangeLlamaWide{{0.27040390814504317}{0.19331230655793222}}
\def\StrengthRangeLlamaNarrow{{0.1974245107398843}{0.1440671794819382}}
\def\StrengthRangeGender{{0.3064012050628662}{0.24256186485290526}}
\def\NnlsRestMin{0.008}
\def\NnlsRestMax{0.023}

%% file: figures/strength_comparison_plot_data.tex
\def\ControlForestA{%
\controlArow{3.5}{9B SVA}{165}{0.040723653}{0.040723653}{0.040723653}{0}{1/2}{1/0}%
\controlArow{2.9}{2B SVA}{62}{0.024339224}{0.011473534}{0.042070648}{1}{0/0}{0/0}%
\controlArow{2.3}{Llama-128K}{188}{0.142250269}{0.117091971}{0.167725505}{1}{10/1}{0/1}%
\controlArow{1.7}{Llama-32K}{112}{0.114815786}{0.076562951}{0.159388481}{1}{3/1}{0/1}%
\controlArow{1.1}{2B gender}{42}{0.033302927}{0.006654789}{0.064566173}{1}{0/0}{0/0}%
\controlArow{0.1}{Llama-128K}{304}{0.160153051}{0.131009537}{0.191322576}{1}{11/1}{0/0}%
}
\def\ControlForestB{%
\controlBrow{3.5}{150}{-0.024065170}{-0.024065170}{-0.024065170}{0}%
\controlBrow{2.9}{62}{0.001271289}{-0.003937448}{0.006431824}{1}%
\controlBrow{2.3}{188}{-0.047637627}{-0.067189995}{-0.029715078}{1}%
\controlBrow{1.7}{109}{-0.029140936}{-0.046580078}{-0.011899102}{1}%
\controlBrow{1.1}{24}{-0.072920370}{-0.072920370}{-0.072920370}{0}%
\controlBrow{0.1}{304}{-0.046392243}{-0.065304912}{-0.028241655}{1}%
}
\def\BoundaryPositionLlama{{0.180890786}{0.020737735}{0.134498543}{304}}
\def\BoundaryPositionGender{{0.084977627}{0.046916437}{0.012057257}{24}}
\def\BoundaryCriterionLlama{{-0.184220472}{-0.222804575}{-0.148879031}{304}}
\def\BoundaryCriterionGender{{0.049919218}{0.008128761}{0.090880133}{40}}
\def\CandidateReductionNine{\candidatereduction{0.723859594}{0.324087319}{1.123631869}}
\def\CandidateReductionTwo{\candidatereduction{0.545924242}{0.363742516}{0.748452219}}
\def\CandidateReductionLlamaWide{\candidatereduction{1.227920294}{1.014873372}{1.425579150}}
\def\CandidateReductionLlamaNarrow{\candidatereduction{1.260216649}{0.962578494}{1.565467330}}
\def\CandidateReductionGender{\node[align=center,fill=white,inner sep=2pt] at ({\redzero+\redscale*.85},0) {\frank{Estimate and interval unavailable}\\Missing case};}

%% file: figures/overview_data.tex
\newcommand{\OverviewNativeMargin}{4.923}
\newcommand{\OverviewEncoderMargin}{-0.233}
\newcommand{\OverviewTotalMargin}{0.034}
\newcommand{\OverviewQoneMargin}{-0.211}

%% file: authors.tex
\author{
  Jinchen Zhang\thanks{Equal contribution.}\quad
  Zheyuan Liu\textsuperscript{*}\quad
  Yihan Zhu\quad
  Meng Jiang \\
  University of Notre Dame \\
  \texttt{\{jzhang47,zliu29\}@nd.edu}
}

%% file: sections/00_abstract.tex
\begin{abstract}
Changing sparse autoencoder (SAE) codes can alter both intervention direction and magnitude, so comparing intervention effects requires a strength control specifying what is held equal.  Norm matching and output
calibration provide different controls, but whether \commentedit{interventions yield different effects across codes after strength matching} remains unclear.
We compare \commentedit{intervention effects under these controls}
for fixed feature groups \commentedit{in Gemma and Llama}
and observe: (1) \commentedit{\textbf{Whole-patch scaling.}}
With one coefficient
per input, full-output Kullback--Leibler (KL) calibration leaves smaller mean
effect gaps from \commentedit{the encoder-based reference intervention} than total-norm matching for nonnegative
least-squares (NNLS) refits. \commentedit{On new Llama inputs,} total-norm matching
loses \ConfirmationTotalLost{} \commentedit{reference} transfers, whereas full-output calibration
preserves every \commentedit{transfer} decision.
(2) \commentedit{\textbf{Positionwise scaling.}} \frank{\commentedit{On a small matched gender-agreement subset,}
positionwise norm matching leaves smaller observed mean gaps than
full-output calibration. (3) \commentedit{\textbf{Answer exclusion.}} Compared with unscaled NNLS, answer-excluded calibration reduces gaps in the measured subject--verb \commentedit{agreement} comparisons but increases them for gender refits.}
\commentedit{Full-output KL calibration} also reduces larger gaps for \commentedit{norm-matched} constructed candidates, although matching coverage
is incomplete in several settings. \commentedit{Since calibration uses the evaluated input's outputs, it serves as a diagnostic of
residual disagreement.}
\end{abstract}

%% file: figures/overview.tex
\begin{figure}[!t]
\centering
\includegraphics[width=7.7cm]{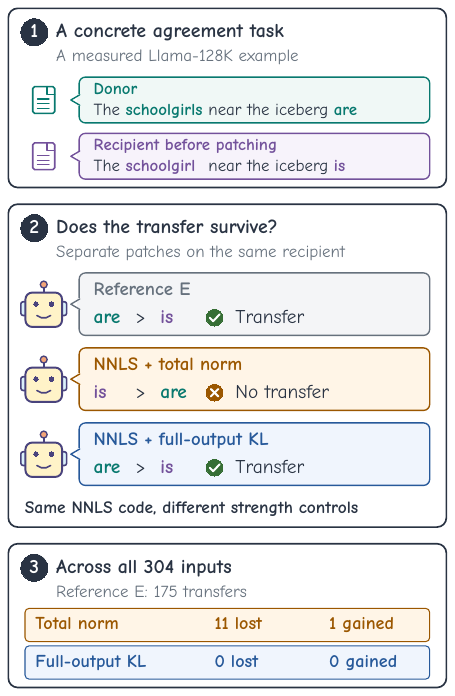}
\caption{\frank{\textbf{Strength control can change a transfer decision.}
A Llama-128K example selected post hoc.
(1) Prompts with task answers appended.
(2) \commentedit{$E$ is the reference intervention built from the released encoder's codes.}
Separate patches on the same native recipient. Rankings ($>$) compare only
the two answer logits. Ticks indicate donor-answer transfer, crosses its absence.
(3) Transfers lost or gained relative to \commentedit{$E$} across all
\ConfirmationInputs{} inputs. \commentedit{Agreement with $E$} is not a faithfulness standard.}}
\label{fig:overview}
\end{figure}

%% file: figures/comparison_design.tex
\begin{figure*}[!t]
\centering
\includegraphics[width=\textwidth]{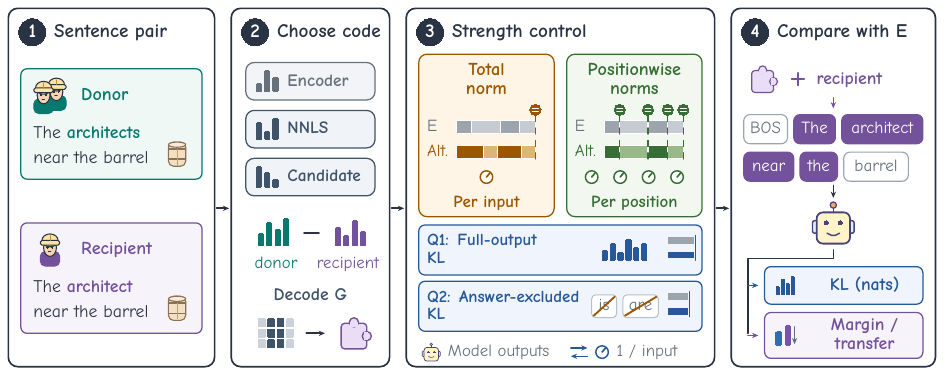}
\caption{\textbf{Four-step comparison \commentedit{with a fixed model, dictionary,
feature group, and sentence pair.}}
\frank{\commentedit{The reference intervention $E$, built from the released encoder's codes,}
supplies the reference targets (gray) for rescaling the
original alternative patch (color).
Norm segments show squared patch norms, matched in sum or per position.
KL controls use trial model outputs to choose one scaling coefficient (knob) per input
for matching scalar KL.
Q2 excludes both answer tokens (crossed) and renormalizes before matching conditional KL.
Step~4 keeps beginning-of-sequence (BOS) and last prompt tokens, comparing
native-to-patched KL (nats), donor-oriented margin change (logits) and
transfer with $E$.
Matching can fail.}}
\label{fig:comparison_design}
\end{figure*}

%% file: sections/01_introduction.tex
\section{Introduction}
\label{sec:intro}

Sparse autoencoders (SAEs) represent model activations through feature
directions and sparse coefficients
\citep{bricken2023monosemanticity,huben2024sparse}.
\arxivrev{The decoder supplies a dictionary of feature directions, and
the encoder assigns coefficients, collectively called a \emph{code},
that specify their contributions to an activation's reconstruction.}
Intervening on selected
features lets researchers test their role in model behavior
\citep{marks2025sparse}.

\arxivrev{One way to test these roles is activation patching, which
transfers selected internal activation values
between runs of the same model and measures the resulting output change
\citep{heimersheim2024patching}.
The input supplying these values is the \emph{donor}.
The \emph{recipient} is the input whose activations are modified
and whose output is evaluated
\citep[][Appendix~D.3]{baumgartner2025cognitive}.}

\arxivrev{We use sentence pairs with contrasting grammatical answers to
test whether patching shifts the recipient toward the donor's answer.
For a fixed group of SAE features, we decode the
donor-minus-recipient coefficient difference into a \emph{patch}.
The unscaled intervention adds this patch to the recipient's activations.
We quantify the resulting shift by the change in the logit margin between
the two task answers.
A \emph{transfer} occurs when the donor answer outranks the recipient answer
(Figure~\ref{fig:overview}).}

\arxivrev{The coefficients used to construct this patch can differ across
sparse inference methods, even when the dictionary and activations are
fixed \citep{oneill2025compute,carbo2026which}.}
\arxivrev{Changing the code can alter both the direction and magnitude
of the patch, so a larger answer effect may reflect a larger perturbation.
Comparing interventions therefore requires a \emph{strength control}
specifying what is held equal across patches.}

\arxivrev{We use the intervention built from the released encoder's codes
as the reference $E$ for both strength matching and effect comparison.}
\arxivrev{We rescale each alternative patch to match $E$ in either
patch norms or Kullback--Leibler (KL) divergence from the unpatched to the
patched next-token distribution. We compare total-norm and positionwise
norm matching with full-output KL calibration and an answer-excluded
variant that omits the two task-answer tokens.
Matching norms or KL need not equate answer effects.
Do these controls lead to the same conclusions about whether alternative
codes preserve the reference intervention's effects?}

\frank{We test subject--verb and reflexive gender agreement in five
selected Gemma and Llama model--dictionary--group configurations
(Table~\ref{tab:settings}). Alternatives to \commentedit{the encoder codes} are ordinary
nonnegative least-squares (NNLS) refits and constructed candidates
constrained by reconstruction and positionwise patch norms.}
\arxivrev{We measure an alternative's \emph{effect gap} as the absolute
difference between its answer-margin effect and that of $E$, and also
compare their transfer decisions.} \arxivrev{For NNLS refits, we find:}
\begin{enumerate}
\item \textbf{Controls differ with one scaling coefficient per input.}
\frank{Full-output KL leaves smaller mean NNLS gaps than total-norm
matching in all five configurations, confirmed prospectively on new
Llama material.}
\item \textbf{Positionwise scaling can improve agreement.}
\frank{On each comparison's common inputs, it lowers observed mean gaps relative to total-norm
matching in four configurations and is descriptively closest to \commentedit{the reference intervention $E$}
on 24 gender inputs, without a paired interval.}
\item \textbf{The output criterion matters.}
\frank{\commentedit{Compared with unscaled NNLS, answer-excluded calibration reduces mean effect gaps in the measured
subject--verb comparisons but increases them in the gender comparison despite successful KL matching.}}
\end{enumerate}

\arxivrev{For constructed candidates, full-output KL also reduces mean
effect gaps on matched inputs relative to the unscaled interventions,
although matching coverage is incomplete in several settings
(Section~\ref{sec:candidate}). For both code families,
per-input output calibration uses the evaluated input's own outputs to
select scaling coefficients, so we interpret it as a diagnostic of
residual disagreement.}

%% file: sections/01_motivation.tex
\section{\frank{Motivation}}
\label{sec:motivation}

\frank{\arxivrev{Figure~\ref{fig:overview} illustrates how strength control
changes a transfer decision for a Llama-128K input selected post hoc.}
\arxivrev{The reference intervention $E$ reverses the singular recipient's
initial preference for \emph{is}
over \emph{are}.}
For the same NNLS code, total-norm matching loses this transfer, whereas
full-output KL matching retains it. \arxivrev{The figure also reports
transfers lost or gained across all \ConfirmationInputs{} inputs in the
prospective comparison (Section~\ref{sec:single_scale}).}}

\arxivrev{To compare code effects under these controls, we hold the model,
dictionary, feature group and input pair fixed while varying the code and
strength control (Figure~\ref{fig:comparison_design}). We compare the
resulting effects with $E$ to assess intervention consistency, without
treating this agreement as a faithfulness standard.}

%% file: sections/02_related_work.tex
\section{\frank{Related Work}}
\label{sec:related}

\paragraph{Sparse codes and interventions.}
\frank{Superposition motivates sparse activation representations
\citep{elhage2022superposition}. SAE scaling and alternative encoders improve
the sparsity--reconstruction trade-off
\citep{gao2025scaling,rajamanoharan2024gated}.
For a fixed decoder, \citet{oneill2025compute} study the amortization gap
between encoder-based and optimized sparse inference.
At the intervention level, \citet[][Appendix~G]{carbo2026which} compare
feature and group ablations across inference methods. We compare
reconstruction-qualified codes for a fixed linguistic group under different
controls on intervention strength.}

\paragraph{Strength calibration.}
\frank{Activation Addition and Contrastive Activation Addition scale steering
vectors \citep{turner2024actadd,rimsky2024steering}. ActAdd also uses
positionwise norm-matched random controls \citep[][Appendix~H]{turner2024actadd}.
Activation-patching conclusions depend on metrics and corruption methods
\citep{zhang2024bestpractices,heimersheim2024patching}.
For output-based strength matching, \citet[][Appendix~C.2]{sun2026price}
match mean per-token KL within each setup, and \citet{luo2026steercheck}
match KL on neutral prompts to study control-direction alignment. Our comparison holds the feature group fixed
and includes controls with equal coefficient counts. Our shared-coefficient
follow-ups use other same-task inputs, so they do not evaluate
SteerCheck's neutral-prompt protocol.}

\paragraph{Evaluating causal explanations.}
\frank{Faithfulness concerns how explanations relate to model behavior
\citep{jacovi-goldberg-2020-towards}. Probe controls separate represented
information from probe learning \citep{hewitt-liang-2019-designing}, and
recoverability alone does not establish model use \citep{belinkov-2022-probing}.
Causal analyses examine agreement mechanisms, sparse circuits and
abstractions \citep{finlayson-etal-2021-causal,marks2025sparse,geiger2021causal}.
Subspace patching can exploit dormant pathways \citep{makelov2024subspace}.
RAVEL measures selective attribute transfer \citep{huang-etal-2024-ravel},
and SAEBench assesses several aspects of SAE quality
\citep{karvonen2025saebench}. Our endpoint is agreement between interventions
on a fixed group after strength control. Encoder agreement is a comparison
outcome, not a ground-truth faithfulness score
(Appendix~\ref{app:related_tests}).}

%% file: sections/02_background.tex
\section{\frank{Background}}
\label{sec:background}

\frank{\arxivrev{The SAE encoder maps} a model activation $h\in\mathbb{R}^d$ to a
nonnegative code $z\in\mathbb{R}_{\geq0}^p$. \arxivrev{The SAE decoder}
$D\in\mathbb{R}^{d\times p}$ contains $p$ feature directions\arxivrev{. The
reconstruction} is $b_{\mathrm{dec}}+Dz$, where
$b_{\mathrm{dec}}\in\mathbb{R}^d$ is the decoder bias. The reconstruction
residual is $h-(b_{\mathrm{dec}}+Dz)$. Throughout the comparison, the
dictionary and bias remain fixed while alternative inference changes the code.}

\frank{The released encoder supplies the reference code. A fixed feature
group $G\subseteq\{1,\ldots,p\}$ selects the decoder columns
$D_G\in\mathbb{R}^{d\times|G|}$ and corresponding code entries $z_G$.
\arxivrev{In the unscaled intervention, replacing} the recipient's group entries with the donor's adds the decoded
group difference to the native activation, retaining the recipient's other
features and reconstruction residual.}

%% file: sections/03_comparison_protocol.tex
\section{\frank{Intervention Comparison Protocol}}
\label{sec:protocol}

\frank{With model, dictionary, group and sentence pair fixed, we qualify
alternative codes and compare paired effects, transfer decisions and
matching coverage under each control. Figure~\ref{fig:comparison_design}
summarizes the four-step comparison.}

\subsection{\frank{Codes and Qualification}}

We compare encoder codes with nonnegative least-squares (NNLS) refits and a constructed
candidate. \frank{Each alternative's complete codes, including features
outside $G$, are nonnegative, use only the encoder's nonzero features and
have reconstruction error no greater than the encoder's, up to numerical tolerance.}
\frank{To qualify a comparison, the donor and recipient sentences must also
pass native-answer and reconstruction-fidelity checks}
(Appendix~\ref{app:eligibility}).

NNLS refits coefficients without a patch-norm constraint.
\frank{To test agreement under a norm constraint, the candidate}
jointly adjusts donor and recipient codes to reduce their patch's projection
onto the encoder patch, subject to equal norms at each position.
Construction uses reconstruction and geometry, not the target linguistic
effect. Feasibility is checked on the realized codes and patches, with optimizer
settings and complete qualification rules in
Appendix~\ref{app:construction}.

\subsection{\frank{Patch Construction and Outcomes}}

\frank{An \emph{input} $i$ is one directed transfer evaluation with a
specified donor and recipient. Each sentence pair yields two inputs.
At aligned patched position $t$, let $z_{d,i,t}^a$ and $z_{r,i,t}^a$
be the donor and recipient codes used by intervention $a$, and let
$h_{i,t}\in\mathbb{R}^d$ be the native recipient activation.
Using the decoder and group from Section~\ref{sec:background}, the decoded
patch $v_{i,t}^a\in\mathbb{R}^d$ and updated activation are}
\frank{\[
\begin{aligned}
v_{i,t}^a&=D_G(z_{d,i,t}^a-z_{r,i,t}^a)_G,\\
\widetilde h_{i,t}^a&=h_{i,t}+v_{i,t}^a,
\end{aligned}
\]}
which describe the unscaled intervention. Each strength control below rescales the original patch before application.
We patch aligned interior positions, leaving the beginning-of-sequence and
last prompt tokens unchanged. The recipient's other features and
reconstruction residual are retained.
\frank{The equations describe decoded displacements. Norm checks and
calibration use their stored execution-precision realizations
(Appendix~\ref{app:scalar_controls}).}

Let $m_i^0$ and $m_i(a)$ be recipient-answer minus donor-answer logit margins
before and after intervention $a$. We measure effect and effect gap by
\[
e_i(a)=m_i^0-m_i(a),\qquad g_i(a)=|e_i(a)-e_i(E)|,
\]
where $E$ denotes the reference intervention built from the released encoder's
codes, not a correctness standard.
\commentedit{Encoder patches, effects and transfers refer to those of $E$.}
A positive effect moves the recipient toward the donor
answer. A \emph{residual gap} is the effect gap after a stated strength control.
\frank{To track changes in the two-answer ranking, we also record a binary
\emph{transfer decision}. A \emph{strict transfer} occurs when $m_i(a)<0$,
so the donor answer outranks the recipient answer, independently of the
outcome under $E$. Comparing these decisions with $E$ gives transfers lost
or gained. Their sum is the number of \emph{transfer disagreements}.}

\subsection{\frank{Strength Controls}}

\frank{Below, $v_{i,t}^a$ denotes the stored execution-precision
realization of the decoded patch. Let $v_i^a$ concatenate these vectors
over input $i$'s aligned interior positions.}
\emph{Total-norm matching} multiplies this entire patch
by $\|v_i^E\|_2/\|v_i^a\|_2$. \emph{Positionwise norm matching} instead uses
$\|v_{i,t}^E\|_2/\|v_{i,t}^a\|_2$ at each position $t$ with a nonzero patch.
\frank{A positive whole-patch coefficient preserves geometric direction
before execution-precision rounding.}
Positionwise matching can redistribute its magnitude across token positions.
\frank{Total-norm coefficients are restricted to $[0,2]$.
Positionwise ratios are unrestricted.}
\frank{Undefined or out-of-tolerance controls remain attempted inputs
and are excluded from matched comparisons. Zero-norm cases and realized-norm
checks are specified in Appendix~\ref{app:scalar_controls}.}

For output-based calibration, we measure functional strength using the forward
KL between the native next-token distribution $p_i^0$ and distribution
$p_i^{a,\alpha}$ after scaling the whole patch by $\alpha$:
\[
K_i^a(\alpha)=D_{\mathrm{KL}}(p_i^0\Vert p_i^{a,\alpha}).
\]
\frank{Both distributions are over the model vocabulary. We use natural
logarithms, so KL is measured in nats.}
\emph{Full-output KL calibration (Q1)} chooses one coefficient per input
to match $K_i^E(1)$. \emph{Answer-excluded KL calibration (Q2)} removes the
two evaluated answer tokens, renormalizes each distribution over the
remaining vocabulary and matches that conditional KL to the corresponding
\commentedit{value under $E$}.

\frank{Both KL controls evaluate trial patches over $\alpha\in[0,2]$
with fixed tolerances and retain the best measured attempt on failure}
(Appendix~\ref{app:search}). A post hoc analysis continues unmatched
candidate searches to $\alpha=4$ (Section~\ref{sec:candidate}). Every control rescales the original
alternative patch.
\frank{Because complete-code qualification precedes this scaling, it does
not guarantee that the calibrated patch is realizable by a code satisfying
the same constraints.}
\frank{Total-norm matching and full-output KL calibration are
\emph{coefficient-count matched}. Full-output KL calibration also requires
evaluation-input output information and additional model evaluations.
It measures target-effect disagreement after matching per-input KL.}

%% file: sections/03_experiments.tex
\section{\frank{Experiments}}
\label{sec:lexical}

\commentedit{We compare NNLS effect gaps from $E$ across total-norm and full-output
KL matching (\textbf{RQ1}), positionwise matching (\textbf{RQ2}), and
answer-excluded calibration (\textbf{RQ3}). \textbf{RQ4} tests whether calibration
reduces norm-matched candidate gaps and with what matching coverage.}

%% file: sections/03_audit_design.tex
\subsection{\frank{Experimental Setup}}
\label{sec:setup}

\begin{table}[t]
\centering
\small
\setlength{\tabcolsep}{3.5pt}
\begin{tabular}{@{}lrrrrr@{}}
\toprule
Configuration & Width & Layer & $|G|$ & NNLS & Cand.\\
\midrule
Gemma-9B-it SVA & 131,072 & 20 & 6 & 166 & 504\\
Gemma-2B SVA & 131,072 & 12 & 6 & 62 & 64\\
Llama-8B SVA & 131,072 & 15 & 12 & 188 & 124\\
\quad New material & 131,072 & 15 & 12 & \ConfirmationInputs{} & ---\\
Llama-8B SVA & 32,768 & 15 & 12 & 112 & 106\\
Gemma-2B gender & 131,072 & 12 & 12 & 42 & 40\\
\bottomrule
\end{tabular}
\caption{\frank{Configurations and eligible inputs before matching.
Width denotes dictionary size, Layer is zero-indexed, and $|G|$ is group size.
SVA denotes subject--verb agreement. Llama-128K/32K indicate dictionary widths.
The 9B NNLS and candidate panels are separate. New Llama material has no candidate comparison.}}
\label{tab:settings}
\end{table}


\noindent\textbf{Tasks.}
Subject--verb agreement across an intervening noun tests hierarchical
syntactic behavior \citep{linzen-etal-2016-assessing,marvin-linzen-2018-targeted}.
Our sentence pairs change subject number and retain a prepositional phrase
containing a potentially distracting noun, called the attractor. We test the
next-token preference for \emph{is} versus \emph{are}.
Reflexive gender agreement supplies a second contrast:
explicit male/female cues change the expected continuation
(\emph{himself}/\emph{herself}), as detailed in Appendix~\ref{app:gender_materials}.

\noindent\textbf{Models and feature groups.}
We evaluate Gemma-2-9B-it and Gemma-2-2B base
\citep{gemmateam2024gemma2} with Llama-3.1-8B base
\citep{grattafiori2024llama3}, using Gemma Scope
\citep{lieberum2024gemmascope} and Llama Scope
\citep{he2024llamascope} dictionaries (Table~\ref{tab:settings}). We select groups with encoder
interventions on development sentences, then fix them for code comparisons.
Gender uses a separate group on the 2B model.

\noindent\textbf{Evaluation and coverage.}
A \emph{panel} is a source set of sentence pairs, a \emph{cohort} the eligible
inputs for a comparison, and a \emph{matched subset} the inputs meeting its
\commentedit{matching} tolerance. Lexical blocks group inputs sharing assigned nouns.
Cases index subject--attractor number combinations for subject--verb agreement
and cue--predicate combinations for gender.
A coverage pass requires prespecified per-case retention and block support,
allowing some inputs to remain unmatched.

\noindent\textbf{Aggregation and uncertainty.}
\frank{Pooled means weight inputs equally. Balanced means weight cases
equally and, where specified, blocks within cases. Counts are unweighted.
Where estimable, pointwise 95\% confidence intervals describe paired
mean contrasts. We condition on retained material and assume independent
lexical blocks, using paired block bootstrap or, for 9B candidate analyses,
block-based Student-$t$ intervals
(Appendix~\ref{app:uncertainty}). These are not simultaneous intervals.}

\noindent\textbf{Study design.}
\frank{The primary comparison is the 9B candidate reduction,
specified before group effects were measured on its lexical panel.
Llama-128K NNLS confirmation fixes new material and analysis before
measurement within an already selected configuration
(Appendix~\ref{app:confirmation}). Other controls, including Q2 on that
panel, are follow-ups on previously inspected material
(Appendix~\ref{app:stops}). The five pre-confirmation NNLS cohorts are the
\emph{earlier cohorts}. We never pool them with the new cohort.}

\input{figures/nnls_controls}

%% file: figures/nnls_controls.tex
\begin{figure*}[t]
\centering
\definecolor{positionwise}{RGB}{62,130,44}
\begin{tikzpicture}[x=1cm,y=1cm,font=\small]
\def\xa{3.45}   
\def\xb{11.27}  
\def\s{18.5}    
\newcommand{\controlArow}[9]{%
  \begin{scope}[y=.85cm]
  \node[anchor=west] at (0,#1) {#2};
  \node at (2.45,#1) {#3};
  \ifnum#7=1
    \draw[blue!60!black,line width=1pt] ({\xa+\s*#5},#1) -- ({\xa+\s*#6},#1);
    \foreach \v in {#5,#6} {\draw[blue!60!black] ({\xa+\s*\v},#1-.07) -- ({\xa+\s*\v},#1+.07);}
  \else
    \node[anchor=west,inner sep=1pt] at ({\xa+\s*#4+.13},#1) {CI n/a};
  \fi
  \fill[blue!60!black] ({\xa+\s*#4},#1) circle (2.5pt);
  \node at (7.8,#1) {#8};
  \node at (9.0,#1) {#9};
  \end{scope}
}
\newcommand{\controlBrow}[6]{%
  \begin{scope}[y=.85cm]
  \node at (10.3,#1) {#2};
  \ifnum#6=1
    \draw[positionwise!85!black,line width=1pt] ({\xb-\s*#4},#1) -- ({\xb-\s*#5},#1);
    \foreach \v in {#4,#5} {\draw[positionwise!85!black] ({\xb-\s*\v},#1-.07) -- ({\xb-\s*\v},#1+.07);}
  \else
    \node[anchor=west,inner sep=1pt] at ({\xb-\s*#3+.13},#1) {CI n/a};
  \fi
  \fill[positionwise!85!black] ({\xb-\s*#3-.065},#1-.065) rectangle ({\xb-\s*#3+.065},#1+.065);
  \end{scope}
}
\fill[black!5] (0,-.15) rectangle (15.35,.32);
\node[anchor=west,font=\small\itshape] at (0,3.40) {Earlier cohorts};
\node[anchor=west,font=\small\itshape] at (0,.52) {Prospective cohort};
\node[anchor=west] at (0,3.83) {Cohort};
\node at (2.45,3.83) {$n$};
\node at (7.8,3.83) {Total norm};
\node at (9.0,3.83) {Q1};
\node at (10.3,3.83) {$n$};
\foreach \x in {\xa,\xb} {
  \foreach \v in {.1,.2} {\draw[black!10] ({\x+\s*\v},-.35) -- ({\x+\s*\v},3.58);}
  \draw[black!45,dashed] (\x,-.35) -- (\x,3.58);
}
\draw[black!20] (9.72,-.15) -- (9.72,4.02);
\ControlForestA
\ControlForestB
\foreach \x/\q in {\xa/Q1,\xb/Positionwise} {
  \draw ({\x-\s*.02},-.35) -- ({\x+\s*.2},-.35);
  \foreach \v/\lab in {0/0,0.1/0.1,0.2/0.2} {
    \draw ({\x+\s*\v},-.35) -- ({\x+\s*\v},-.42);
    \node[anchor=north] at ({\x+\s*\v},-.44) {\lab};
  }
  \node at ({\x+\s*.09},-1.12) {Gap reduction (logit)};
}
\node[font=\small\bfseries,text height=1.5ex,text depth=.25ex] at (4.86,-1.67) {(a) Q1 versus total norm};
\node[font=\small\bfseries,text height=1.5ex,text depth=.25ex] at (12.55,-1.67) {(b) Positionwise versus total norm};
\end{tikzpicture}
\caption{\textbf{NNLS gap reductions relative to total-norm matching.}
\frank{Full-output KL calibration \commentedit{(Q1, panel a)} reduces mean gaps in every cohort.} Positionwise matching (b) improves on
total-norm matching in most observed means.
Points show mean paired differences in absolute effect gaps from the
\commentedit{encoder-based reference intervention $E$} (total-norm minus alternative), so positive
values favor the alternative.
Each comparison uses its $n$ common matched inputs. Means are case/block-balanced (pooled for 9B). Bars show pointwise 95\% paired block-bootstrap intervals, and ``CI n/a''
marks an unavailable interval. \commentedit{Transfer counts report lost/gained relative to $E$.}}
\label{fig:nnls_controls}
\end{figure*}
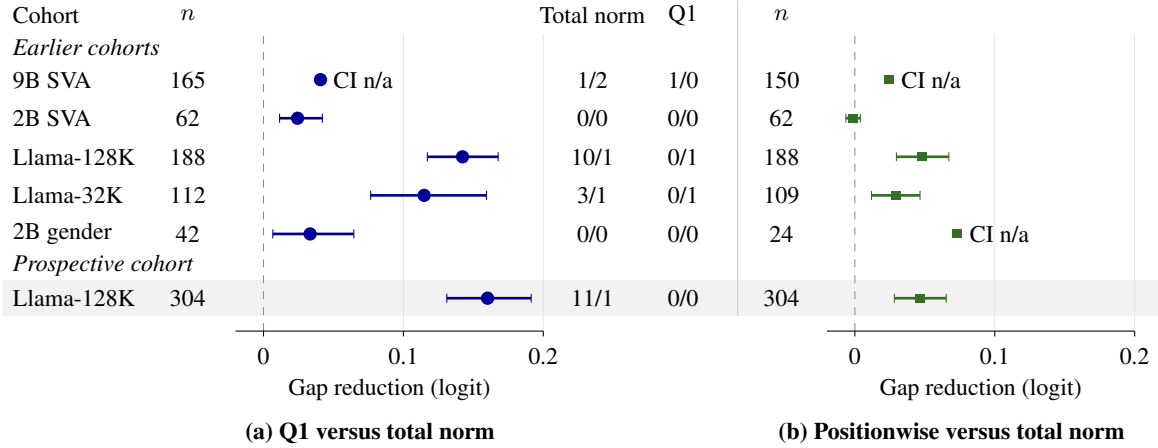

%% file: sections/04_lexical_calibration.tex
\subsection{\frank{RQ1: Total-Norm versus Full-Output KL}}
\label{sec:single_scale}

\frank{Total-norm and full-output KL matching both rescale the original
NNLS patch with one coefficient per input. We compare them on each cohort's
common inputs (Figure~\ref{fig:nnls_controls}a).}

We find smaller \commentedit{mean} residual gaps under \frank{full-output KL calibration} in all five earlier cohorts
(Figure~\ref{fig:nnls_controls}a), with paired intervals excluding zero in
the four cohorts where they are estimable.
The prospective comparison confirms this ordering on new material within
the selected Llama-128K configuration. Both controls match all
\ConfirmationInputs{} inputs, \frank{with balanced mean gaps of 0.181 under
total-norm matching and \ConfirmationQoneGap{} under full-output KL calibration. Their paired
difference is \ConfirmationPrimary{} logit.} The interval in Figure~\ref{fig:nnls_controls}a
meets the prespecified precision target (Appendix~\ref{app:confirmation}).

\frank{Although control differences are small relative to encoder effects
in the five earlier common-input cohorts (Appendix~\ref{app:effect_scale}),
they can change decisions near the transfer threshold.} On the new Llama inputs, total-norm matching
loses \ConfirmationTotalLost{} of $E$'s \ConfirmationEncoderTransfers{}
transfers and gains \ConfirmationTotalGained{} elsewhere.
\frank{Figure~\ref{fig:overview} illustrates one such loss.
Full-output KL calibration preserves every}
decision on this panel. In the earlier 9B cohort, however,
both controls lose one transfer. Smaller mean gaps therefore do not
guarantee preservation of individual transfer decisions.

\input{figures/control_boundaries}

\subsection{\frank{RQ2: Positionwise Norm Matching}}
\label{sec:positionwise}

\frank{Both controls in RQ1 scale all patched positions together.
We next compare total-norm matching with positionwise matching, which
allows a separate coefficient at each position. These ratios also lack
the whole-patch coefficient bound, so the comparison does not isolate
granularity from the allowed range.}
Figure~\ref{fig:nnls_controls}b uses the inputs matched by these two controls,
separately from the \frank{full-output KL} comparison in panel (a).

Positionwise norm matching leaves a smaller observed mean gap than total-norm
matching in four earlier cohorts and the prospective cohort
(Figure~\ref{fig:nnls_controls}b). The remaining 2B subject--verb comparison
has a slightly higher mean gap under positionwise matching, but the paired
analysis does not establish a clear advantage for either norm control.
On the new Llama inputs, \frank{full-output KL calibration} still leaves smaller \commentedit{mean} gaps than either norm
control (Figure~\ref{fig:control_boundaries}a).

\frank{On the 24 gender inputs matched by all three controls, positionwise
matching yields the closest observed agreement with \commentedit{$E$}}
(Figure~\ref{fig:control_boundaries}a).
\frank{Their balanced gaps are \GenderCommonPosition{} for positionwise
matching, \GenderCommonQone{} for full-output KL and \GenderCommonTotal{}
for total-norm matching. This 24-input comparison is descriptive, with no
estimable paired interval. The total-norm/full-output KL comparison in
RQ1 uses all 42 eligible inputs.}
\frank{A geometric diagnostic of the 24-input subset finds that the original
NNLS patches are nearly parallel to the encoder patches but distribute
their norm differently across positions} (Appendix~\ref{app:scalar_controls}).
The contrasting Llama and gender orderings show that whole-patch output
calibration need not give the closest agreement, though differences in
model and feature group prevent attributing the contrast to the task alone.

%% file: figures/control_boundaries.tex
\begin{figure}[t]
\centering
\definecolor{positionwise}{RGB}{62,130,44}
\begin{tikzpicture}[x=1cm,y=1cm,font=\small]
\newcommand{\positionmarks}[4]{%
  \node at (2.20,0) {(#4)};
  \draw[black!40,line width=.8pt] ({2.6+22*#1},0) -- ({2.6+22*#3},0);
  \draw[orange!70!black,fill=white,line width=1pt] ({2.6+22*#1},0) circle (3pt);
  \fill[blue!60!black] ({2.6+22*#2},0) circle (2.8pt);
  \fill[positionwise!85!black] ({2.6+22*#3-.075},-.075) rectangle ({2.6+22*#3+.075},.075);
}
\pgfmathsetmacro{\boundaryCriterionScale}{(7-2.6)/(0.1+0.25)}
\pgfmathsetmacro{\boundaryCriterionZero}{2.6+0.25*\boundaryCriterionScale}
\newcommand{\criterionmarks}[4]{%
  \node at (2.20,0) {\commentedit{(#4)}};
  \draw[black!75,line width=1pt] ({\boundaryCriterionZero+\boundaryCriterionScale*#2},0) -- ({\boundaryCriterionZero+\boundaryCriterionScale*#3},0);
  \foreach \v in {#2,#3} {\draw[black!75] ({\boundaryCriterionZero+\boundaryCriterionScale*\v},-.09) -- ({\boundaryCriterionZero+\boundaryCriterionScale*\v},.09);}
  \fill[black!75] ({\boundaryCriterionZero+\boundaryCriterionScale*#1},0) circle (2.8pt);
}
\draw[orange!70!black,fill=white,line width=1pt] (.1,2.25) circle (3pt);
\node[anchor=west] at (.27,2.25) {Total norm};
\fill[blue!60!black] (2.78,2.25) circle (2.8pt);
\node[anchor=west] at (2.95,2.25) {Q1};
\fill[positionwise!85!black] (4.08,2.175) rectangle (4.23,2.325);
\node[anchor=west] at (4.33,2.25) {Positionwise};
\foreach \v in {0,.1,.2} {\draw[black!12] ({2.6+22*\v},.5) -- ({2.6+22*\v},1.9);}
\begin{scope}[yshift=1.6cm]
  \node[anchor=west] at (0,0) {Llama-128K};
  \expandafter\positionmarks\BoundaryPositionLlama
\end{scope}
\begin{scope}[yshift=.95cm]
  \node[anchor=west] at (0,0) {2B gender};
  \expandafter\positionmarks\BoundaryPositionGender
\end{scope}
\draw (2.6,.5) -- (7,.5);
\foreach \v/\lab in {0/0,0.1/0.1,0.2/0.2} {
  \draw ({2.6+22*\v},.5) -- ({2.6+22*\v},.43);
  \node[anchor=north] at ({2.6+22*\v},.4) {\lab};
}
\node at (4.8,-.3) {Mean effect gap (logit), $\leftarrow$ closer to $E$};
\node[font=\small\bfseries] at (3.5,-.8) {(a) Positionwise matching versus Q1};
\begin{scope}[yshift=-3.25cm]
\draw[black!45,dashed] (\boundaryCriterionZero,.5) -- (\boundaryCriterionZero,1.9);
\begin{scope}[yshift=1.6cm]
  \node[anchor=west] at (0,0) {Llama-128K};
  \expandafter\criterionmarks\BoundaryCriterionLlama
\end{scope}
\begin{scope}[yshift=.95cm]
  \node[anchor=west] at (0,0) {2B gender};
  \expandafter\criterionmarks\BoundaryCriterionGender
\end{scope}
\draw (2.6,.5) -- (7,.5);
\foreach \v/\lab in {-0.2/{-0.2},-0.1/{-0.1},0/0,0.1/0.1} {
  \draw ({\boundaryCriterionZero+\boundaryCriterionScale*\v},.5) -- ({\boundaryCriterionZero+\boundaryCriterionScale*\v},.43);
  \node[anchor=north] at ({\boundaryCriterionZero+\boundaryCriterionScale*\v},.4) {$\lab$};
}
\node at (4.8,-.3) {Change in mean gap (logit)};
\node[anchor=west,inner sep=0pt] at (2.6,-.75) {$\leftarrow$ Q2 closer to $E$};
\node[anchor=east,inner sep=0pt] at (7,-.75) {farther $\rightarrow$};
\node[font=\small\bfseries] at (3.5,-1.25) {(b) Q2 versus unscaled NNLS};
\end{scope}
\end{tikzpicture}
\caption{\textbf{NNLS control orderings differ across settings.}
\frank{Q1 denotes full-output KL, Q2 answer-excluded KL.}
(a) Case/block-balanced mean gaps from \commentedit{the encoder-based reference intervention $E$} on inputs matched by all three controls
in each row, shown descriptively without intervals. 
(b) Case/block-balanced mean paired gap differences (Q2 minus unscaled NNLS), with pointwise 95\% block-bootstrap intervals. \commentedit{Parentheses give input counts. Gender subsets differ between panels.} Llama rows use the new cohort, where Q2 is a
follow-up after inspection.}
\label{fig:control_boundaries}
\end{figure}
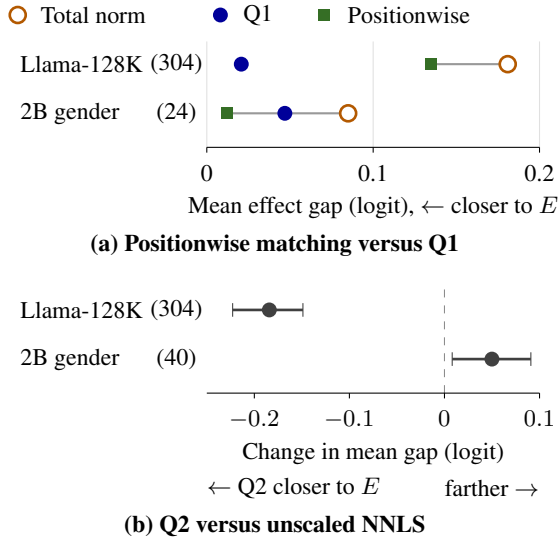

%% file: sections/05_answer_excluded.tex
\subsection{\frank{RQ3: Answer-Excluded Calibration}}
\label{sec:excluded}

\frank{Full-output KL includes the answer tokens used to measure the target effect.
We compare it with answer-excluded KL under the same search bounds to test
how removing those tokens changes NNLS effect agreement.}

In the earlier 2B and Llama NNLS cohorts for subject--verb agreement, \frank{answer-excluded KL}
matches every input and leaves pooled mean residual gaps of
\NnlsRestMin{}--\NnlsRestMax{} logit
(Appendices~\ref{app:2b_analysis} and~\ref{app:llama}). The earlier 9B NNLS
cohort has no \frank{answer-excluded KL} measurement.

\commentedit{On the new Llama panel, this follow-up after inspection of the
confirmation results matches all \FollowupQtwoInputs{} inputs and reduces
the balanced mean gap relative to unscaled NNLS
(Figure~\ref{fig:control_boundaries}b). It leaves a larger gap than full-output
KL (\FollowupQtwoGap{} versus \ConfirmationQoneGap{} logit) and loses
\FollowupQtwoLost{} transfers that full-output KL preserves
(Table~\ref{tab:confirmation_q2}, Appendix~\ref{app:confirmation_q2}).}

On the 40 gender inputs eligible for both NNLS and candidate comparisons,
\frank{answer-excluded KL successfully matches all inputs} but increases the balanced mean gap relative to
unscaled NNLS. \frank{The paired interval for this increase lies above zero} (Figure~\ref{fig:control_boundaries}b).
The reduction observed in subject--verb
agreement therefore does not extend to this gender comparison
(\frank{Appendix~\ref{app:gender_comparisons}}, Table~\ref{tab:gender_criteria}).

To characterize the gender mismatch, we separate full KL into a term for the total
probability assigned to the two answers and two terms for the normalized
distributions within the answer and remaining-vocabulary sets. The latter
terms are weighted by each set's native probability. Under \frank{answer-excluded KL}, the weighted
remaining-vocabulary term moves closer to the encoder's corresponding value,
but the within-answer term moves farther away. This decomposition locates
the mismatch without identifying its cause
(Appendix~\ref{app:kl_partition}).

%% file: figures/core_calibration.tex
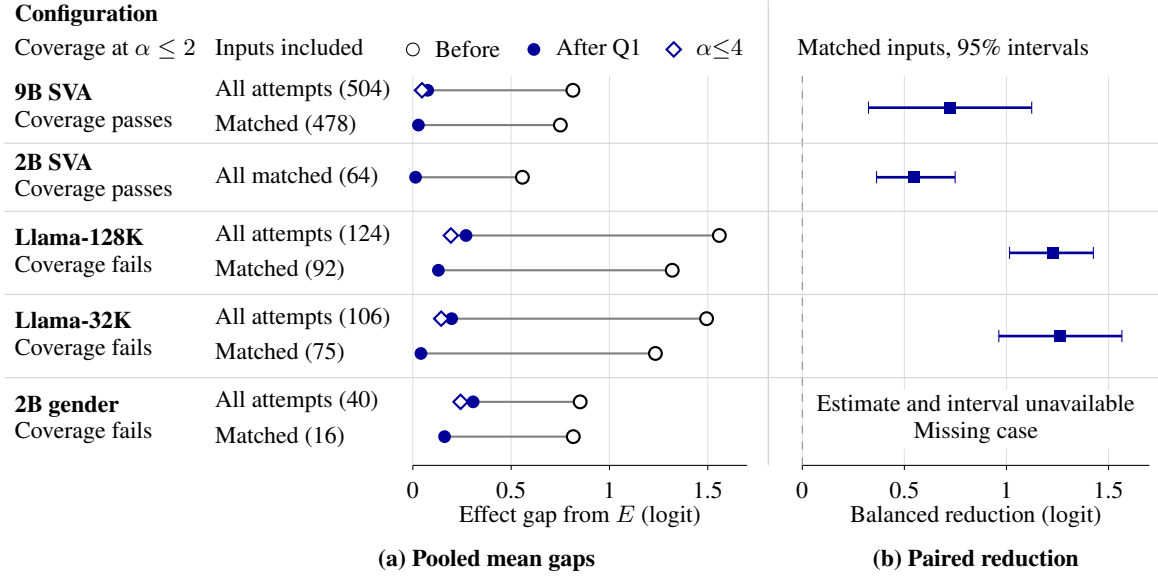
\begin{figure*}[t]
\centering
\begin{tikzpicture}[x=1cm,y=1cm,font=\small]
\def\gapzero{5.4}
\def\gapscale{2.6}
\def\redzero{10.55}
\def\redscale{2.7}
\newcommand{\gaprow}[4]{%
  \node[anchor=west] at (2.65,#1) {#2};
  \draw[black!50,line width=.8pt] ({\gapzero+\gapscale*#3},#1) -- ({\gapzero+\gapscale*#4},#1);
  \draw[black,fill=white,line width=.8pt] ({\gapzero+\gapscale*#3},#1) circle (2.4pt);
  \fill[blue!60!black] ({\gapzero+\gapscale*#4},#1) circle (2.4pt);
}
\newcommand{\rangemark}[2]{%
  \draw[blue!60!black,densely dashed,line width=.7pt] ({\gapzero+\gapscale*#1},.23) -- ({\gapzero+\gapscale*#2},.23);
  \draw[blue!60!black,fill=white,line width=.8pt]
    ({\gapzero+\gapscale*#2},.325) -- ({\gapzero+\gapscale*#2+.095},.23) -- ({\gapzero+\gapscale*#2},.135) -- ({\gapzero+\gapscale*#2-.095},.23) -- cycle;
}
\newcommand{\candidatevalues}[6]{%
  \ifnum#1=#4
    \gaprow{0}{All matched (#1)}{#2}{#3}
  \else
    \gaprow{.23}{All attempts (#1)}{#2}{#3}
    \gaprow{-.23}{Matched (#4)}{#5}{#6}
  \fi
}
\newcommand{\candidatereduction}[3]{%
  \draw[blue!60!black,line width=1pt] ({\redzero+\redscale*#2},0) -- ({\redzero+\redscale*#3},0);
  \foreach \v in {#2,#3} {\draw[blue!60!black] ({\redzero+\redscale*\v},-.08) -- ({\redzero+\redscale*\v},.08);}
  \fill[blue!60!black] ({\redzero+\redscale*#1-.075},-.075) rectangle ({\redzero+\redscale*#1+.075},.075);
}
\foreach \zero/\scale in {\gapzero/\gapscale,\redzero/\redscale} {
  \foreach \x/\lab in {0/0,0.5/0.5,1/1,1.5/1.5} {
    \draw[black!12] ({\zero+\scale*\x},.1) -- ({\zero+\scale*\x},5.25);
    \draw ({\zero+\scale*\x},.1) -- ({\zero+\scale*\x},.04);
    \node[anchor=north] at ({\zero+\scale*\x},.02) {\lab};
  }
  \draw (\zero,.1) -- ({\zero+\scale*1.7},.1);
}
\draw[black!45,dashed] (\redzero,.1) -- (\redzero,5.25);
\draw[black!20] (10.1,.1) -- (10.1,6.25);
\foreach \y in {4.36,3.46,2.36,1.26} {\draw[black!18] (0,\y) -- (15.3,\y);}
\begin{scope}[shift={(0,4.83)}]
\node[anchor=west,align=left] at (0,0) {\textbf{9B SVA}\\Coverage passes};
\expandafter\candidatevalues\StrengthCandidateNine
\expandafter\rangemark\StrengthRangeNine
\CandidateReductionNine
\end{scope}
\begin{scope}[shift={(0,3.91)}]
\node[anchor=west,align=left] at (0,0) {\textbf{2B SVA}\\Coverage passes};
\expandafter\candidatevalues\StrengthCandidateTwo
\CandidateReductionTwo
\end{scope}
\begin{scope}[shift={(0,2.91)}]
\node[anchor=west,align=left] at (0,0) {\textbf{Llama-128K}\\Coverage fails};
\expandafter\candidatevalues\StrengthCandidateLlamaWide
\expandafter\rangemark\StrengthRangeLlamaWide
\CandidateReductionLlamaWide
\end{scope}
\begin{scope}[shift={(0,1.81)}]
\node[anchor=west,align=left] at (0,0) {\textbf{Llama-32K}\\Coverage fails};
\expandafter\candidatevalues\StrengthCandidateLlamaNarrow
\expandafter\rangemark\StrengthRangeLlamaNarrow
\CandidateReductionLlamaNarrow
\end{scope}
\begin{scope}[shift={(0,.71)}]
\node[anchor=west,align=left] at (0,0) {\textbf{2B gender}\\Coverage fails};
\expandafter\candidatevalues\StrengthCandidateGender
\expandafter\rangemark\StrengthRangeGender
\CandidateReductionGender
\end{scope}
\node[anchor=west,font=\small\bfseries] at (0,6.05) {Configuration};
\node[anchor=west] at (0,5.6) {Coverage at $\alpha\le2$};
\node[anchor=west] at (2.65,5.6) {Inputs included};
\draw[black,fill=white] (5.4,5.6) circle (2.4pt);
\node[anchor=west] at (5.56,5.6) {Before};
\fill[blue!60!black] (7.0,5.6) circle (2.4pt);
\node[anchor=west] at (7.16,5.6) {After Q1};
\draw[blue!60!black,fill=white,line width=.8pt]
  (8.85,5.695) -- (8.945,5.6) -- (8.85,5.505) -- (8.755,5.6) -- cycle;
\node[anchor=west] at (9.01,5.6) {$\alpha{\le}4$};
\node[anchor=west] at (10.35,5.6) {\frank{Matched inputs, 95\% intervals}};
\node at (7.61,-.57) {Effect gap from $E$ (logit)};
\node at (12.845,-.57) {Balanced reduction (logit)};
\node[font=\small\bfseries,text height=1.5ex,text depth=.25ex] at (6.375,-1.12) {(a) Pooled mean gaps};
\node[font=\small\bfseries,text height=1.5ex,text depth=.25ex] at (12.845,-1.12) {(b) Paired reduction};
\end{tikzpicture}
\caption{\frank{\textbf{Constructed candidates under full-output KL calibration (Q1).}
\textbf{(a)} Pooled gaps \commentedit{from the encoder-based reference intervention $E$}
before (open circles) and after (filled circles)
search within $\alpha\le2$. All-attempt rows retain failed searches at their
best measured point. Diamonds show post hoc continuation to $\alpha\le4$
(Appendix~\ref{app:range}).
\textbf{(b)} Paired reductions on original matched inputs, case-balanced
for 9B and case/block-balanced otherwise. Pointwise 95\% intervals use
lexical-block Student-$t$ for 9B and paired block bootstrap elsewhere.
The 9B primary interval's half-width exceeds the planned
\CandidatePrecisionGoal{} logit. Gender lacks a balanced estimate and interval because one case
has no matches. Coverage follows the original per-case rule. Llama
intervals describe selected subsets.}}
\label{fig:core_calibration}
\end{figure*}

%% file: sections/04_candidate_stress.tex
\subsection{\frank{RQ4: Constructed Candidate Stress Tests}}
\label{sec:candidate}

\frank{For RQ4, the} candidate comparisons test whether \frank{full-output KL calibration} also reduces the larger effect gaps
allowed by the reconstruction and positionwise norm constraints.
\frank{Figure~\ref{fig:core_calibration} separates matched-input and
all-attempt results, with paired intervals on the original matched subsets.}

\frank{The prespecified 9B primary comparison uses 478 successful matches
out of 504 attempts. On this subset, full-output KL reduces the
case-balanced mean gap from 0.752 to 0.028 logit. The paired reduction from
the unscaled candidate is \CandidatePrimaryReduction{} logit, with an
interval wider than planned (Figure~\ref{fig:core_calibration}b).}
Transfer disagreements fall from 77 to 8 on this
subset (case-level and block-deletion checks in
Appendix~\ref{app:lexical_details}).

All 64 qualified 2B subject--verb candidate inputs match, so calibration
introduces no further subset restriction. Their balanced residual gap is
0.015 logit, with no remaining transfer disagreements
(Appendix~\ref{app:2b_analysis}). Within the prespecified range $\alpha\le2$,
\commentedit{mean} candidate gaps decrease on matched inputs in the Llama configurations and
the gender setting, but these comparisons fail the per-case coverage requirements.
The gender candidate matches only 16 of 40 inputs,
including none in one case, so the Llama and gender matched results describe
retained subsets.

Answer exclusion also reduces candidate gaps in the measured subject--verb
settings, subject to the same coverage qualifications
(Appendices~\ref{app:answer_excluded}, \ref{app:2b_analysis} and~\ref{app:llama}).
On the common 40 gender inputs, \commentedit{mean} candidate gaps decrease across \frank{answer-excluded KL} attempts
even though NNLS gaps increase (Table~\ref{tab:gender_criteria}).
Only 13 candidate inputs match within the original range, limiting this to
a descriptive contrast between codes with unequal coverage.

Coverage partly reflects the finite search range. A post hoc continuation
to $\alpha=4$ matches \RangeRescued{} of \RangeSearches{} previously failed
search records, which overlap across \frank{the two KL controls}, without changing any original
selection or coverage decision. Diamonds in Figure~\ref{fig:core_calibration}a
show the extended all-attempt means. Newly matched inputs generally have
larger gaps than the original matched subset, and \frank{the Llama-32K
full-output KL comparison and both gender comparisons still fail coverage}
(Appendix~\ref{app:range}).

%% file: sections/06_global_transfer.tex
\subsection{\frank{Additional Analyses}}
\paragraph{\frank{Shared coefficients.}}
\label{sec:global}
\frank{To examine coefficient selection from other inputs, we test shared
candidate coefficients using mean full-output KL on the two previously
inspected Gemma subject--verb panels.} \frank{The 9B selector uses a word-disjoint calibration split.
The 2B selector leaves out each evaluation block.}
This changes both coefficient granularity and selection
information. \frank{Balanced mean gaps fall from 0.876 to 0.704 in 9B and
from 0.561 to 0.313 in 2B. Individual gaps nevertheless increase on
\GlobalNineWorsened{} of \GlobalNineInputs{} inputs in 9B and
\GlobalTwoWorsened{} of \GlobalTwoInputs{} in 2B, relative to the unscaled candidates}
(Appendix~\ref{app:global_distributions}, Figure~\ref{fig:global_distribution}).
\frank{Even the effect-optimal shared coefficient chosen in hindsight on
the tested nine-point grid leaves larger mean gaps than per-input
full-output KL on these panels.}

%% file: sections/05_multibudget.tex
\paragraph{\frank{Strength budgets.}}
\label{sec:calibration_scope}
\label{sec:budgets}
\frank{To examine agreement at other strength levels, we vary target
budgets of answer-excluded KL and independently scale encoder and
candidate patches to each target.}
\frank{Among inputs where both patches match, mean gaps remain small.
Coverage falls at higher budgets, restricting those means to smaller subsets}
(Appendix~\ref{app:response_details}).

%% file: sections/07_interpretation.tex
\section{Discussion}
\label{sec:interpretation}

\frank{Small mean effect gaps do not ensure preservation of encoder
transfer decisions (Figure~\ref{fig:nnls_controls}a). Reporting paired gaps
and transfer gains and losses on the same inputs reveals disagreements
that mean improvements or net transfer counts can hide. The differing
control orderings in Figure~\ref{fig:control_boundaries}a also require
specifying the matched quantity, coefficient granularity and selection
information. Agreement with $E$ alone does not establish which code better
captures the model's linguistic mechanism.}

\paragraph{Selection information.}
\frank{Per-input KL calibration measures disagreement after using reference
outputs and repeated evaluations to match a scalar KL value. Comparisons with total-norm matching change the matching criterion, selection
information and computation together. 
Answer-related information may remain in the vocabulary even after
the two evaluated answer tokens are excluded.
Using shared coefficients selected on other inputs changes both
input-specific flexibility and selection information, so it does
not isolate the effect of information access. Applications requiring coefficients fixed before
evaluation need results under that selection protocol.}

\paragraph{\frank{Coverage.}}
\frank{Retaining failed searches exposes larger residual gaps
(Figure~\ref{fig:core_calibration}). Matched-subset summaries can therefore
omit the hardest comparisons. Report the search range and coverage by
condition alongside both matched-subset and all-attempt results, including
inputs whose gaps increase despite a mean reduction.}

%% file: sections/10_conclusion.tex
\section{Conclusion}
\label{sec:conclusion}

With one coefficient per input, \frank{full-output KL calibration} leaves smaller mean NNLS gaps than
total-norm matching across five earlier cohorts, with prospective confirmation
in Llama-128K. On that new panel, total-norm matching loses encoder transfers
that \frank{full-output KL calibration preserves}.
\frank{Full-output KL need not give the closest agreement: positionwise
matching does so on a descriptive 24-input gender subset. Answer exclusion
also increases gender NNLS gaps relative to the unscaled patches. Calibration also reduces larger gaps from constructed candidates,
with matching coverage dependent on the search range.}

%% file: sections/11_limitations.tex
\section*{Limitations}
\label{sec:limitations}

\begingroup
\interlinepenalty=10000
\noindent\frank{\textbf{Scope and selection.} Our five selected configurations cover
two restricted next-token phenomena, with model, layer, dictionary and
feature group varying together. They do not isolate task effects or
represent unrestricted generation. The new-material Llama confirmation
strengthens evidence within a configuration chosen using earlier outcomes.
Its Q2 extension remains a follow-up, and neither analysis resolves adaptive
selection or shared-template dependence.}\par
\endgroup

\noindent\frank{\textbf{Reference and calibration.} Encoder agreement does not
establish explanation faithfulness. Q1 uses evaluation-input outputs and
additional model evaluations, so equal coefficient counts do not equalize
information or computation. Q2 can retain answer-related information outside
the excluded tokens. Scaled patches need separate qualification to satisfy
complete-code constraints.}

\noindent\frank{\textbf{Coverage and finite search.} Qualification excludes many
settings, and matching further restricts comparisons
(Appendix~\ref{app:stops}). Candidate coverage fails within the original
range. Expanding the range recovers matches but leaves some failures
(Appendix~\ref{app:range}). Gender positionwise results concern a small
subset. Finite searches limit conclusions about unmeasured scales.
Feature-group specificity at original strengths remains unestablished
(Appendix~\ref{app:random}).}

\noindent\frank{\textbf{Statistical uncertainty.} Pointwise intervals condition on
retained lexical blocks and assume block independence, which shared
templates or semantic structure can violate (Appendix~\ref{app:uncertainty}).
They do not adjust for selection across the research program. The
prespecified 9B candidate estimate misses planned precision. Earlier 9B NNLS
and gender positionwise comparisons lack block intervals.}

%% file: sections/12_ethics.tex
\section*{Ethics Statement}
\label{sec:ethics}

The experiments study model outputs on templated sentence prefixes, without
human participants or personal records. The main lexical inventory was
authored for the study and contains common nouns denoting people and roles.
It is not a representative sample of English usage or of any demographic
group. The gender materials manipulate explicit male/female cues to test
the model's reflexive continuations. They do not operationalize gender
identity or support conclusions about fairness. Neither phenomenon is a
test of social bias or model safety.

Gemma 2 weights are distributed under the Gemma Terms of Use, and the Gemma
Scope SAE repositories used here specify CC BY 4.0.\footnote{Official sources:
\url{https://ai.google.dev/gemma/terms};
\url{https://huggingface.co/google/gemma-scope-9b-it-res};
\url{https://huggingface.co/google/gemma-scope-2b-pt-res}.}
Public availability does not remove these conditions.
Llama weights were accessed through the official gated release under the
Llama 3.1 license. The Llama Scope umbrella card declares Apache 2.0, but
the two pinned checkpoint repositories have no separate license metadata
in the archived asset records. We therefore distinguish that umbrella
declaration from checkpoint-specific redistribution clearance and do not
redistribute either model or SAE weights with this manuscript.\footnote{
\url{https://huggingface.co/meta-llama/Llama-3.1-8B};
\url{https://huggingface.co/OpenMOSS-Team/Llama-Scope}.}
The earlier public-group qualification uses sentence materials from
the Sparse Feature Circuits repository \citep{marks2025sparse}, whose MIT
notice is retained in the research archive. We have not established a separate
dataset-specific license for those materials and do not infer rights to
underlying third-party content from the repository notice.

The intended use is to improve interpretation of intervention comparisons.
Small discrepancies after calibration should not be taken as evidence that an
explanation is correct or that a model is safe to deploy. Conversely, a code
discrepancy does not establish that the underlying linguistic behavior is
unreliable. Keeping these distinctions explicit limits the risk of presenting
a measurement control as a safety assurance.

%% file: sections/b_experimental_details.tex
\section{\frank{Appendix: Experimental Protocol and Gemma-9B Details}}
\label{app:experimental}

\frank{This appendix specifies the Gemma-9B assets, code construction,
qualification and calibration rules. The refit and bounded-search procedures
also apply to the other settings, whose materials, weighting and coverage
rules are given in Appendices~\ref{app:2b} and~\ref{app:llama}.
Norm-control implementation is specified in Appendix~\ref{app:norm_controls}.}

\subsection{Model, SAE and Reconstruction Conventions}
\label{app:assets}

\begingroup\raggedright
The model and tokenizer are \nolinkurl{google/gemma-2-9b-it}.
We use raw Gemma Scope parameters from
\nolinkurl{google/gemma-scope-9b-it-res}, file
\nolinkurl{layer_20/width_131k/average_l0_153/params.npz}.
\par\endgroup

\frank{The intervention replaces the output of the zero-indexed transformer block 20.}
Model execution uses an FP16 body and an FP32 output head, with one sentence
per forward pass and no key--value cache.

For the raw SAE parameters, encoding applies JumpReLU
\citep{rajamanoharan2024jumping} to
$u=W_{\mathrm{enc}}^{\mathsf T}h+b_{\mathrm{enc}}$:
$z_j=u_j\mathbf{1}[u_j>\theta_j]$, where $\theta_j$ is the released threshold.
Here $W_{\mathrm{enc}}$ has shape $d\times p$ and $h$ is a column vector.
No decoder-bias subtraction precedes encoding. The decoder in our notation
is $D=W_{\mathrm{dec}}^{\mathsf T}$, and reconstruction is $b_{\mathrm{dec}}+Dz$.
Encoder evaluation uses FP32 parameters and activations. \frank{To evaluate reconstruction
error,} we subtract $b_{\mathrm{dec}}$ from $h$ and decode the sparse
coefficients in FP64. Physical reconstruction casts the decoded vector to
FP32, adds the FP32 bias, then casts to FP16. This sparse arithmetic convention
is also used to qualify the candidate reconstructions. The group patch itself
contains only the donor--recipient difference $D_G(z_{d,G}-z_{r,G})$;
the decoder bias is not removed from the recipient state.

\subsection{Same-Support NNLS Refitting}
\label{app:nnls}

The refit minimizes reconstruction error over the encoder's active support.
For activation $h$, decoder bias $b_{\mathrm{dec}}$ and encoder support $S$,
write $y=h-b_{\mathrm{dec}}$. We solve
\[
 u^\star=\underset{u\in\mathbb{R}^{|S|},\,u\geq0}{\arg\min}
 \|D_Su-y\|_2^2,
\]
then set coefficients outside $S$ to zero. There is no regularization term or
new feature selection. All settings use the same audited refit routine.
\frank{It uses SciPy's nonnegative least-squares solver in FP64, with the library's default limit of
$3|S|$ iterations.} The 2B and Llama environments pin SciPy 1.17.1. Each solve
is repeated on identical inputs and must return identical coefficients.

Numerical acceptance checks the raw solution before discarding coefficients
at or below $10^{-10}$. \frank{Let $q=D_S^{\mathsf T}(D_Su-y)$, half the gradient
of the squared-error objective.} The checks use scaling factors
\begin{align*}
 \gamma&=\max\!\left(1,\max_{j\in S}\|D_j\|_2\|y\|_2\right),\\
 \eta&=\max(1,\|u\|_\infty).
\end{align*}
The normalized primal violation, dual violation and complementarity residual
must satisfy
\begin{align*}
 \|(-u)_+\|_\infty/\eta&\leq10^{-8},\\
 \frank{\|(-q)_+\|_\infty/\gamma}&\frank{\leq10^{-8},}\\
 \frank{\max_j|u_jq_j|/(\gamma\eta)}&\frank{\leq10^{-8},}
\end{align*}
where $(x)_+$ takes the elementwise positive part.
These are floating-point optimality diagnostics. The retained coefficients
must separately satisfy the reconstruction-error bound after truncation.
FP32 coefficient-cast diagnostics are recorded, but patches use the retained
FP64 coefficients with the decoder's FP32 values promoted to FP64 before
execution-precision casting. Nonconvergence or failed numerical checks marks
the refit unavailable; reconstruction-behavior qualification follows only
after geometric acceptance.

\subsection{Feature Group and Candidate Construction}
\label{app:construction}

The group contains feature IDs
$\{31256,49323,97461,101074,101300,101972\}$ at layer 20.
It originated in a development search with 177 proposals on 45 pairs.
We keep that selection history separate from subsequent evaluation on new
words.

\frank{At each position, let $S_d$ and $S_r$ be the donor and recipient
encoder supports, and set $m=|S_d|+|S_r|$. Concatenate their positive encoder
coefficients into $x_E\in\mathbb{R}_{\geq0}^{m}$. The linear map
$L\in\mathbb{R}^{d\times m}$ takes these coefficients to the decoded
donor--recipient group difference in activation dimension $d$.} Starting at $x_E$, we take the negative
gradient of the sum of the two relative reconstruction errors in coordinates
scaled by $s_j=\max(|x_{E,j}|,1)$. We project this direction onto the tangent
of the equal-norm constraint in those scaled coordinates, then normalize and
convert back to obtain \frank{$w\in\mathbb{R}^{m}$} with
$\langle Lw,Lx_E\rangle=0$ and $\|Lw\|=\|Lx_E\|$.
The seed lies on the rational curve
\[
 \frank{x(\tau)=\frac{(1-\tau^2)x_E+2\tau w}{1+\tau^2},
 \qquad 0\leq\tau\leq1.}
\]
This curve preserves the ideal patch norm. Nonnegativity gives quadratic
constraints in \frank{$\tau$}; each reconstruction-error constraint gives a polynomial
of degree at most 4 after clearing the denominator. We compute their first
positive real boundary numerically, capped at 1, and take 99\% of that value.
The error constraints use the internal $1.0000005$ multiplier below.
A full-space check must accept the resulting seed. There is no clipping,
smaller-step retry or alternate tangent. A zero encoder patch or a numerically
degenerate physical tangent retains the literal encoder code. Numerical root
finding here is not a certified isolation of every feasible interval.

The candidate optimization starts from this archived seed. At each
position with a nonzero patch, it jointly optimizes the donor and recipient
coefficients on their respective encoder supports. The local objective
minimizes the projection of the candidate patch onto the encoder patch,
normalized by the encoder patch's squared norm. It constrains candidate and
encoder patch norms to be equal and retains each code's nonnegativity and
reconstruction constraints. We use sequential least-squares programming
(SLSQP), limited to 200 iterations and a 30-second callback time limit.
\frank{Positions with zero patch displacement and seeds that retain the encoder
code receive no optimization restart.} The construction uses geometry, not
the measured linguistic effect, as its objective.

A separate full-space check determines eligibility, rather than the solver's
termination flag. Each squared reconstruction error is bounded by
$1.000001$ times its encoder error. The optimizer uses an internal
$1.0000005$ multiplier as a numerical reserve. The ideal relative norm error
must be at most $10^{-7}$; after casting, realized per-position norm mismatch
must be at most $0.001$. The geometric objective may not worsen relative to
the seed beyond $10^{-10}$. Reconstruction-behavior qualification is checked
after construction and is distinct from these geometric conditions.
This local procedure makes no global-optimality claim.

\subsection{\frank{Materials, Token Alignment and Eligibility}}
\label{app:eligibility}

\frank{Token positions $1,\ldots,T-2$ are patched, with zero-based indexing and
sequence length $T$ including the beginning-of-sequence token. Positions 0
and $T-1$ remain unchanged. We tokenize the five-word prefixes directly,
without a chat template. Each answer must extend its prefix by exactly one
token, with distinct IDs for the two answers and no change to the prefix
tokenization. Paired prefixes must have identical token offsets at all five
word boundaries and identical token suffixes after the subject. Equal total
length alone is insufficient. The recorded token IDs, answer IDs and boundary
offsets are checked again before causal evaluation.}

Both members of a pair must have correct native answers and aligned tokens.
Reconstructing the patched positions must preserve the native top-1 output
and keep forward KL at or below 0.02 on each member. These checks precede
alternative group-effect measurement.

The source inventory contains 280 authored count nouns with regular plurals.
A fixed deterministic ordering allocates them to 70 blocks of 4 nouns: 2 subjects and
2 attractors per block, with no noun shared across blocks or roles. Each
block crosses these nouns with 4 prepositions and 2 attractor numbers to give
32 sentence pairs, balanced across the four original grammatical cases.
The first 8 blocks were reserved for planning; the remaining 62 supply the
1,984-pair source panel. This allocation is reproducible, not random sampling
from English or evidence that block effects are independent.

The main panel is selected from these 1,984 pairs.
\commentedit{Table~\ref{tab:qualification_counts} reports the number of pairs
retained at each qualification stage.}
\noindent\begin{minipage}{\columnwidth}
\centering
\begin{tabular}{lr}
\toprule
Condition & Retained \\
\midrule
Source panel & 1,984 \\
Native/encoder qualification & 337 \\
Encoder/NNLS common mask & 325 \\
\commentedit{Geometry and patch norms} & 320 \\
Seed behavior (of 320) & 290 \\
Candidate behavior (of 320) & 259 \\
All-code common mask & 252 \\
\bottomrule
\end{tabular}
\captionof{table}{\commentedit{Sentence pairs retained during 9B candidate
qualification. Seed and candidate behavior are parallel filters on the same
320 pairs. The final intersection yields 252 pairs and 504 directed inputs.}}
\label{tab:qualification_counts}
\end{minipage}
NNLS denotes a nonnegative least-squares refit on encoder support.
The seed and candidate behavior rows are parallel filters on the same 320
pairs, not consecutive filters. The 252-pair intersection includes encoder,
NNLS, seed, and candidate eligibility. Although the main calibration
comparison uses encoder and candidate patches, it preserves that previously
\frank{defined eligibility subset}. Each selected pair supplies both recipient inputs.
No outcomes are imputed for source pairs \frank{outside this subset}.

\subsection{Bounded Per-Input Calibration}
\label{app:search}

All KL values use natural logarithms (nats). Intervention KL measures a
patch's output change and is distinct from the reconstruction KL used in
qualification. Calibration scales the archived, realized displacement rather than decoding
the coefficients again. If $x$ is the archived patched activation and $h$ the
native activation, the new activation is
$\operatorname{FP16}(h+\alpha(x-h))$, with the subtraction and scaling computed
in FP64. The same scalar applies to every patched position of that input.

Calibration evaluates $\alpha=0,0.25,0.5,0.75,1,1.25,1.5,1.75,2$, reusing \commentedit{evaluations
of the native state and unscaled intervention} at 0 and 1. If the \commentedit{unscaled intervention}
already meets tolerance, the search stops there. A target at or below
$10^{-8}$ is marked unmatched. The matching tolerance is
$\max(10^{-6},0.01K_i^E(1))$, with the appropriate conditional-KL target for
Q2.

If no grid point matches, the search selects the adjacent sign-changing
bracket closest to 1, breaking ties by the lower endpoint, and bisects it
at most 12 times. The algorithm does not assume global monotonicity or try
additional brackets after failure. Among matching observed points, it selects
the coefficient closest to 1, then the smallest KL error, then the smaller
coefficient. On failure it retains the smallest-error observed point, breaking
ties by distance from 1 and then by coefficient. These final attempts remain
unsuccessful matches even when their effects improve.

For Q2, the two contextual answer IDs are verified
as distinct single-token continuations before execution. The conditional
distribution is computed directly from logits on the remaining vocabulary.
Excluded logits do not enter its normalization or a probability-mass weight.
Search range, tolerances, and refinement budget are unchanged.

\subsection{Global Coefficient Selection}
\label{app:global}

The fixed split has 79 calibration pairs and 173 evaluation pairs, yielding
158 and 346 inputs. Calibration and evaluation words are disjoint, but
the full panel's earlier outcomes were already exposed.
For every coefficient on the same 9-point grid, the selector averages
candidate KL within each original grammatical case and then across the 4
cases. It minimizes the absolute difference from the analogously averaged
encoder KL. It does not minimize the mean absolute per-input KL difference.

Scores within $10^{-12}$ nats of the minimum are ties, resolved by distance
from 1 and then by the smaller coefficient. No interpolation or refinement
is used. Linguistic effects and evaluation rows are excluded from the
selection input. The selected coefficient, 1.25, is unchanged under all 17
calibration-block deletions; these checks do not choose an ensemble or an
alternative coefficient. All 346 evaluation inputs are retained.
The per-input diagnostic on those inputs includes its 11 failed
matching attempts and is not a same-budget competitor.

\subsection{Aggregation and Conditional Uncertainty}
\label{app:uncertainty}

\frank{The estimator below applies to the case-balanced Gemma-9B summaries.
Case/block-balanced bootstrap comparisons are specified separately for the
NNLS controls (Appendix~\ref{app:control_inference}), Gemma-2B
(Appendix~\ref{app:2b_analysis}) and Llama
(Appendices~\ref{app:llama_inference} and~\ref{app:confirmation}).
Gender has an additional interval-availability rule
(Appendix~\ref{app:gender_materials}). These analyses retain their own
cohorts and weighting rules.}

Generalizing beyond sampled words requires accounting for lexical-item
variation \citep{clark1973language}. Work on confirmatory language experiments
also emphasizes modeling dependence associated with sampled units
\citep{barr2013random}. We use the lexical blocks defined above, not a fitted
crossed random-effects model; these references do not validate independence
between our blocks.

For row statistic $y_i$, case $c$, and $n_c$ rows in that case, define
$\bar y_c=n_c^{-1}\sum_{i:c(i)=c}y_i$ and
$\hat\mu=\frac14\sum_c\bar y_c$. The case assignment is inherited from the
original pair, including when donor and recipient roles are reversed.
For block $b$, the score is
\[
 u_b=\frac14\sum_c\frac{1}{n_c}
       \sum_{i:b(i)=b,\,c(i)=c}(y_i-\bar y_c).
\]
With $B$ represented blocks, the reported standard error is
$\sqrt{\frac{B}{B-1}\sum_bu_b^2}$.
The two-sided conditional interval uses a Student-$t$ critical value with
degrees of freedom equal to the minimum number of supported blocks in any
case minus 1. Missing cases make the estimate undefined. Numerically zero
variance is not treated as a stability certificate.

Let $n_{bc}$ be the row count in block $b$ and case $c$. The balanced block
weight is $w_b=\frac14\sum_c n_{bc}/n_c$ and the effective-block diagnostic is
$1/\sum_bw_b^2$. Duplicating recipient inputs does not create additional
independent lexical information.
Intervals assume independent block-level fluctuations and sufficiently stable
case counts under block sampling for the ratio-based approximation above.
Pre-experiment stress tests show that rare large block-level fluctuations or
correlations across blocks can invalidate coverage. Accordingly, the primary
interpretation is the finite authored
panel; block deletion and conditional intervals do not establish generalization
to English sentences.

For the 9B candidate panel, the prespecified reporting requirements are
at least 80\% input retention in each case, at least 20 effective matched
blocks, at least 20 supported blocks per case, and maximum balanced block
weight at most 0.10. These are reporting floors, not a generalizability theorem.
The planned 0.25-logit interval half-width is a precision goal, not an
equivalence margin. The global-transfer design retains its earlier planning
half-width of 0.437 under assumed block SD 1, exceeding the same 0.25 goal.

\subsection{Numerical Replay}
\label{app:replay}

Independent replay reproduces saved reference and final intervention outputs
for the lexical and Q2 comparisons, including failed matching attempts.
The original lexical replay does not regenerate intermediate search outputs;
the later shared-coefficient evaluation reproduces its measured outputs and
selections.

\subsection{Calibration Cost}
\label{app:compute}

A bounded per-input search can examine nine grid coefficients and at most
12 additional bisection points for each code and KL objective. Early matches
terminate the search, and existing measurements can be reused. These are
coefficient evaluations, not a count of all model calls: native and encoder
references, qualification checks and repeat forwards add work. A transferred
global coefficient requires no coefficient search on the evaluation input.

The per-input search requires the encoder reference and alternative outputs
on the evaluated input. A shared coefficient avoids that search at evaluation
time, but requires a separate calibration set. These different information
and computational requirements preclude a cost-matched method comparison.

\frank{The measured execution environment and reproduction cost of the new
Llama confirmation chain are reported in Appendix~\ref{app:confirmation_compute}.}

\input{sections/a_identification_details}

%% file: sections/a_identification_details.tex
\subsection{Scope of Identification Claims}
\label{app:identification}

Let $\codeset$ contain the eligible donor and recipient code collections for
fixed inputs, decoder, and patch rule. Let $e(z)$ denote the resulting signed
effect defined in \frank{Section~\ref{sec:protocol}}. Point identification on $\codeset$
means that $e(z)$ is constant for all $z\in\codeset$.
Two eligible codes with different effects refute constancy. A finite search
that finds no such pair cannot establish it: untested codes may have different
effects. The smallest observed effect is an upper bound on the infimum over
$\codeset$, and the largest is a lower bound on its supremum. The observed
range therefore need not cover the full range allowed by \frank{the stated code constraints}.
This distinction follows from the definition, without an additional
identification theorem.

Optimizing a coefficient is also different from optimizing the effect of the
nonlinear model suffix. Coefficient extrema alone do not give effect extrema,
and separate coordinate optima need not belong to one jointly feasible code.

Rescaling the patch by $\alpha$ produces a separate physical control.
Treating this control as a comparison within $\codeset$ requires complete
codes that realize it and satisfy the original constraints.
Agreement of two such controls also does not show constancy on
the subset of all codes with a prescribed scalar KL.

%% file: sections/i_strength_controls.tex
\section{\frank{Appendix: Norm Controls and Gender Analyses}}
\label{app:scalar_controls}

\frank{The norm-control comparisons support RQ1 and RQ2
(Sections~\ref{sec:single_scale} and~\ref{sec:positionwise}).
We first specify the implementation, shared evaluation subsets and estimators, then
introduce the gender materials and the cohorts used to compare norm and KL
controls. The final KL decomposition supports the answer-excluded analysis
in Section~\ref{sec:excluded}.}

\subsection{\frank{Norm-Control Implementation}}
\label{app:norm_controls}

Total-norm matching uses the Euclidean
norm of the concatenation of all patched-position vectors. Positionwise norm
matching instead uses one ratio at each nonzero position. Each realized patch
must satisfy its own norm tolerance before entering a matched comparison.
The ratios use the archived realized displacement, the patched activation
minus its \commentedit{native activation}. Scaling materializes a new activation using
the same execution-precision convention as the KL controls.

For total-norm matching, we compute the ratio in double precision from the
stored half-precision activations and accept only coefficients in $[0,2]$.
We neither clip an out-of-range ratio nor search for a replacement.
If the \commentedit{alternative patch} has zero norm and the \commentedit{reference patch} has positive norm, the ratio
is undefined. If only the \commentedit{reference patch} has zero norm, the coefficient is zero.
If both have zero norm, we use coefficient one and flag the control as
uninformative. After adding the scaled displacement to the native activation
and rounding to half precision, the realized norm must have relative error
at most $0.001$ from the \commentedit{reference patch norm}. A zero target requires an exactly zero
realized displacement. Nonfinite realizations and tolerance failures do not
enter matched comparisons, but remain accounted for among attempts.
Positionwise matching applies the same realized-norm tolerance separately at
every position, including the exact-zero rule. The $[0,2]$ bound above applies
to the total-norm coefficient, not to the positionwise ratios.

\subsection{\frank{Common-Input Comparisons and Weighting}}
\label{app:control_inference}

\frank{Table~\ref{tab:practical} reports inputs matched by both total-norm matching and Q1.
Table~\ref{tab:positionwise} instead retains inputs matched by both positionwise and total-norm
controls within each cohort.} A coefficient exists for all
570 attempted total-norm controls across the five earlier cohorts,
but \commentedit{inclusion in a paired comparison} also
depends on the other control. In particular, Table~\ref{tab:practical} retains 165 of 166
9B NNLS inputs. Table~\ref{tab:positionwise} retains 150 for which positionwise measurements
are available, and 109 of 112 Llama-32K inputs pass positionwise matching.
No Q2 measurement exists for 9B NNLS.

The earlier 9B NNLS cohort comprises 166 inputs from a frozen 83-pair
panel, separate from the 504-input candidate panel. The counts of 165 and
150 above apply \frank{different control-specific matching requirements} to that NNLS panel.
The common-code comparison in Appendix~\ref{app:earlier} instead retains
70 of those 83 pairs (140 inputs) for which \commentedit{both NNLS and candidate
interventions match under Q1}. These subsets are not interchangeable.

The paired bootstrap resamples lexical blocks and recomputes the difference
using the same sampled inputs for both controls, with 10,000 draws and
seed 20260925 for the five earlier cohorts. The prospective cohort uses
seed 20260926 (Appendix~\ref{app:confirmation}). Case/block-balanced estimates
give each retained case equal weight and each retained block within that case
equal weight. Intervals assume independent blocks and are pointwise conditional
intervals, without simultaneous coverage or adjustment for the adaptive research
history. The earlier 9B vocabulary structure does not support a block interval;
its pooled estimate is accompanied by case and word-deletion diagnostics.
For the gender positionwise comparison, too few resampled datasets retain the
required cells, so its interval is unavailable.

\begin{table*}[t]
\centering
\input{tables/strength_display_a}
\caption{\textbf{Coefficient-count matched NNLS controls.}
Q1 leaves smaller mean effect gaps on inputs matched by both controls.
Differences are total norm minus Q1 (logits). Balanced means equally weight
cases and blocks within case. Pointwise 95\% paired block-bootstrap
intervals assume independent blocks; the 9B interval is unavailable.
Lost/gained transfers are unweighted counts relative to $E$. The prospective
cohort uses new material in the same Llama-128K configuration and is estimated
separately from the earlier panel. These are the full values underlying
Figure~\ref{fig:nnls_controls}a.}
\label{tab:practical}
\end{table*}

\begin{table}[t]
\centering
\input{tables/strength_display_b}
\caption{\textbf{Positionwise minus total-norm effect gap.}
Negative values favor positionwise norm matching. Each row uses its own
common matched inputs; Figure~\ref{fig:nnls_controls}b plots the negative of
these values.
Values are in logits, with case/block balancing except for pooled 9B.
Intervals use the assumptions in Table~\ref{tab:practical}. The gender interval is unavailable
because too few bootstrap draws preserve the required cells.}
\label{tab:positionwise}
\end{table}

\paragraph{\frank{Transfer counts and secondary intervals.}}
\label{app:practical_details}
\frank{On the complete 112-input Llama-32K cohort, $E$, NNLS, attempted
positionwise norm matching and Q1 produce 96, 90, 95 and 97
transfers. The positionwise-control count includes 3 inputs that miss the
norm tolerance, unlike the 109-input common-input row in
Table~\ref{tab:positionwise}.

The case/block-balanced NNLS reductions under Q1 have paired 95\%
block-bootstrap intervals wholly above zero in both Llama configurations
(Table~\ref{tab:supplementary_intervals}). These secondary estimates use
the complete qualified NNLS cohorts, separately from the candidate subsets.}

\subsection{Encoder Effect Scale}
\label{app:effect_scale}

Table~\ref{tab:encoder_scale} reports the median absolute encoder effect on
the five earlier cohorts in Table~\ref{tab:practical}, using their exact
\frank{common-input subsets}. Medians weight inputs
equally and provide descriptive scale, without changing the paired
estimators. We do not interpret a ratio between a balanced mean gap and an
unweighted median as an input-level relative effect. Encoder transfers
supply the reference counts for paired losses and gains.

\begin{table}[t]
\centering
\input{tables/encoder_effect_scale}
\caption{\commentedit{Effect magnitude of the encoder-based reference intervention $E$} on the five earlier total-norm/Q1 common-input cohorts.
Absolute-effect medians are in logits; transfer counts are unweighted.
These background quantities describe the same inputs as the paired
comparison, separately for each cohort.}
\label{tab:encoder_scale}
\end{table}

\subsection{\frank{Gender Materials and Qualification}}
\label{app:gender_materials}

The gender setting uses explicit male/female cues and the reflexive
continuations \emph{himself} and \emph{herself}
with the predicates \emph{admire} and \emph{hurt}.
The template is \emph{The [male/female] [occupation] beside the [object] could
[hurt/admire]}, so the cue is separated from the final prompt token by
intervening patched content. The model and dictionary are the same as the
2B subject--verb configuration, but the group and materials are separate.
Development and reserved validation each contain 16 lexical blocks, with
disjoint occupation and object words across splits. Each block fixes those
two words and gives one male/female pair for each predicate, yielding
32 pairs and 64 inputs per split. Cases cross recipient cue with predicate.
Both material lists are fixed before tokenizer inspection.

The native and reconstruction checks retain 26 development pairs.
Group discovery evaluates all active singleton contrasts using the
case/block-balanced utility $U(G)$ defined in \frank{Appendix~\ref{app:group_utility}}.
The shortlist combines the top 16 features in each case and fills to 64 by
overall utility, breaking ties by feature ID. One greedy path extends groups
up to twelve features, ranking extensions by efficacy qualification, covered
cases, minimum case utility, overall utility and feature IDs, in that order.
Among all evaluated qualifying groups, final selection maximizes overall
utility, with ties favoring fewer features and then feature IDs.
The selected IDs are 7358, 28833, 31252, 31284, 31595, 55304,
59302, 62256, 76680, 80491, 86279 and 123681.

Before donor and group evaluation, each split must retain at least 12
bilaterally qualified pairs across eight blocks, including four pairs per
predicate. At aligned interior positions, the native-body donor control
replaces recipient activations with donor activations. The decoded-body
donor control replaces the recipient's full SAE reconstruction with the
donor's, retaining the recipient's reconstruction residual. Both leave the
beginning-of-sequence and final prompt positions unchanged.
These donor controls and the encoder-group intervention must each transfer
on at least half of the qualified inputs
and in at least eight blocks, with a transfer and positive mean effect in
every case. The selected group is fixed before validation model evaluations.
\frank{Encoder-group transfer succeeds on all 42 qualified validation inputs;
inputs that fail reconstruction fidelity remain excluded. NNLS retains all
42 inputs, candidate qualification retains 40, and positionwise norm matching
\commentedit{succeeds} on 24 of the 42 NNLS inputs. These filters define the distinct
comparisons in Appendix~\ref{app:gender_comparisons}.}

For gender calibration, the prespecified coverage rule requires at least
80\% of each original case, twelve matched blocks overall and four matched
blocks per case. These availability thresholds differ from those of the
larger subject--verb panels. Balanced estimates average within case--block
cells, then give equal weight to represented blocks within each case and to
the four cases. The paired bootstrap resamples whole lexical blocks
10,000 times with seed 20260925 and reports a conditional 95\% percentile interval only if at
least 99\% of draws contain every case. Missing-case draws are not replaced.
The 24-input positionwise comparison fails this interval-availability rule,
so its means remain descriptive.

\subsection{\frank{Gender Comparisons and Positional Allocation}}
\label{app:gender_comparisons}

\frank{The Q2 comparison in Table~\ref{tab:gender_criteria} retains the 40
inputs qualified for both NNLS and the candidate. The codes therefore use
identical inputs but have different matching coverage; the candidate row
includes failed attempts. The complete 42-input NNLS analysis below and the
24-input positionwise analysis answer separate comparisons.}

\begin{table*}[t]
\centering
\input{tables/gender_criteria}
\caption{\textbf{Answer exclusion is a boundary, not a uniform robustness result.}
Q2 worsens NNLS effect agreement on the gender common-input cohort despite
complete matching. \commentedit{Mean} candidate gaps decrease across all attempts, with failed
coverage. Means and reductions are case/block-balanced logits; final means
retain unmatched attempts. Intervals are pointwise conditional 95\%
block-bootstrap intervals. A negative reduction denotes deterioration.}
\label{tab:gender_criteria}
\end{table*}

\frank{On the complete 42-input NNLS cohort, the pooled original gap is 0.082 logit,
compared with residual gaps of 0.054 under Q1 and 0.185 under Q2. The balanced
Q2 reduction has a paired 95\% block-bootstrap interval wholly below zero
(Table~\ref{tab:supplementary_intervals}). Thus the deterioration is present on the complete cohort
as well as the common-40 comparison. Positionwise matching \commentedit{succeeds on} only
24 of the 42 inputs, which is why its comparison is reported separately.}

\frank{On the 24 common gender inputs, the pooled NNLS gap is 0.089 before control,
0.081 under total-norm matching, 0.014 under positionwise norm matching,
0.061 under Q1 and 0.183 under Q2. The corresponding balanced means for
positionwise, Q1 and total-norm matching are \GenderCommonPosition{},
\GenderCommonQone{} and \GenderCommonTotal{}, corresponding to the points in
Figure~\ref{fig:control_boundaries}a. Both summaries describe the same 24-input subset,
with different weighting.}

\frank{On this same subset, the original NNLS patch has mean cosine similarity near 0.998 to
the encoder patch on these 24 inputs. Q1 and Q2 retain that orientation.
For realized displacements $v_{i,t}^{a}$ and $v_{i,t}^{E}$ at patched
positions $t$, let $v_i^E$ concatenate the encoder displacements.
For $\|v_i^E\|_2>0$, define the normalized positional norm-profile distance as
\[
d_i^{\mathrm{profile}}(a)=
\frac{\sqrt{\sum_t(\|v_{i,t}^{a}\|_2-\|v_{i,t}^{E}\|_2)^2}}
     {\|v_i^E\|_2}.
\]
Normalization uses
\commentedit{the reference patch's total norm}, so the distance is sensitive to overall scale
as well as allocation across positions. Its pooled mean is about 0.045
before control, near zero under positionwise matching and larger under
Q1 and Q2.
Orientation and allocation summaries are diagnostics of the measured patches,
not an intervention that isolates the cause of the effect difference.}

\subsection{Partitioning the Output KL}
\label{app:kl_partition}

Let $p$ denote the native distribution and $q$ an intervened distribution.
Partition the vocabulary into answer tokens $A$ and the remainder $R$.
Write $p_A=\sum_{v\in A}p(v)$ and $q_A=\sum_{v\in A}q(v)$ for their answer
masses; $p(\cdot\mid A)$ and $q(\cdot\mid A)$ are the normalized answer
distributions, with analogous definitions for $R$.
\frank{These conditional distributions require $0<p_A<1$ and $0<q_A<1$.}
Let $K_A=D_{\mathrm{KL}}(p(\cdot\mid A)\Vert q(\cdot\mid A))$,
$K_R=D_{\mathrm{KL}}(p(\cdot\mid R)\Vert q(\cdot\mid R))$, and let $K_M$ be
the KL between the Bernoulli distributions with probabilities $p_A$ and $q_A$.
The chain rule gives
\[
D_{\mathrm{KL}}(p\Vert q)=K_M+p_AK_A+(1-p_A)K_R.
\]
Q2 targets $K_R$ before multiplication by $1-p_A$.
It neither fixes the answer mass nor the conditional distribution within the
answer tokens. The numerical decomposition reproduces full KL to floating-point
precision. On gender NNLS inputs, Q2 reduces \commentedit{the discrepancy from $E$
in the weighted remaining-vocabulary term $(1-p_A)K_R$} and increases
\commentedit{it in the weighted within-answer term $p_AK_A$}. This identity locates
the output differences; it is not evidence of a causal mediation mechanism.

%% file: tables/strength_display_a.tex
\begin{tabular}{@{}lrlrrrlrr@{}}
\toprule
 & & & \multicolumn{2}{c}{Mean gap (logit)} & \multicolumn{2}{c}{Paired difference (logit)} & \multicolumn{2}{c}{Lost/gained} \\
Cohort & $n$ & Weighting & Total norm & Q1 & Estimate & 95\% interval & Total norm & Q1 \\
\midrule
\multicolumn{9}{l}{Earlier panels (follow-up controls)} \\
9B SVA & 165 & pooled & 0.056 & 0.015 & $0.041$ & unavailable & 1/2 & 1/0 \\
2B SVA & 62 & balanced & 0.033 & 0.008 & $0.024$ & $[0.011, 0.042]$ & 0/0 & 0/0 \\
Llama-128K & 188 & balanced & 0.160 & 0.017 & $0.142$ & $[0.117, 0.168]$ & 10/1 & 0/1 \\
Llama-32K & 112 & balanced & 0.127 & 0.012 & $0.115$ & $[0.077, 0.159]$ & 3/1 & 0/1 \\
2B gender & 42 & balanced & 0.075 & 0.041 & $0.033$ & $[0.007, 0.065]$ & 0/0 & 0/0 \\
\midrule
\multicolumn{9}{l}{Prospective confirmation (new material)} \\
Llama-128K & 304 & balanced & 0.181 & 0.021 & $0.160$ & $[0.131, 0.191]$ & 11/1 & 0/0 \\
\bottomrule
\end{tabular}

%% file: tables/strength_display_b.tex
\begin{tabular}{@{}lrr@{\enspace}l@{}}
\toprule
Cohort & $n$ & \frank{Estimate} & 95\% interval \\
\midrule
9B SVA & 150 & $-0.024$ & unavailable \\
2B SVA & 62 & $0.001$ & $[-0.004, 0.006]$ \\
Llama-128K & 188 & $-0.048$ & $[-0.067, -0.030]$ \\
Llama-32K & 109 & $-0.029$ & $[-0.047, -0.012]$ \\
2B gender & 24 & $-0.073$ & unavailable \\
\midrule
\multicolumn{4}{l}{Prospective confirmation} \\
Llama-128K & 304 & $-0.046$ & $[-0.065, -0.028]$ \\
\bottomrule
\end{tabular}

%% file: tables/encoder_effect_scale.tex
\begin{tabular}{lrrr}
\toprule
Cohort & $n$ & \shortstack{Median\\$|e_i(E)|$} & \shortstack{Encoder\\transfers} \\
\midrule
9B SVA & 165 & 3.92 & 85 \\
2B SVA & 62 & 4.61 & 55 \\
Llama-128K & 188 & 3.33 & 126 \\
Llama-32K & 112 & 4.05 & 96 \\
2B gender & 42 & 4.47 & 42 \\
\bottomrule
\end{tabular}

%% file: tables/gender_criteria.tex
\begin{tabular}{lrrrrl}
\toprule
Code & Inputs & Q2 matches & Before & Final & \frank{Reduction and 95\% interval} \\
\midrule
NNLS & 40 & 40 & 0.089 & 0.139 & $-0.050$ $[-0.091, -0.008]$ \\
Candidate & 40 & 13 & 0.754 & 0.401 & $0.353$ $[0.265, 0.447]$ \\
\bottomrule
\end{tabular}

%% file: sections/c_additional_results.tex
\section{\frank{Appendix: Additional Calibration Results and Sensitivity Analyses}}
\label{app:results}

\frank{This appendix gives the complete calibration estimates and the
diagnostics supporting the candidate stress tests in
Section~\ref{sec:candidate}. The matched-input and all-attempt summaries
retain their specified populations. We then examine patch geometry,
unmatched inputs and the later search-range continuation before detailing
the Q2 comparison and the low-strength random-group control.}

\subsection{Complete Calibration Summaries}
\label{app:candidate_summary}

\frank{Table~\ref{tab:supplementary_intervals} reports the primary and
supporting estimates with each row's specified population and interval
method. Table~\ref{tab:lexical} gives the balanced Gemma gap means,
distinct from the pooled descriptions in Figure~\ref{fig:core_calibration}a.
The earlier Llama matching and
all-attempt accounting for both selectors are reported with their
configuration-specific analysis in Appendix~\ref{app:llama_inference}
(Table~\ref{tab:llama_calibration}).}

\begin{table*}[t]
\centering
\small
\input{tables/supplementary_intervals}
\caption{\textbf{Primary and supporting calibration estimates with pointwise
95\% intervals.} All quantities are in logits. A reduction is the
original gap minus the selected gap, so a negative reduction denotes
deterioration. The 9B rows use case-balanced means and block-based
Student-$t$ intervals. Other rows use case/block-balanced means and lexical-block
bootstrap intervals. Candidate Q1 rows use matched inputs, including the
prespecified 9B primary reduction. The 9B Q2-minus-Q1 row retains all attempts,
and the gender row uses the complete 42-input NNLS cohort, distinct from the
common-40 comparison in Table~\ref{tab:gender_criteria}. Each row preserves its
own input set and inferential assumptions.}
\label{tab:supplementary_intervals}
\end{table*}

\begin{table}[t]
\centering
\begin{tabular}{lrrr}
\toprule
\multicolumn{4}{c}{Balanced mean absolute gap (logit)} \\
\midrule
Panel & $n$ & Before & After \\
\midrule
9B: all attempts & 504 & 0.816 & 0.076 \\
9B: matched & 478 & 0.752 & 0.028 \\
2B: all, matched & 64 & 0.561 & 0.015 \\
\bottomrule
\end{tabular}
\caption{Candidate absolute effect gaps from \commentedit{the encoder-based reference intervention $E$}, in logits, before and after
Q1. Means use each setting's specified balancing;
$n$ counts inputs. The 9B all-attempt row includes 26 failed matches.
The 9B matched row uses the inputs for the prespecified primary comparison,
whose estimand is the paired reduction. All 2B attempts match.}
\label{tab:lexical}
\end{table}

\subsection{Earlier Positionwise-Norm-Matched Comparisons}
\label{app:earlier}

The earlier 12-pair example has 10 strict encoder transfers and 1 candidate
transfer in 24 inputs before Q1. Q1 calibration of the candidate patch
restores identical input-level decisions and 10 transfers;
the absolute effect gap falls from 1.936 to 0.041 logit.
The original equal-positionwise-norm counterexample remains valid under its
\frank{stated code constraints}. It does not show a disagreement that survives this functional
control.

On a separate exposed panel, 70 of 83 pairs have a common matched comparison.
On the same 140 inputs, the NNLS mean absolute effect gap from $E$
falls from 0.126 to 0.012 logit under Q1, and transfer
disagreements fall from 7 to 1. The constructed candidate's gap falls from
0.617 to 0.032, with disagreements falling from 16 to 1. Unlike the candidate,
the unscaled NNLS patch is not constrained to have the encoder's norm.
The singular--singular case retains only 13 of 18 pairs, below its
required 15, so the full case-coverage condition fails. Neither this subset
nor the earlier example is pooled with the 9B candidate panel.

\subsection{Candidate Patch Geometry}
\label{app:geometry}

To characterize how much the candidate changes patch direction, we concatenate
the realized displacements across token positions into vectors $v_i^E$ and
$v_i^{\mathrm{candidate}}$. Each displacement is the stored patched state minus the same native
recipient state, after execution-precision materialization. We compute
$\|v_i^{\mathrm{candidate}}-v_i^E\|_2/\|v_i^E\|_2$ and their cosine for each input, then
take unweighted quantiles across inputs. The unchanged boundary positions
contribute zero; none of the 504 encoder displacement vectors has zero norm.

\frank{Across all 504 inputs in the 9B candidate panel, relative distance has
median 0.576 and an interquartile range from 0.490 to 0.661.
Cosine similarity has median 0.834 and an interquartile range from
0.782 to 0.880.} On the 478 Q1 matches, the corresponding medians are
0.568 and 0.839. These are geometric descriptions of the constructed patches,
not evidence that direction is irrelevant to effects or that the patches
represent naturally occurring inference choices.

\subsection{Influence and Unmatched Inputs}
\label{app:lexical_details}

Table~\ref{tab:lexical_cases} separates matching coverage from the effect
comparison. Within each case, both gap means use exactly the same matched
inputs. The main estimate gives these four cases equal weight, despite
their different retained counts. The matched subset spans 49 lexical blocks,
with an effective count of approximately 31 under this weighting
(Appendix~\ref{app:uncertainty}). Table~\ref{tab:supplementary_intervals}
reports uncertainty for both the primary reduction and the secondary matched
\commentedit{residual gap}. The narrower \commentedit{gap} interval does not repair the missed precision
target for the primary estimate or bound each input's gap.

\begin{table}[t]
\centering
\begin{tabular}{lrrr}
\toprule
Subject--attractor & Matched & Before & After \\
\midrule
Plural--plural & 118/124 & 0.643 & 0.020 \\
Plural--singular & 117/124 & 0.702 & 0.036 \\
Singular--plural & 130/136 & 0.740 & 0.025 \\
Singular--singular & 113/120 & 0.923 & 0.033 \\
\bottomrule
\end{tabular}
\caption{Matching coverage and mean absolute effect gaps (logit) before and
after Q1, by original subject and attractor number. Counts
are matched/attempted inputs. Both means exclude unmatched inputs;
the case label is retained when donor and recipient are reversed.}
\label{tab:lexical_cases}
\end{table}

For the 9B candidate panel, every single-block deletion leaves mean paired
reduction between 0.604 and 0.777 logit. Restricting to the 232 pairs matched
in both donor--recipient orders gives 0.701 logit. These analyses change the evaluation
subset and supplement, rather than replace, the \frank{primary matched-input analysis}.
The 26 unmatched inputs retain a larger case-balanced original gap of
1.975 logit (1.986 with equal weight per input).
Their best attempted coefficients are 2 for 21 inputs, 1.5 for 3, and
1.25 and 1.75 for 1 each. These are not all measurements at the search cap.
At those selected attempts, the pooled mean absolute gap falls from 1.986
to 0.946 logit, and candidate transfers increase from 6 to 9, compared with
15 for \commentedit{$E$}. These descriptive quantities use equal weight per
input, unlike the case-balanced main estimates. None of the 26 attempts
meets the KL tolerance; the partial improvement is not a successful calibration.

\paragraph{\frank{Complete-record accounting.}}
\frank{The archived lexical analysis retains all 504 inputs, including the
26 unsuccessful matching attempts, and evaluates deletions of all 50 blocks
represented in the attempted panel. A block with no matched input does
not contribute to the matched estimate. The answer-excluded analysis retains
the same 504 inputs with separate matching flags for the two calibration
rules. Global calibration retains its 346 evaluation inputs and reports
deletions of the 33 represented evaluation blocks. These records distinguish
attempted observations from those included in each reported estimate; the
summaries above do not replace that accounting. Appendix~\ref{app:artifact}
describes the CPU checks for these scalar comparisons and the patch and
model-replay checks that still depend on internal records.}

\subsection{\frank{Candidate Search-Range Sensitivity}}
\label{app:range}

\frank{The original Q1 and Q2 searches evaluate $\alpha\in\{0,0.25,\dots,2\}$ and
refine a sign-crossing bracket at most 12 times, with tolerance
$\max(10^{-6},0.01\,K_i^E(1))$ for Q1, using the corresponding conditional-KL
target for Q2. The retained Q1 traces contain 26 unmatched 9B
searches, 32 in Llama-128K, 31 in Llama-32K and 24 in gender. Retained Q2
traces used for this continuation contain 33, 30 and 27 searches in the latter
three configurations. The continuation did not include 9B Q2, and 2B
subject--verb searches all matched.
At every observed scale in these failed searches, the selector-specific KL is
below its encoder target. In \RangeRisingAtCap{} of the \RangeSearches{}
search records, the best attempt lies at $\alpha=2$ and KL increases over the
last grid segment. That finite difference does not establish monotonicity or
reachability beyond the grid. The 9B Q2 scalar traces were subsequently
recovered from the original run records and are included in the CPU supplement.}

\frank{We specified a continuation after inspecting these traces and froze it before
any new forward pass. It continues only the originally unmatched searches in
the seven setting--criterion cells, evaluates the eight new points
$\alpha\in\{2.25,2.5,\dots,4\}$, and applies the original tolerance,
tie rules and refinement budget within the new interval. Original matched
selections, coverage decisions and the prespecified 9B endpoint are unchanged.
A complete replay on a second host reproduces every new output exactly.
Table~\ref{tab:range} reports the resulting matched subsets and pooled gaps without
intervals.}

\begin{table*}[t]
\centering
\setlength{\tabcolsep}{5pt}
\input{tables/range_sensitivity}
\caption{\frank{\textbf{Post hoc search-range sensitivity for unmatched candidate
searches.} Matched counts use the original denominators; $^\dagger$ marks
matched subsets that do not meet the original per-case coverage rule. Gaps are pooled
mean absolute effect gaps from \commentedit{the encoder-based reference intervention $E$} (logits). All-attempt gaps retain each
unmatched search at its best attempt. The matched-gap columns compare inputs
matched within $\alpha\le2$ with inputs first matched by the continuation.
Lost counts are encoder transfers lost over the full cohort. Search records
overlap across Q1 and Q2 and are not independent inputs.}}
\label{tab:range}
\end{table*}

\frank{The continuation matches \RangeRescued{} of the \RangeSearches{} records. The
extended matched subsets would satisfy the original coverage rule in both Llama-128K
cells and in Llama-32K under Q2, but not in Llama-32K under Q1 or in either
gender cell. The only case with no original match, in gender, becomes fully
matched under both criteria. Rescued searches match at median coefficients of
\RangeAlphaMin{}--\RangeAlphaMax{}; because candidates share the encoder's
norm at each position, these patches exceed twice the encoder's magnitude and
may leave the usual activation range.}

\frank{Rescued inputs retain larger matched gaps than inputs matched within the
original range in every cell except gender under Q2, where the two are close.
The all-attempt gap falls in six cells. In gender under Q2 it rises slightly
when inputs are pooled but falls from \RangeGenderQtwoBalancedBefore{} to
\RangeGenderQtwoBalancedAfter{} logit under the original case/block
balancing, so we report neither an overall improvement nor a deterioration.
Transfer losses fall except in Llama-32K. Searches that remain unmatched
describe this second finite range; they do not show that no coefficient could
match.}

\subsection{Answer-Excluded Comparison}
\label{app:answer_excluded}

Q2 and Q1 each match 478 of 504 inputs.
Of these, 477 match under both rules; 25 fail under both, and each rule has
1 failure that the other matches. The main comparison retains all 504
inputs. On the secondary common matched subset, the case-balanced mean
absolute gap is 0.754 logit before calibration, 0.036 after Q2 and 0.028 after Q1. This subset comparison does
not replace the all-attempt estimates.

Across all 504 inputs, Q2 leaves a larger mean \commentedit{residual gap} than Q1,
with a positive paired difference under the block-based interval analysis
(Table~\ref{tab:supplementary_intervals}). The two procedures therefore
should not be described as giving identical effects. Q2 loses 12 encoder
transfers and gains 4, compared with 11 losses and 3 gains under Q1.
The answer-excluded mean reduction remains positive after every block
deletion, ranging from 0.620 to 0.787 logit, while 57 individual inputs
have larger gaps than before calibration. These checks retain both the
unmatched attempts and the worsening inputs.

\subsection{Random-Group Comparison at a Common Low KL}
\label{app:random}

This comparison uses one previously selected random six-feature group on
17 pairs, giving 34 recipient inputs. We hold the random-group
intervention fixed and attenuate the agreement-group intervention toward its
output KL. The control asks whether matching scalar functional strength
necessarily matches the target effect across these different feature groups.

The bounded search matches 25 inputs. Mean KL is 0.0010766 for the
attenuated agreement group and 0.0010759 for the random group.
Mean signed target effects are 0.730 and 0.098 logit, respectively.
The paired difference is positive on 24 of the 25 inputs, but both
groups have zero strict transfers at this low strength.
Only 9 of 17 pairs match in both inputs, and full-panel coverage fails.
All 9 unmatched attempts are retained. The comparison supports a
low-strength, matched-subset distinction, not original-strength specificity,
equivalence of transfer rates, or a distributional claim about random groups.

%% file: tables/supplementary_intervals.tex
\begin{tabular}{lllr rrr}
\toprule
Configuration & Code & Quantity & $n$ & Estimate & Lower & Upper \\
\midrule
9B SVA & Candidate & Q1 reduction & 478 & 0.7239 & 0.3241 & 1.1236 \\
9B SVA & Candidate & \commentedit{Q1 gap} & 478 & 0.0284 & 0.0223 & 0.0344 \\
9B SVA & Candidate & \commentedit{Q2 gap $-$ Q1 gap} & 504 & 0.0071 & 0.0042 & 0.0100 \\
2B SVA & Candidate & Q1 reduction & 64 & 0.5459 & 0.3637 & 0.7485 \\
2B SVA & Candidate & \commentedit{Q1 gap} & 64 & 0.0149 & 0.0106 & 0.0201 \\
2B SVA & Candidate & Q2 reduction & 64 & 0.5410 & 0.3590 & 0.7446 \\
2B SVA & Candidate & \commentedit{Q2 gap} & 64 & 0.0198 & 0.0140 & 0.0265 \\
Llama-128K & NNLS & Q1 reduction & 188 & 0.1492 & 0.1173 & 0.1828 \\
Llama-128K & Candidate & Q1 reduction & 92 & 1.2279 & 1.0149 & 1.4256 \\
Llama-32K & NNLS & Q1 reduction & 112 & 0.1519 & 0.1137 & 0.1943 \\
Llama-32K & Candidate & Q1 reduction & 75 & 1.2602 & 0.9626 & 1.5655 \\
2B gender & NNLS & Q2 reduction & 42 & -0.0556 & -0.0944 & -0.0129 \\
\bottomrule
\end{tabular}

%% file: tables/range_sensitivity.tex
\begin{tabular}{@{}llrrrrrrrr@{}}
\toprule
 & & \multicolumn{2}{c}{Matched inputs} & \multicolumn{2}{c}{All-attempt gap} & \multicolumn{2}{c}{Matched gap} & Rescued & Lost \\
Setting & Sel. & $\alpha{\le}2$ & $\alpha{\le}4$ & $\alpha{\le}2$ & $\alpha{\le}4$ & original & rescued & median $\alpha$ & $\alpha{\le}2{\to}4$ \\
\midrule
9B SVA & Q1 & 478/504 & 498/504 & 0.076 & 0.047 & 0.028 & 0.392 & 3.03 & 11$\to$6 \\
Llama-128K & Q1 & 92/124$^\dagger$ & 116/124 & 0.270 & 0.193 & 0.130 & 0.241 & 2.53 & 13$\to$7 \\
Llama-128K & Q2 & 91/124$^\dagger$ & 117/124 & 0.302 & 0.234 & 0.174 & 0.274 & 2.39 & 14$\to$9 \\
Llama-32K & Q1 & 75/106$^\dagger$ & 92/106$^\dagger$ & 0.197 & 0.144 & 0.041 & 0.087 & 2.38 & 10$\to$10 \\
Llama-32K & Q2 & 76/106$^\dagger$ & 94/106 & 0.205 & 0.173 & 0.051 & 0.112 & 2.47 & 10$\to$10 \\
2B gender & Q1 & 16/40$^\dagger$ & 34/40$^\dagger$ & 0.306 & 0.243 & 0.161 & 0.287 & 2.33 & 7$\to$3 \\
2B gender & Q2 & 13/40$^\dagger$ & 29/40$^\dagger$ & 0.414 & 0.420 & 0.396 & 0.417 & 2.55 & 8$\to$4 \\
\bottomrule
\end{tabular}

%% file: sections/g_two_setting_details.tex
\section{\frank{Appendix: Gemma-2B Details and Cross-Setting Response Analyses}}
\label{app:2b}

\frank{The first two subsections specify the Gemma-2B group, eligible inputs
and statistical analysis. The response analyses then compare the Gemma-9B
and Gemma-2B settings across answer-excluded KL budgets, shared coefficients
and finite-grid hindsight optima. These follow-ups retain their own evaluation
panels and selection rules; they do not extend the original lexical
comparison to every input or coefficient.}

\subsection{Assets, Group Selection and Eligibility}
\label{app:group_utility}

\begingroup\raggedright
The 2B subject--verb setting uses \nolinkurl{google/gemma-2-2b}
and the residual-stream SAE from \nolinkurl{google/gemma-scope-2b-pt-res}.
Its asset is \nolinkurl{layer_12/width_131k/average_l0_129/params.npz}.
\par\endgroup
The fixed group contains features 5153, 9422, 11917, 19857, 46415 and 89045.
This is distinct from the earlier public two-feature candidate that failed
qualification (Appendix~\ref{app:secondmodel}).

Development uses 16 pairs built from 8 subject--attractor lemma pairs, with
the fixed preposition \emph{near}. Each lexical block contains one subject
and one attractor, crossed over their two grammatical numbers: two sentence
pairs and four recipient inputs. Twelve pairs qualify for discovery.
For an encoder intervention on proposed group $G$, define its development
utility as
\[
 U(G)=\sum_i w_i\,\operatorname{clip}\!\left(
 \frac{m_i^0-m_i(E;G)}{m_i^0},-1,2\right),
\]
where $m_i(E;G)$ is the \commentedit{margin under the reference intervention for group $G$} and $w_i$ gives equal weight to
cases, blocks within case, and inputs within each case--block cell.
\frank{Clipping bounds the normalized margin decrease between $-1$ and $2$.} All native margins in this
calculation are positive.

Discovery evaluates every feature with a nonzero donor--recipient coefficient
contrast on a qualified input: 1,449 singleton features. It retains the top
64 by $U$, breaking ties by feature ID, and greedily adds one feature at a
time from that list. An extension must improve utility by more than $10^{-6}$;
the search stops at six features. The run evaluates 305 joint proposals.
The development group must transfer in at least 4 blocks, with at least one
transfer and a positive mean margin decrease in every case.

The selected group is fixed before evaluation on 32 new subject--attractor
lemma pairs, disjoint from the development vocabulary. Lemmas are used in
only one block and one role within this validation panel; the construction
again crosses both numbers with \emph{near}. Case labels refer to recipient
subject and attractor number. Validation requires group transfers on at least
half of the qualified inputs, spread over at least 8 blocks, with a
transfer and positive mean decrease in every case. The native-body and
decoded-body donor controls defined in Appendix~\ref{app:gender_materials}
must also satisfy the breadth conditions. These
are selection and validation rules for this group, not guarantees for an
arbitrary group. Encoder-group effects on the validation material were known
before the alternative-code comparison was specified.

The source panel contains 64 pairs, 128 inputs and 32 lexical blocks;
120 inputs pass token alignment. Bilateral encoder and candidate checks
retain 32 pairs, or 64 inputs in 23 represented blocks. Candidate case
counts are 11, 21, 11 and 21 in recipient subject--attractor order:
singular--singular, singular--plural, plural--singular and plural--plural.
NNLS retains 31 pairs, 62 inputs and 22 blocks: one input in the
additional pair has reconstruction KL 0.02498, above 0.02, so both inputs
are excluded. Its case counts are 10, 21, 10 and 21 in the same order.

The construction uses the same feasible-seed and geometry-objective procedure
as the 9B comparison, with 200 iterations and a 30-second local limit.
Candidate norm qualification is checked on the materialized physical patch,
not only the optimizer's coefficients. Computation uses FP16 eager model
execution, an FP32 output head, disabled TF32 and one mathematical thread.
Interventions exclude the beginning-of-sequence and final prompt positions.

\input{figures/budget_response}
\input{figures/global_distribution}
\input{figures/global_fullrange}

\subsection{Weighting and Inference}
\label{app:2b_analysis}

For 2B continuous summaries, we first average inputs within each
case--block cell, then average represented blocks within case, then average
the four cases. Counts remain unweighted. Reported 95\% bootstrap intervals
resample lexical blocks 10,000 times with seed 20260921 and use percentile
endpoints. All retained draws contain every case. These intervals condition
on the eligible blocks and the assumed block-level sampling model.
They do not treat inputs, resampling draws or execution replays as
independent observations.

Both Q1 and Q2 reduce candidate gaps, with paired 95\% block-bootstrap
intervals wholly above zero (Table~\ref{tab:supplementary_intervals}).
Candidate leave-one-block reductions under answer exclusion range from
0.49279 to 0.57707. The 9B planned 0.25 precision threshold was not a
2B decision rule.

Q2 matches all 64 candidate and all 62 eligible
NNLS inputs. The balanced candidate gap falls from 0.561 to 0.020 logit,
with no remaining transfer disagreement; the NNLS gap falls from 0.103
to 0.008. These reductions use the \commentedit{unscaled interventions} as baselines, whereas
Table~\ref{tab:practical} compares total-norm matching with Q1 on common inputs.

The \commentedit{shared-coefficient} comparison leaves out one lexical block at a time. Each fold
selects a coefficient using only other blocks' balanced mean \commentedit{full-output} KL;
held-out effects do not enter selection. All candidate folds choose 1.25
and all NNLS folds choose 1. Overlapping training folds are not independent
replicates, \frank{so we do not estimate uncertainty by treating folds as independent observations.}
The candidate has 40 improvements and 24 deteriorations; its 59 transfers
versus $E$'s 56 comprise four gains and one loss. A higher transfer count
is not identical agreement with the reference.

\subsection{Budget-Specific Coverage and \commentedit{Residual Gaps}}
\label{app:response_details}

Let $K_{i,R}^a(\alpha)$ denote the answer-excluded KL for \commentedit{intervention} $a$ on
input $i$ after scaling its patch by $\alpha$. Each relative budget
$q\in\{0.25,0.5,0.75,1,1.25,1.5\}$ sets a common target
$qK_{i,R}^E(1)$. We independently select coefficients
$\alpha_{i,q}^E$ and $\alpha_{i,q}^a$ for the encoder and alternative patches
using the bounded search in Appendix~\ref{app:search}. Joint matching
requires both patches to reach this target within tolerance.
Writing $e_i(a,\alpha)$ for the effect of the scaled patch, the budget-specific
gap is
\[
 g_{i,q}(a)=\left|e_i(a,\alpha_{i,q}^a)-e_i(E,\alpha_{i,q}^E)\right|.
\]
Thus the reference effect is measured with the encoder patch calibrated to
the same budget, rather than held at its original scale.
The 9B comparison uses the 346-input candidate evaluation panel;
2B uses its 64 candidate and 62 NNLS inputs.

Table~\ref{tab:budgets} records the budget-specific matched means and
counts shown in Figure~\ref{fig:budget_response}. The plotted fixed-intersection
means are reported below, separately from the changing matched subsets.

\begin{table}[t]
\centering
\begin{tabular}{lrrrr}
\toprule
& \multicolumn{2}{c}{9B: pooled} & \multicolumn{2}{c}{2B: balanced} \\
$q$ & Matched & Gap & Matched & Gap \\
\midrule
0.25 & 339/346 & 0.074 & 64/64 & 0.019 \\
0.50 & 343/346 & 0.056 & 64/64 & 0.017 \\
0.75 & 343/346 & 0.043 & 64/64 & 0.019 \\
1.00 & 334/346 & 0.037 & 64/64 & 0.020 \\
1.25 & 303/346 & 0.038 & 56/64 & 0.027 \\
1.50 & 209/346 & 0.042 & 35/64 & 0.028 \\
\bottomrule
\end{tabular}
\caption{Candidate joint matching coverage and absolute effect gaps (logit) at each
answer-excluded KL budget. Means condition on \commentedit{both patches matching}: pooled
inputs in 9B and case/block-balanced in 2B. The 9B panel is the global
evaluation subset, not all 504 lexical inputs. High-budget coverage
failures prevent extending these subset means to the full panels.}
\label{tab:budgets}
\end{table}

\begin{table}[t]
\centering
\begin{tabular}{lrrr}
\toprule
$q$ & NNLS matched & Gap & Candidate \\
\midrule
0.25 & 62/62 & 0.00782 & 0.01927 \\
0.50 & 62/62 & 0.00798 & 0.01679 \\
0.75 & 62/62 & 0.00986 & 0.01902 \\
1.00 & 62/62 & 0.00821 & 0.01977 \\
1.25 & 62/62 & 0.01217 & 0.04455 \\
1.50 & 50/62 & 0.01225 & 0.13646 \\
\bottomrule
\end{tabular}
\caption{Additional 2B budget accounting. NNLS gaps condition on joint
matching. The candidate column retains all 64 inputs, using nearest
measured attempts for failed matches; its larger-budget entries are not
matched comparisons. All gaps are balanced mean absolute differences in logits.}
\label{tab:2b_budgets}
\end{table}

In 2B, candidate coverage at $q=1.25$ \frank{is 16 of 21 inputs in one case}, below its
specified 80\% floor. At $q=1.5$, multiple case floors fail for the candidate,
and NNLS also fails coverage. On the 35 candidate inputs matched at all
budgets, gaps are 0.02286, 0.01908, 0.02207, 0.02169, 0.02485 and 0.02811.
The corresponding NNLS common subset contains 50 inputs, with gaps
0.00825, 0.00725, 0.00957, 0.00823, 0.01165 and 0.01225. At $q=0.5$ each code
loses one encoder transfer; other matched-budget comparisons have no transfer
gains or losses. These finite counts do not certify identical decisions.
Across budget-specific matched subsets, the largest candidate gap is
0.162 logit at $q=1.25$.

The 9B common subset contains 205 inputs; its pooled mean gaps are
0.0798, 0.0591, 0.0458, 0.0393, 0.0381 and 0.0421. These averages conceal tails:
at $q=0.25$, the maximum gap is 1.421 and the 90th percentile is 0.183.
The means are descriptive, with no equivalence bound or confidence interval.
Full-panel coverage deteriorates at high budgets; changing subsets do not
justify an unrestricted shared-response-curve claim.

\subsection{\commentedit{Shared-Coefficient} Calibration: Means and Complete Distributions}
\label{app:global_distributions}

These follow-ups test how much of the per-input agreement a shared
coefficient preserves for constructed candidates in the two Gemma
subject--verb settings. \commentedit{In 9B, one coefficient is shared across all
evaluation inputs. In 2B, each held-out block uses a coefficient selected on
the remaining blocks.} Both protocols were specified after earlier outcomes on their
panels had been inspected. Table~\ref{tab:global} reports balanced means,
and Figure~\ref{fig:global_distribution} shows the input-level distributions.
Figure~\ref{fig:global_fullrange} retains their full measured range.

\begin{table}[t]
\centering
\begin{tabular}{lrrr}
\toprule
\multicolumn{4}{c}{Balanced mean absolute gap (logit)} \\
\midrule
Setting & Before & \commentedit{Shared} & Per-input \\
\midrule
9B ($n=346$) & 0.876 & 0.704 & 0.046 \\
2B ($n=64$) & 0.561 & 0.313 & 0.015 \\
\bottomrule
\end{tabular}
\caption{Balanced candidate effect gaps (logit) on the same inputs within
each setting. Both selectors return $\alpha=1.25$; every 2B fold selects that
value. Per-input Q1 uses each evaluation input's reference
and separate search, including 11 failed attempts in 9B. It is an
information-richer diagnostic, not a same-budget comparator.}
\label{tab:global}
\end{table}

Shared coefficients reduce mean gaps but worsen 216 of 346 inputs in 9B
and 24 of 64 in 2B. A retrospective breakdown in 9B places most deterioration
among inputs whose individual coefficient is below the shared value; those
inputs already start close to $E$. The more concentrated 2B coefficients
do not remove the trade-off under its selector
(selection rules in Appendices~\ref{app:global} and~\ref{app:2b_analysis}).

\frank{For 9B, global calibration gives positive case mean reductions of
0.113--0.236 logit and block-deletion reductions of 0.053--0.206 logit.
These summaries describe the same finite panel and do not establish
independent lexical replication.}

The 9B evaluation mean KL is 0.09685 nats under global calibration and
0.09811 under $E$. \frank{For the 2B candidate, the per-input coefficient median
is 1.156, with an interquartile range from 1.068 to 1.238.}

\subsection{Finite-Grid Hindsight Comparisons}

The nine-point global grid is $0,0.25,\ldots,2$. In 9B, selecting the best
coefficient after observing evaluation effects returns 1.5, with balanced gap
0.571. \frank{It worsens 233 of 346 inputs}, and the pooled median absolute gap rises
from 0.102 to 0.480. Every one-block deletion returns the same grid optimum.
Allowing a separate effect-optimal grid point per input gives 0.109, but
uses the outcomes and is not a deployable KL selector.
In 2B, the single-coefficient hindsight optimum is 1.25 for the candidate and
1 for NNLS. None of these optima addresses coefficients between grid points.

\subsection{Numerical Verification Scope}
\label{app:gemma_replay}

Independent executions reproduce the final outputs of the 2B code comparison
and the complete measured searches of its response extension. Two recomputed
metadata norms differ by one unit in the last place (ULP) at double precision,
but applied patches and model outputs agree exactly.

The 9B budget and shared-coefficient extensions also reproduce their measured
search points and selections. This scope exceeds the final-output-only replay
of the original lexical comparison. Some response-extension records retain
output verification summaries rather than full arrays, so arbitrary new
vocabulary statistics would require model re-execution.


%% file: figures/budget_response.tex
\begin{figure*}[t]
\centering
\begin{tikzpicture}[x=1cm,y=1cm,font=\small]
\newcommand{\budgetpoint}[5]{%
  \pgfmathsetmacro{\bx}{(#1-.25)*4}
  \draw[black!65,fill=white] (\bx,{2.6+30*#3}) circle (3.8pt);
  \fill[blue!60!black] (\bx,{2.6+30*#2}) circle (2pt);
  \ifdim#1pt=.25pt
    \node[anchor=south west] at ({\bx+.07},{2.6+30*max(#2,#3)+.18}) {#4};
  \else
    \node[anchor=south] at (\bx,{2.6+30*max(#2,#3)+.18}) {#4};
  \fi
}
\newcommand{\budgetaxes}[2]{%
  \node[font=\small\bfseries,text height=1.5ex,text depth=.25ex] at (2.5,1.31) {#1};
  \node[anchor=west] at (-.65,5.92) {#2};
  \foreach \g/\lab in {0/0,.03/0.03,.06/0.06,.09/0.09} {
    \draw[black!12] (0,{2.6+30*\g}) -- (5,{2.6+30*\g});
    \node[anchor=east] at (-.12,{2.6+30*\g}) {\lab};
  }
  \draw (0,5.3) -- (0,2.6) -- (5,2.6);
  \node[rotate=90] at (-1.24,3.95) {Mean absolute gap (logit)};
  \foreach \x/\q in {0/0.25,1/0.50,2/0.75,3/1.00,4/1.25,5/1.50} {
    \node[anchor=north] at (\x,2.52) {\q};
    \draw (\x,2.6) -- (\x,2.54);
  }
  \node at (2.5,1.86) {Relative answer-excluded KL budget $q$};
}
\begin{scope}
\budgetaxes{(a) Gemma-9B: pooled means}{346 attempts; 205 inputs match all budgets}
\BudgetNineCommon
\BudgetNinePoints
\end{scope}
\begin{scope}[shift={(7.9,0)}]
\budgetaxes{(b) Gemma-2B: balanced means}{64 attempts; 35 inputs match all budgets}
\BudgetTwoCommon
\BudgetTwoPoints
\node[align=center] at (2.5,4.65) {Per-case coverage fails\\at $q=1.25,1.50$};
\end{scope}
\fill[blue!60!black] (1.25,7.08) circle (2.5pt);
\node[anchor=west] at (1.40,7.08) {Matched at each budget};
\draw[black!65,densely dashed] (7.10,7.08) -- (7.7,7.08);
\draw[black!65,fill=white] (7.4,7.08) circle (3.8pt);
\node[anchor=west] at (7.8,7.08) {Matched at all six budgets};
\end{tikzpicture}
\caption{Candidate \commentedit{residual gaps} across answer-excluded KL budgets.
Small mean gaps persist on both changing and fixed matched subsets.
\commentedit{$E$ is the encoder-based reference intervention. Both reference and candidate patches are rescaled to each budget.}
Filled points use each budget's joint $E$/candidate matches (counts out of
346 in 9B, 64 in 2B). Open rings and dashed lines use the all-six-budget
intersection; superposed symbols indicate overlapping means. Both axes are
shared, but weighting differs: pooled in 9B, case/block-balanced in 2B.
The 2B coverage failures are labeled; 9B counts are descriptive.
Lines connect measured means, without uncertainty bands or an equivalence
claim. Exact values and coverage are in Appendix~\ref{app:response_details}.}
\label{fig:budget_response}
\end{figure*}
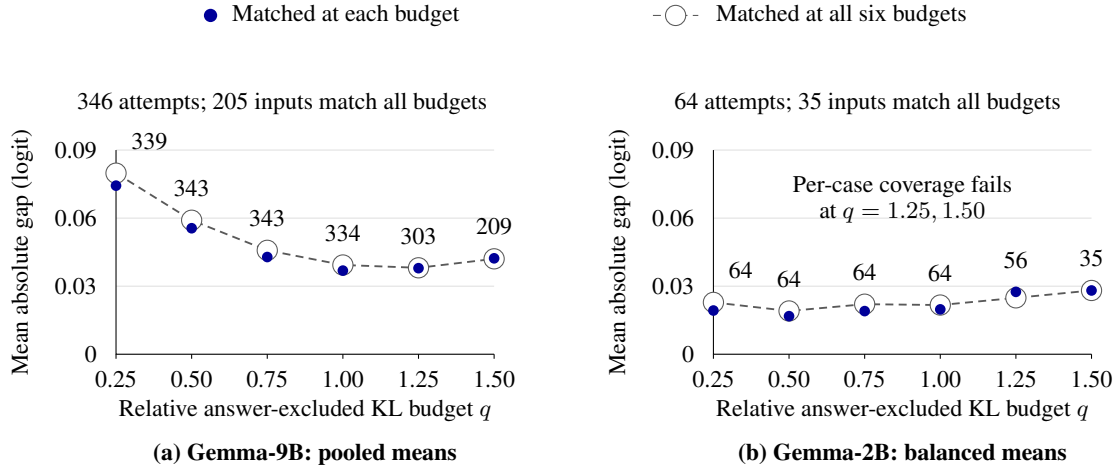

%% file: figures/global_distribution.tex
\begin{figure*}[t]
\centering
\newcommand{\calibrationpanel}[4]{%
\begin{tikzpicture}[x=1.5cm,y=3.0cm,font=\small,
 globalcurve/.style={black,line width=.9pt},
 inputcurve/.style={blue!60!black,densely dashed,line width=.9pt}]
\fill[black!6] (-1,0) rectangle (0,1);
\foreach \y/\label in {0/0,.25/25,.5/50,.75/75,1/100} {
  \draw[black!15] (-1,\y) -- (3,\y);
  \node[anchor=east] at (-1.08,\y) {\label};
}
\draw[dotted] (0,0) -- (0,1);
\begin{scope}
\clip (-1,0) rectangle (3,1.01);
\input{#2}
\end{scope}
\draw (-1,0) -- (3,0);
\draw (-1,0) -- (-1,1);
\foreach \x in {-1,0,1,2,3} {
  \draw (\x,0) -- (\x,-.02);
  \node[anchor=north] at (\x,-.025) {\x};
}
\node[rotate=90] at (-1.70,.5) {Cumulative share (\%)};
\node at (1,-.22) {Reduction in absolute effect gap (logit)};
\node[font=\small\bfseries,text height=1.5ex,text depth=.25ex] at (1,-.40) {#1};
\node[anchor=south] at (1,1.035) {\commentedit{Shared} worsens #3/#4 inputs};
\draw[globalcurve] (.85,.42) -- (1.15,.42);
\node[anchor=west] at (1.23,.42) {\commentedit{Shared}};
\draw[inputcurve] (.85,.18) -- (1.15,.18);
\node[anchor=west,align=left] at (1.23,.18) {Q1\\(own-input KL)};
\draw[white,fill=black,line width=.7pt] (0,{#3/#4}) circle (3pt);
\end{tikzpicture}%
}
\begin{minipage}{.49\textwidth}\centering
\begin{tikzpicture}[x=1.5cm,y=3.0cm,font=\small,
 globalcurve/.style={black,line width=.9pt},
 inputcurve/.style={blue!60!black,densely dashed,line width=.9pt}]
\fill[black!6] (-1,0) rectangle (0,1);
\foreach \y/\label in {0/0,.25/25,.5/50,.75/75,1/100} {
  \draw[black!15] (-1,\y) -- (3,\y);
  \node[anchor=east] at (-1.08,\y) {\label};
}
\draw[dotted] (0,0) -- (0,1);
\begin{scope}
\clip (-1,0) rectangle (3,1.01);
\input{figures/global_ecdf_paths}
\end{scope}
\draw (-1,0) -- (3,0);
\draw (-1,0) -- (-1,1);
\foreach \x in {-1,0,1,2,3} {
  \draw (\x,0) -- (\x,-.02);
  \node[anchor=north] at (\x,-.025) {\x};
}
\node[rotate=90] at (-1.70,.5) {Cumulative share (\%)};
\node at (1,-.22) {Reduction in absolute effect gap (logit)};
\node[font=\small\bfseries,text height=1.5ex,text depth=.25ex] at (1,-.40) {(a) 9B: word-split selection};
\node[anchor=south] at (1,1.035) {\commentedit{Shared} worsens \GlobalNineWorsened/\GlobalNineInputs inputs};
\draw[globalcurve] (.85,.42) -- (1.15,.42);
\node[anchor=west] at (1.23,.42) {\commentedit{Shared}};
\draw[inputcurve] (.85,.18) -- (1.15,.18);
\node[anchor=west,align=left] at (1.23,.18) {Q1\\(own-input KL)};
\draw[white,fill=black,line width=.7pt] (0,{\GlobalNineWorsened/\GlobalNineInputs}) circle (3pt);
\end{tikzpicture}%

\end{minipage}\hfill
\begin{minipage}{.49\textwidth}\centering
\begin{tikzpicture}[x=1.5cm,y=3.0cm,font=\small,
 globalcurve/.style={black,line width=.9pt},
 inputcurve/.style={blue!60!black,densely dashed,line width=.9pt}]
\fill[black!6] (-1,0) rectangle (0,1);
\foreach \y/\label in {0/0,.25/25,.5/50,.75/75,1/100} {
  \draw[black!15] (-1,\y) -- (3,\y);
  \node[anchor=east] at (-1.08,\y) {\label};
}
\draw[dotted] (0,0) -- (0,1);
\begin{scope}
\clip (-1,0) rectangle (3,1.01);
\input{figures/global_ecdf_2b_paths}
\end{scope}
\draw (-1,0) -- (3,0);
\draw (-1,0) -- (-1,1);
\foreach \x in {-1,0,1,2,3} {
  \draw (\x,0) -- (\x,-.02);
  \node[anchor=north] at (\x,-.025) {\x};
}
\node[rotate=90] at (-1.70,.5) {Cumulative share (\%)};
\node at (1,-.22) {Reduction in absolute effect gap (logit)};
\node[font=\small\bfseries,text height=1.5ex,text depth=.25ex] at (1,-.40) {(b) 2B: block-held-out selection};
\node[anchor=south] at (1,1.035) {\commentedit{Shared} worsens \GlobalTwoWorsened/\GlobalTwoInputs inputs};
\draw[globalcurve] (.85,.42) -- (1.15,.42);
\node[anchor=west] at (1.23,.42) {\commentedit{Shared}};
\draw[inputcurve] (.85,.18) -- (1.15,.18);
\node[anchor=west,align=left] at (1.23,.18) {Q1\\(own-input KL)};
\draw[white,fill=black,line width=.7pt] (0,{\GlobalTwoWorsened/\GlobalTwoInputs}) circle (3pt);
\end{tikzpicture}%

\end{minipage}
\caption{Candidate effect-gap reductions under \commentedit{shared-coefficient} and per-input calibration.
\commentedit{Gaps are measured from the encoder-based reference intervention $E$.}
\commentedit{Shared} coefficients improve means while worsening many individual comparisons.
Negative reductions denote larger gaps. Dots at zero mark the \commentedit{shared-coefficient}
worsening fraction. The curves weight inputs equally, retaining 11 failed
per-input Q1 attempts in 9B. Table~\ref{tab:global} reports balanced means. Beyond the displayed
3-logit limit lie 2 \commentedit{shared-coefficient} and 40 per-input values in 9B, none in 2B
(full curves: Appendix~\ref{app:global_distributions},
Figure~\ref{fig:global_fullrange}). Selection protocols differ
between panels. \commentedit{Shared-coefficient} selection holds out evaluation inputs, which were
exposed in earlier analyses. Per-input Q1 uses each evaluation input's outputs.
Curves are descriptive, without confidence intervals.}
\label{fig:global_distribution}
\end{figure*}
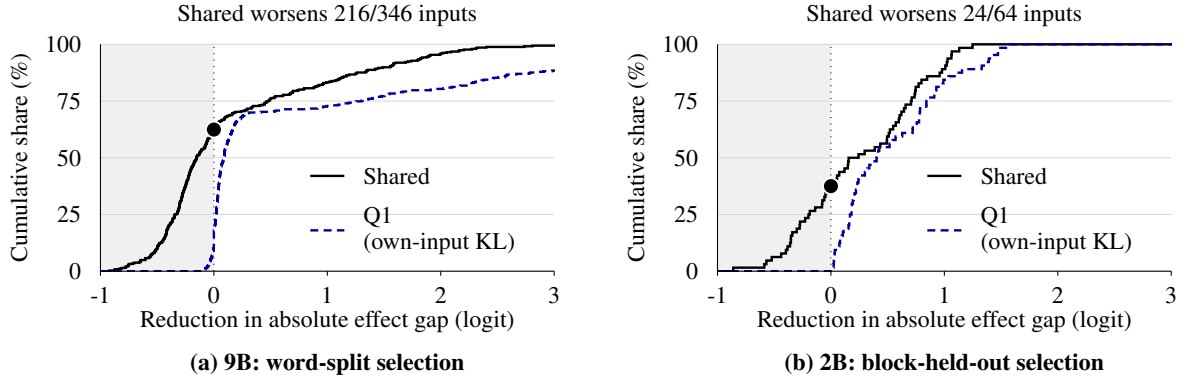

%% file: figures/global_ecdf_paths.tex
\draw[globalcurve] plot coordinates {
(-1,0)
(-0.921697617,0.000000000)
(-0.921697617,0.002890173)
(-0.876040459,0.002890173)
(-0.876040459,0.005780347)
(-0.871612549,0.005780347)
(-0.871612549,0.008670520)
(-0.830803871,0.008670520)
(-0.830803871,0.011560694)
(-0.811252594,0.011560694)
(-0.811252594,0.014450867)
(-0.779499054,0.014450867)
(-0.779499054,0.017341040)
(-0.779273033,0.017341040)
(-0.779273033,0.020231214)
(-0.761912346,0.020231214)
(-0.761912346,0.023121387)
(-0.754117966,0.023121387)
(-0.754117966,0.026011561)
(-0.751956940,0.026011561)
(-0.751956940,0.028901734)
(-0.748444557,0.028901734)
(-0.748444557,0.031791908)
(-0.716489792,0.031791908)
(-0.716489792,0.034682081)
(-0.702882767,0.034682081)
(-0.702882767,0.037572254)
(-0.658994675,0.037572254)
(-0.658994675,0.040462428)
(-0.648430824,0.040462428)
(-0.648430824,0.043352601)
(-0.646542549,0.043352601)
(-0.646542549,0.046242775)
(-0.644898415,0.046242775)
(-0.644898415,0.049132948)
(-0.607496262,0.049132948)
(-0.607496262,0.052023121)
(-0.602975845,0.052023121)
(-0.602975845,0.054913295)
(-0.602465630,0.054913295)
(-0.602465630,0.057803468)
(-0.573493958,0.057803468)
(-0.573493958,0.060693642)
(-0.564509392,0.060693642)
(-0.564509392,0.063583815)
(-0.560060501,0.063583815)
(-0.560060501,0.066473988)
(-0.559425354,0.066473988)
(-0.559425354,0.069364162)
(-0.552597046,0.069364162)
(-0.552597046,0.072254335)
(-0.552467346,0.072254335)
(-0.552467346,0.075144509)
(-0.540815353,0.075144509)
(-0.540815353,0.078034682)
(-0.535293579,0.078034682)
(-0.535293579,0.080924855)
(-0.522898674,0.080924855)
(-0.522898674,0.083815029)
(-0.522866249,0.083815029)
(-0.522866249,0.086705202)
(-0.515070915,0.086705202)
(-0.515070915,0.089595376)
(-0.514256477,0.089595376)
(-0.514256477,0.092485549)
(-0.509862900,0.092485549)
(-0.509862900,0.095375723)
(-0.509327888,0.095375723)
(-0.509327888,0.098265896)
(-0.499870300,0.098265896)
(-0.499870300,0.101156069)
(-0.498447418,0.101156069)
(-0.498447418,0.104046243)
(-0.498027802,0.104046243)
(-0.498027802,0.106936416)
(-0.495676994,0.106936416)
(-0.495676994,0.109826590)
(-0.493281364,0.109826590)
(-0.493281364,0.112716763)
(-0.486561775,0.112716763)
(-0.486561775,0.115606936)
(-0.485524178,0.115606936)
(-0.485524178,0.118497110)
(-0.482274055,0.118497110)
(-0.482274055,0.121387283)
(-0.475710869,0.121387283)
(-0.475710869,0.124277457)
(-0.465665817,0.124277457)
(-0.465665817,0.127167630)
(-0.457242966,0.127167630)
(-0.457242966,0.130057803)
(-0.454710007,0.130057803)
(-0.454710007,0.132947977)
(-0.449229240,0.132947977)
(-0.449229240,0.135838150)
(-0.446031570,0.135838150)
(-0.446031570,0.138728324)
(-0.442016602,0.138728324)
(-0.442016602,0.141618497)
(-0.438523293,0.141618497)
(-0.438523293,0.144508671)
(-0.435147285,0.144508671)
(-0.435147285,0.147398844)
(-0.432906151,0.147398844)
(-0.432906151,0.150289017)
(-0.429461479,0.150289017)
(-0.429461479,0.153179191)
(-0.428216934,0.153179191)
(-0.428216934,0.156069364)
(-0.428064346,0.156069364)
(-0.428064346,0.158959538)
(-0.426345825,0.158959538)
(-0.426345825,0.161849711)
(-0.417707443,0.161849711)
(-0.417707443,0.164739884)
(-0.416407585,0.164739884)
(-0.416407585,0.167630058)
(-0.416193008,0.167630058)
(-0.416193008,0.170520231)
(-0.415925026,0.170520231)
(-0.415925026,0.173410405)
(-0.415729523,0.173410405)
(-0.415729523,0.176300578)
(-0.415233612,0.176300578)
(-0.415233612,0.179190751)
(-0.413433075,0.179190751)
(-0.413433075,0.182080925)
(-0.413100243,0.182080925)
(-0.413100243,0.184971098)
(-0.408112526,0.184971098)
(-0.408112526,0.187861272)
(-0.405100822,0.187861272)
(-0.405100822,0.190751445)
(-0.404241562,0.190751445)
(-0.404241562,0.193641618)
(-0.402569771,0.193641618)
(-0.402569771,0.196531792)
(-0.401338577,0.196531792)
(-0.401338577,0.199421965)
(-0.395509720,0.199421965)
(-0.395509720,0.202312139)
(-0.392255783,0.202312139)
(-0.392255783,0.205202312)
(-0.384598732,0.205202312)
(-0.384598732,0.208092486)
(-0.383907318,0.208092486)
(-0.383907318,0.210982659)
(-0.382470131,0.210982659)
(-0.382470131,0.213872832)
(-0.374760628,0.213872832)
(-0.374760628,0.216763006)
(-0.367554665,0.216763006)
(-0.367554665,0.219653179)
(-0.366669655,0.219653179)
(-0.366669655,0.222543353)
(-0.366399765,0.222543353)
(-0.366399765,0.225433526)
(-0.363548279,0.225433526)
(-0.363548279,0.228323699)
(-0.361033440,0.228323699)
(-0.361033440,0.231213873)
(-0.359697342,0.231213873)
(-0.359697342,0.234104046)
(-0.348663330,0.234104046)
(-0.348663330,0.236994220)
(-0.348574638,0.236994220)
(-0.348574638,0.239884393)
(-0.346912384,0.239884393)
(-0.346912384,0.242774566)
(-0.345941544,0.242774566)
(-0.345941544,0.245664740)
(-0.345380783,0.245664740)
(-0.345380783,0.248554913)
(-0.340892792,0.248554913)
(-0.340892792,0.251445087)
(-0.324113846,0.251445087)
(-0.324113846,0.254335260)
(-0.323609352,0.254335260)
(-0.323609352,0.257225434)
(-0.321969986,0.257225434)
(-0.321969986,0.260115607)
(-0.321726799,0.260115607)
(-0.321726799,0.263005780)
(-0.315438271,0.263005780)
(-0.315438271,0.265895954)
(-0.315038681,0.265895954)
(-0.315038681,0.268786127)
(-0.312290192,0.268786127)
(-0.312290192,0.271676301)
(-0.310267448,0.271676301)
(-0.310267448,0.274566474)
(-0.308104515,0.274566474)
(-0.308104515,0.277456647)
(-0.307603836,0.277456647)
(-0.307603836,0.280346821)
(-0.305766106,0.280346821)
(-0.305766106,0.283236994)
(-0.305351257,0.283236994)
(-0.305351257,0.286127168)
(-0.303412437,0.286127168)
(-0.303412437,0.289017341)
(-0.303064346,0.289017341)
(-0.303064346,0.291907514)
(-0.303043365,0.291907514)
(-0.303043365,0.294797688)
(-0.302972794,0.294797688)
(-0.302972794,0.297687861)
(-0.298986435,0.297687861)
(-0.298986435,0.300578035)
(-0.298591614,0.300578035)
(-0.298591614,0.303468208)
(-0.298493385,0.303468208)
(-0.298493385,0.306358382)
(-0.292904854,0.306358382)
(-0.292904854,0.309248555)
(-0.289285660,0.309248555)
(-0.289285660,0.312138728)
(-0.288133621,0.312138728)
(-0.288133621,0.315028902)
(-0.285640717,0.315028902)
(-0.285640717,0.317919075)
(-0.285335541,0.317919075)
(-0.285335541,0.320809249)
(-0.283708572,0.320809249)
(-0.283708572,0.323699422)
(-0.281812668,0.323699422)
(-0.281812668,0.326589595)
(-0.281230927,0.326589595)
(-0.281230927,0.329479769)
(-0.277777672,0.329479769)
(-0.277777672,0.332369942)
(-0.277097702,0.332369942)
(-0.277097702,0.335260116)
(-0.276494026,0.335260116)
(-0.276494026,0.338150289)
(-0.276448250,0.338150289)
(-0.276448250,0.341040462)
(-0.276329041,0.341040462)
(-0.276329041,0.343930636)
(-0.271755219,0.343930636)
(-0.271755219,0.346820809)
(-0.270928383,0.346820809)
(-0.270928383,0.349710983)
(-0.267055511,0.349710983)
(-0.267055511,0.352601156)
(-0.264657974,0.352601156)
(-0.264657974,0.355491329)
(-0.264315605,0.355491329)
(-0.264315605,0.358381503)
(-0.261323929,0.358381503)
(-0.261323929,0.361271676)
(-0.261112213,0.361271676)
(-0.261112213,0.364161850)
(-0.259478569,0.364161850)
(-0.259478569,0.367052023)
(-0.256216049,0.367052023)
(-0.256216049,0.369942197)
(-0.252365112,0.369942197)
(-0.252365112,0.372832370)
(-0.249584198,0.372832370)
(-0.249584198,0.375722543)
(-0.246877670,0.375722543)
(-0.246877670,0.378612717)
(-0.246495247,0.378612717)
(-0.246495247,0.381502890)
(-0.246035576,0.381502890)
(-0.246035576,0.384393064)
(-0.245235443,0.384393064)
(-0.245235443,0.387283237)
(-0.241374969,0.387283237)
(-0.241374969,0.390173410)
(-0.241304874,0.390173410)
(-0.241304874,0.393063584)
(-0.237860680,0.393063584)
(-0.237860680,0.395953757)
(-0.237083435,0.395953757)
(-0.237083435,0.398843931)
(-0.236013412,0.398843931)
(-0.236013412,0.401734104)
(-0.228679657,0.401734104)
(-0.228679657,0.404624277)
(-0.228276253,0.404624277)
(-0.228276253,0.407514451)
(-0.227176666,0.407514451)
(-0.227176666,0.410404624)
(-0.225460052,0.410404624)
(-0.225460052,0.413294798)
(-0.223587036,0.413294798)
(-0.223587036,0.416184971)
(-0.222888947,0.416184971)
(-0.222888947,0.419075145)
(-0.220705986,0.419075145)
(-0.220705986,0.421965318)
(-0.219232559,0.421965318)
(-0.219232559,0.424855491)
(-0.217552185,0.424855491)
(-0.217552185,0.427745665)
(-0.216546059,0.427745665)
(-0.216546059,0.430635838)
(-0.212764740,0.430635838)
(-0.212764740,0.433526012)
(-0.211300850,0.433526012)
(-0.211300850,0.436416185)
(-0.209987640,0.436416185)
(-0.209987640,0.439306358)
(-0.209644318,0.439306358)
(-0.209644318,0.442196532)
(-0.207283974,0.442196532)
(-0.207283974,0.445086705)
(-0.203545570,0.445086705)
(-0.203545570,0.447976879)
(-0.201404572,0.447976879)
(-0.201404572,0.450867052)
(-0.198425293,0.450867052)
(-0.198425293,0.453757225)
(-0.198219299,0.453757225)
(-0.198219299,0.456647399)
(-0.194248199,0.456647399)
(-0.194248199,0.459537572)
(-0.187953949,0.459537572)
(-0.187953949,0.462427746)
(-0.184656143,0.462427746)
(-0.184656143,0.465317919)
(-0.184140205,0.465317919)
(-0.184140205,0.468208092)
(-0.180498123,0.468208092)
(-0.180498123,0.471098266)
(-0.178127289,0.471098266)
(-0.178127289,0.473988439)
(-0.175459862,0.473988439)
(-0.175459862,0.476878613)
(-0.174259186,0.476878613)
(-0.174259186,0.479768786)
(-0.172586441,0.479768786)
(-0.172586441,0.482658960)
(-0.168443680,0.482658960)
(-0.168443680,0.485549133)
(-0.161853790,0.485549133)
(-0.161853790,0.488439306)
(-0.157595634,0.488439306)
(-0.157595634,0.491329480)
(-0.157295227,0.491329480)
(-0.157295227,0.494219653)
(-0.155895233,0.494219653)
(-0.155895233,0.497109827)
(-0.154024124,0.497109827)
(-0.154024124,0.500000000)
(-0.153310776,0.500000000)
(-0.153310776,0.502890173)
(-0.152868271,0.502890173)
(-0.152868271,0.505780347)
(-0.146203041,0.505780347)
(-0.146203041,0.508670520)
(-0.145656586,0.508670520)
(-0.145656586,0.511560694)
(-0.131012917,0.511560694)
(-0.131012917,0.514450867)
(-0.130600929,0.514450867)
(-0.130600929,0.517341040)
(-0.127526283,0.517341040)
(-0.127526283,0.520231214)
(-0.125220299,0.520231214)
(-0.125220299,0.523121387)
(-0.121762276,0.523121387)
(-0.121762276,0.526011561)
(-0.119831085,0.526011561)
(-0.119831085,0.528901734)
(-0.111995697,0.528901734)
(-0.111995697,0.531791908)
(-0.111320496,0.531791908)
(-0.111320496,0.534682081)
(-0.105574608,0.534682081)
(-0.105574608,0.537572254)
(-0.091126442,0.537572254)
(-0.091126442,0.540462428)
(-0.084359169,0.540462428)
(-0.084359169,0.543352601)
(-0.079088211,0.543352601)
(-0.079088211,0.546242775)
(-0.079008102,0.546242775)
(-0.079008102,0.549132948)
(-0.078322411,0.549132948)
(-0.078322411,0.552023121)
(-0.071575165,0.552023121)
(-0.071575165,0.554913295)
(-0.070875168,0.554913295)
(-0.070875168,0.557803468)
(-0.070692062,0.557803468)
(-0.070692062,0.560693642)
(-0.066237450,0.560693642)
(-0.066237450,0.563583815)
(-0.063291550,0.563583815)
(-0.063291550,0.566473988)
(-0.062157631,0.566473988)
(-0.062157631,0.569364162)
(-0.061696053,0.569364162)
(-0.061696053,0.572254335)
(-0.058944702,0.572254335)
(-0.058944702,0.575144509)
(-0.058536530,0.575144509)
(-0.058536530,0.578034682)
(-0.056322098,0.578034682)
(-0.056322098,0.580924855)
(-0.052096367,0.580924855)
(-0.052096367,0.583815029)
(-0.049073219,0.583815029)
(-0.049073219,0.586705202)
(-0.034590721,0.586705202)
(-0.034590721,0.589595376)
(-0.033253670,0.589595376)
(-0.033253670,0.592485549)
(-0.028285027,0.592485549)
(-0.028285027,0.595375723)
(-0.025573730,0.595375723)
(-0.025573730,0.598265896)
(-0.020833969,0.598265896)
(-0.020833969,0.601156069)
(-0.020428658,0.601156069)
(-0.020428658,0.604046243)
(-0.019931793,0.604046243)
(-0.019931793,0.606936416)
(-0.018461227,0.606936416)
(-0.018461227,0.609826590)
(-0.018085480,0.609826590)
(-0.018085480,0.612716763)
(-0.014788628,0.612716763)
(-0.014788628,0.615606936)
(-0.013855934,0.615606936)
(-0.013855934,0.618497110)
(-0.010731697,0.618497110)
(-0.010731697,0.621387283)
(-0.003083229,0.621387283)
(-0.003083229,0.624277457)
(0.000323296,0.624277457)
(0.000323296,0.627167630)
(0.001665115,0.627167630)
(0.001665115,0.630057803)
(0.002117157,0.630057803)
(0.002117157,0.632947977)
(0.007539749,0.632947977)
(0.007539749,0.635838150)
(0.011287689,0.635838150)
(0.011287689,0.638728324)
(0.019765854,0.638728324)
(0.019765854,0.641618497)
(0.021918297,0.641618497)
(0.021918297,0.644508671)
(0.023512840,0.644508671)
(0.023512840,0.647398844)
(0.024512291,0.647398844)
(0.024512291,0.650289017)
(0.027198792,0.650289017)
(0.027198792,0.653179191)
(0.034635544,0.653179191)
(0.034635544,0.656069364)
(0.044414520,0.656069364)
(0.044414520,0.658959538)
(0.053852081,0.658959538)
(0.053852081,0.661849711)
(0.066573143,0.661849711)
(0.066573143,0.664739884)
(0.072826385,0.664739884)
(0.072826385,0.667630058)
(0.090333939,0.667630058)
(0.090333939,0.670520231)
(0.093681335,0.670520231)
(0.093681335,0.673410405)
(0.095726013,0.673410405)
(0.095726013,0.676300578)
(0.099332809,0.676300578)
(0.099332809,0.679190751)
(0.110420227,0.679190751)
(0.110420227,0.682080925)
(0.120834351,0.682080925)
(0.120834351,0.684971098)
(0.136110306,0.684971098)
(0.136110306,0.687861272)
(0.158171654,0.687861272)
(0.158171654,0.690751445)
(0.162189484,0.690751445)
(0.162189484,0.693641618)
(0.163934708,0.693641618)
(0.163934708,0.696531792)
(0.176882744,0.696531792)
(0.176882744,0.699421965)
(0.191590309,0.699421965)
(0.191590309,0.702312139)
(0.238702774,0.702312139)
(0.238702774,0.705202312)
(0.265020370,0.705202312)
(0.265020370,0.708092486)
(0.282681465,0.708092486)
(0.282681465,0.710982659)
(0.291701317,0.710982659)
(0.291701317,0.713872832)
(0.296968460,0.713872832)
(0.296968460,0.716763006)
(0.330034256,0.716763006)
(0.330034256,0.719653179)
(0.331898689,0.719653179)
(0.331898689,0.722543353)
(0.341801643,0.722543353)
(0.341801643,0.725433526)
(0.345854759,0.725433526)
(0.345854759,0.728323699)
(0.402714729,0.728323699)
(0.402714729,0.731213873)
(0.426353455,0.731213873)
(0.426353455,0.734104046)
(0.445158005,0.734104046)
(0.445158005,0.736994220)
(0.447686195,0.736994220)
(0.447686195,0.739884393)
(0.450215340,0.739884393)
(0.450215340,0.742774566)
(0.461974144,0.742774566)
(0.461974144,0.745664740)
(0.465442657,0.745664740)
(0.465442657,0.748554913)
(0.481079102,0.748554913)
(0.481079102,0.751445087)
(0.488229752,0.751445087)
(0.488229752,0.754335260)
(0.493735313,0.754335260)
(0.493735313,0.757225434)
(0.496958733,0.757225434)
(0.496958733,0.760115607)
(0.503817558,0.760115607)
(0.503817558,0.763005780)
(0.530211449,0.763005780)
(0.530211449,0.765895954)
(0.538161278,0.765895954)
(0.538161278,0.768786127)
(0.544902802,0.768786127)
(0.544902802,0.771676301)
(0.606380463,0.771676301)
(0.606380463,0.774566474)
(0.613012314,0.774566474)
(0.613012314,0.777456647)
(0.663827896,0.777456647)
(0.663827896,0.780346821)
(0.673636436,0.780346821)
(0.673636436,0.783236994)
(0.677024841,0.783236994)
(0.677024841,0.786127168)
(0.693037033,0.786127168)
(0.693037033,0.789017341)
(0.696691513,0.789017341)
(0.696691513,0.791907514)
(0.697780609,0.791907514)
(0.697780609,0.794797688)
(0.737749100,0.794797688)
(0.737749100,0.797687861)
(0.750115395,0.797687861)
(0.750115395,0.800578035)
(0.790797234,0.800578035)
(0.790797234,0.803468208)
(0.797835350,0.803468208)
(0.797835350,0.806358382)
(0.879270554,0.806358382)
(0.879270554,0.809248555)
(0.885089874,0.809248555)
(0.885089874,0.812138728)
(0.900024414,0.812138728)
(0.900024414,0.815028902)
(0.916410446,0.815028902)
(0.916410446,0.817919075)
(0.926499367,0.817919075)
(0.926499367,0.820809249)
(0.954527855,0.820809249)
(0.954527855,0.823699422)
(0.964322090,0.823699422)
(0.964322090,0.826589595)
(0.966232300,0.826589595)
(0.966232300,0.829479769)
(0.989961624,0.829479769)
(0.989961624,0.832369942)
(1.016181946,0.832369942)
(1.016181946,0.835260116)
(1.038110733,0.835260116)
(1.038110733,0.838150289)
(1.068450928,0.838150289)
(1.068450928,0.841040462)
(1.089482307,0.841040462)
(1.089482307,0.843930636)
(1.099287033,0.843930636)
(1.099287033,0.846820809)
(1.109002113,0.846820809)
(1.109002113,0.849710983)
(1.120049477,0.849710983)
(1.120049477,0.852601156)
(1.133699417,0.852601156)
(1.133699417,0.855491329)
(1.145616531,0.855491329)
(1.145616531,0.858381503)
(1.169005394,0.858381503)
(1.169005394,0.861271676)
(1.170662880,0.861271676)
(1.170662880,0.864161850)
(1.195034981,0.864161850)
(1.195034981,0.867052023)
(1.244953156,0.867052023)
(1.244953156,0.869942197)
(1.254439354,0.869942197)
(1.254439354,0.872832370)
(1.260347366,0.872832370)
(1.260347366,0.875722543)
(1.338329315,0.875722543)
(1.338329315,0.878612717)
(1.342985153,0.878612717)
(1.342985153,0.881502890)
(1.345363617,0.881502890)
(1.345363617,0.884393064)
(1.357028961,0.884393064)
(1.357028961,0.887283237)
(1.419325829,0.887283237)
(1.419325829,0.890173410)
(1.427719116,0.890173410)
(1.427719116,0.893063584)
(1.449411392,0.893063584)
(1.449411392,0.895953757)
(1.487504005,0.895953757)
(1.487504005,0.898843931)
(1.553956032,0.898843931)
(1.553956032,0.901734104)
(1.559155464,0.901734104)
(1.559155464,0.904624277)
(1.572160721,0.904624277)
(1.572160721,0.907514451)
(1.580820084,0.907514451)
(1.580820084,0.910404624)
(1.581570625,0.910404624)
(1.581570625,0.913294798)
(1.584914207,0.913294798)
(1.584914207,0.916184971)
(1.599875450,0.916184971)
(1.599875450,0.919075145)
(1.688829422,0.919075145)
(1.688829422,0.921965318)
(1.710987091,0.921965318)
(1.710987091,0.924855491)
(1.740424156,0.924855491)
(1.740424156,0.927745665)
(1.793855667,0.927745665)
(1.793855667,0.930635838)
(1.801634789,0.930635838)
(1.801634789,0.933526012)
(1.818881035,0.933526012)
(1.818881035,0.936416185)
(1.824797630,0.936416185)
(1.824797630,0.939306358)
(1.837049484,0.939306358)
(1.837049484,0.942196532)
(1.879178047,0.942196532)
(1.879178047,0.945086705)
(1.893764496,0.945086705)
(1.893764496,0.947976879)
(1.899817467,0.947976879)
(1.899817467,0.950867052)
(1.911255836,0.950867052)
(1.911255836,0.953757225)
(1.992321014,0.953757225)
(1.992321014,0.956647399)
(1.999379158,0.956647399)
(1.999379158,0.959537572)
(2.015516281,0.959537572)
(2.015516281,0.962427746)
(2.053991318,0.962427746)
(2.053991318,0.965317919)
(2.092730522,0.965317919)
(2.092730522,0.968208092)
(2.157390594,0.968208092)
(2.157390594,0.971098266)
(2.166426659,0.971098266)
(2.166426659,0.973988439)
(2.203158379,0.973988439)
(2.203158379,0.976878613)
(2.274284363,0.976878613)
(2.274284363,0.979768786)
(2.309514999,0.979768786)
(2.309514999,0.982658960)
(2.319361687,0.982658960)
(2.319361687,0.985549133)
(2.401414871,0.985549133)
(2.401414871,0.988439306)
(2.735945702,0.988439306)
(2.735945702,0.991329480)
(2.808104515,0.991329480)
(2.808104515,0.994219653)
(3.211774826,0.994219653)
(3.211774826,0.997109827)
(3.592141151,0.997109827)
(3.592141151,1.000000000)
(6,1)
};
\draw[inputcurve] plot coordinates {
(-1,0)
(-0.108776093,0.000000000)
(-0.108776093,0.002890173)
(-0.080340385,0.002890173)
(-0.080340385,0.005780347)
(-0.077370644,0.005780347)
(-0.077370644,0.008670520)
(-0.075239182,0.008670520)
(-0.075239182,0.011560694)
(-0.074202538,0.011560694)
(-0.074202538,0.014450867)
(-0.061696053,0.014450867)
(-0.061696053,0.017341040)
(-0.057372093,0.017341040)
(-0.057372093,0.020231214)
(-0.055396080,0.020231214)
(-0.055396080,0.023121387)
(-0.048828125,0.023121387)
(-0.048828125,0.026011561)
(-0.045791626,0.026011561)
(-0.045791626,0.028901734)
(-0.044024467,0.028901734)
(-0.044024467,0.031791908)
(-0.035203934,0.031791908)
(-0.035203934,0.034682081)
(-0.034371376,0.034682081)
(-0.034371376,0.037572254)
(-0.032317162,0.037572254)
(-0.032317162,0.040462428)
(-0.031192780,0.040462428)
(-0.031192780,0.043352601)
(-0.031018257,0.043352601)
(-0.031018257,0.046242775)
(-0.029425621,0.046242775)
(-0.029425621,0.049132948)
(-0.029253960,0.049132948)
(-0.029253960,0.052023121)
(-0.026247025,0.052023121)
(-0.026247025,0.054913295)
(-0.023991585,0.054913295)
(-0.023991585,0.057803468)
(-0.020537376,0.057803468)
(-0.020537376,0.060693642)
(-0.019435883,0.060693642)
(-0.019435883,0.063583815)
(-0.016783714,0.063583815)
(-0.016783714,0.066473988)
(-0.016373634,0.066473988)
(-0.016373634,0.069364162)
(-0.015058517,0.069364162)
(-0.015058517,0.072254335)
(-0.014232635,0.072254335)
(-0.014232635,0.075144509)
(-0.012382507,0.075144509)
(-0.012382507,0.078034682)
(-0.010263443,0.078034682)
(-0.010263443,0.080924855)
(-0.010263443,0.080924855)
(-0.010263443,0.083815029)
(-0.008370399,0.083815029)
(-0.008370399,0.086705202)
(-0.008307457,0.086705202)
(-0.008307457,0.089595376)
(-0.006370544,0.089595376)
(-0.006370544,0.092485549)
(-0.004253387,0.092485549)
(-0.004253387,0.095375723)
(-0.002460480,0.095375723)
(-0.002460480,0.098265896)
(-0.002004623,0.098265896)
(-0.002004623,0.101156069)
(0.000000000,0.101156069)
(0.000000000,0.104046243)
(0.000000000,0.104046243)
(0.000000000,0.106936416)
(0.000000000,0.106936416)
(0.000000000,0.109826590)
(0.000000000,0.109826590)
(0.000000000,0.112716763)
(0.000000000,0.112716763)
(0.000000000,0.115606936)
(0.000000000,0.115606936)
(0.000000000,0.118497110)
(0.000000000,0.118497110)
(0.000000000,0.121387283)
(0.000000000,0.121387283)
(0.000000000,0.124277457)
(0.000000000,0.124277457)
(0.000000000,0.127167630)
(0.000000000,0.127167630)
(0.000000000,0.130057803)
(0.000000000,0.130057803)
(0.000000000,0.132947977)
(0.000000000,0.132947977)
(0.000000000,0.135838150)
(0.000000000,0.135838150)
(0.000000000,0.138728324)
(0.000000000,0.138728324)
(0.000000000,0.141618497)
(0.000000000,0.141618497)
(0.000000000,0.144508671)
(0.000000000,0.144508671)
(0.000000000,0.147398844)
(0.000000000,0.147398844)
(0.000000000,0.150289017)
(0.000000000,0.150289017)
(0.000000000,0.153179191)
(0.000000000,0.153179191)
(0.000000000,0.156069364)
(0.000000000,0.156069364)
(0.000000000,0.158959538)
(0.000000000,0.158959538)
(0.000000000,0.161849711)
(0.000000000,0.161849711)
(0.000000000,0.164739884)
(0.000000000,0.164739884)
(0.000000000,0.167630058)
(0.000000000,0.167630058)
(0.000000000,0.170520231)
(0.000000000,0.170520231)
(0.000000000,0.173410405)
(0.000000000,0.173410405)
(0.000000000,0.176300578)
(0.000000000,0.176300578)
(0.000000000,0.179190751)
(0.001097679,0.179190751)
(0.001097679,0.182080925)
(0.002044678,0.182080925)
(0.002044678,0.184971098)
(0.002761841,0.184971098)
(0.002761841,0.187861272)
(0.004152298,0.187861272)
(0.004152298,0.190751445)
(0.006630898,0.190751445)
(0.006630898,0.193641618)
(0.007412910,0.193641618)
(0.007412910,0.196531792)
(0.008168221,0.196531792)
(0.008168221,0.199421965)
(0.009707451,0.199421965)
(0.009707451,0.202312139)
(0.009882927,0.202312139)
(0.009882927,0.205202312)
(0.010493279,0.205202312)
(0.010493279,0.208092486)
(0.011522293,0.208092486)
(0.011522293,0.210982659)
(0.013235092,0.210982659)
(0.013235092,0.213872832)
(0.013545990,0.213872832)
(0.013545990,0.216763006)
(0.014237404,0.216763006)
(0.014237404,0.219653179)
(0.014495850,0.219653179)
(0.014495850,0.222543353)
(0.015273094,0.222543353)
(0.015273094,0.225433526)
(0.016195297,0.225433526)
(0.016195297,0.228323699)
(0.016372681,0.228323699)
(0.016372681,0.231213873)
(0.016725540,0.231213873)
(0.016725540,0.234104046)
(0.017350197,0.234104046)
(0.017350197,0.236994220)
(0.018731117,0.236994220)
(0.018731117,0.239884393)
(0.019000053,0.239884393)
(0.019000053,0.242774566)
(0.019214630,0.242774566)
(0.019214630,0.245664740)
(0.019676208,0.245664740)
(0.019676208,0.248554913)
(0.020458221,0.248554913)
(0.020458221,0.251445087)
(0.020468712,0.251445087)
(0.020468712,0.254335260)
(0.020979881,0.254335260)
(0.020979881,0.257225434)
(0.021486282,0.257225434)
(0.021486282,0.260115607)
(0.022758484,0.260115607)
(0.022758484,0.263005780)
(0.023287773,0.263005780)
(0.023287773,0.265895954)
(0.024775505,0.265895954)
(0.024775505,0.268786127)
(0.025491714,0.268786127)
(0.025491714,0.271676301)
(0.025591850,0.271676301)
(0.025591850,0.274566474)
(0.025613785,0.274566474)
(0.025613785,0.277456647)
(0.026303291,0.277456647)
(0.026303291,0.280346821)
(0.026551247,0.280346821)
(0.026551247,0.283236994)
(0.027061462,0.283236994)
(0.027061462,0.286127168)
(0.029011726,0.286127168)
(0.029011726,0.289017341)
(0.029538155,0.289017341)
(0.029538155,0.291907514)
(0.029949188,0.291907514)
(0.029949188,0.294797688)
(0.030074120,0.294797688)
(0.030074120,0.297687861)
(0.030097961,0.297687861)
(0.030097961,0.300578035)
(0.030569077,0.300578035)
(0.030569077,0.303468208)
(0.031017303,0.303468208)
(0.031017303,0.306358382)
(0.032470703,0.306358382)
(0.032470703,0.309248555)
(0.033316612,0.309248555)
(0.033316612,0.312138728)
(0.033376694,0.312138728)
(0.033376694,0.315028902)
(0.033397675,0.315028902)
(0.033397675,0.317919075)
(0.033614159,0.317919075)
(0.033614159,0.320809249)
(0.033756256,0.320809249)
(0.033756256,0.323699422)
(0.035354614,0.323699422)
(0.035354614,0.326589595)
(0.035623550,0.326589595)
(0.035623550,0.329479769)
(0.035650253,0.329479769)
(0.035650253,0.332369942)
(0.036245346,0.332369942)
(0.036245346,0.335260116)
(0.036437988,0.335260116)
(0.036437988,0.338150289)
(0.036928177,0.338150289)
(0.036928177,0.341040462)
(0.037925720,0.341040462)
(0.037925720,0.343930636)
(0.038696289,0.343930636)
(0.038696289,0.346820809)
(0.039712906,0.346820809)
(0.039712906,0.349710983)
(0.040022850,0.349710983)
(0.040022850,0.352601156)
(0.040402412,0.352601156)
(0.040402412,0.355491329)
(0.040812492,0.355491329)
(0.040812492,0.358381503)
(0.041164398,0.358381503)
(0.041164398,0.361271676)
(0.041851044,0.361271676)
(0.041851044,0.364161850)
(0.041916847,0.364161850)
(0.041916847,0.367052023)
(0.042443275,0.367052023)
(0.042443275,0.369942197)
(0.043488503,0.369942197)
(0.043488503,0.372832370)
(0.043491364,0.372832370)
(0.043491364,0.375722543)
(0.043935776,0.375722543)
(0.043935776,0.378612717)
(0.043938637,0.378612717)
(0.043938637,0.381502890)
(0.044615746,0.381502890)
(0.044615746,0.384393064)
(0.044834137,0.384393064)
(0.044834137,0.387283237)
(0.046386719,0.387283237)
(0.046386719,0.390173410)
(0.046483994,0.390173410)
(0.046483994,0.393063584)
(0.047964096,0.393063584)
(0.047964096,0.395953757)
(0.048169136,0.395953757)
(0.048169136,0.398843931)
(0.048281670,0.398843931)
(0.048281670,0.401734104)
(0.049042702,0.401734104)
(0.049042702,0.404624277)
(0.049091339,0.404624277)
(0.049091339,0.407514451)
(0.049182892,0.407514451)
(0.049182892,0.410404624)
(0.049502373,0.410404624)
(0.049502373,0.413294798)
(0.051596642,0.413294798)
(0.051596642,0.416184971)
(0.051953316,0.416184971)
(0.051953316,0.419075145)
(0.053852081,0.419075145)
(0.053852081,0.421965318)
(0.053971291,0.421965318)
(0.053971291,0.424855491)
(0.054196358,0.424855491)
(0.054196358,0.427745665)
(0.054553032,0.427745665)
(0.054553032,0.430635838)
(0.054925919,0.430635838)
(0.054925919,0.433526012)
(0.056351662,0.433526012)
(0.056351662,0.436416185)
(0.057215691,0.436416185)
(0.057215691,0.439306358)
(0.057555199,0.439306358)
(0.057555199,0.442196532)
(0.059920311,0.442196532)
(0.059920311,0.445086705)
(0.061487198,0.445086705)
(0.061487198,0.447976879)
(0.061651230,0.447976879)
(0.061651230,0.450867052)
(0.064482689,0.450867052)
(0.064482689,0.453757225)
(0.064531326,0.453757225)
(0.064531326,0.456647399)
(0.065914154,0.456647399)
(0.065914154,0.459537572)
(0.066601753,0.459537572)
(0.066601753,0.462427746)
(0.066942215,0.462427746)
(0.066942215,0.465317919)
(0.067325592,0.465317919)
(0.067325592,0.468208092)
(0.069304466,0.468208092)
(0.069304466,0.471098266)
(0.069907665,0.471098266)
(0.069907665,0.473988439)
(0.073431015,0.473988439)
(0.073431015,0.476878613)
(0.078474045,0.476878613)
(0.078474045,0.479768786)
(0.079440117,0.479768786)
(0.079440117,0.482658960)
(0.079928398,0.482658960)
(0.079928398,0.485549133)
(0.081627846,0.485549133)
(0.081627846,0.488439306)
(0.083364487,0.488439306)
(0.083364487,0.491329480)
(0.084113121,0.491329480)
(0.084113121,0.494219653)
(0.085491180,0.494219653)
(0.085491180,0.497109827)
(0.088352203,0.497109827)
(0.088352203,0.500000000)
(0.092400551,0.500000000)
(0.092400551,0.502890173)
(0.092604637,0.502890173)
(0.092604637,0.505780347)
(0.093832970,0.505780347)
(0.093832970,0.508670520)
(0.094056129,0.508670520)
(0.094056129,0.511560694)
(0.095314980,0.511560694)
(0.095314980,0.514450867)
(0.095438957,0.514450867)
(0.095438957,0.517341040)
(0.095532417,0.517341040)
(0.095532417,0.520231214)
(0.095983505,0.520231214)
(0.095983505,0.523121387)
(0.096186638,0.523121387)
(0.096186638,0.526011561)
(0.096360207,0.526011561)
(0.096360207,0.528901734)
(0.096861839,0.528901734)
(0.096861839,0.531791908)
(0.097108841,0.531791908)
(0.097108841,0.534682081)
(0.097919464,0.534682081)
(0.097919464,0.537572254)
(0.099804878,0.537572254)
(0.099804878,0.540462428)
(0.100025177,0.540462428)
(0.100025177,0.543352601)
(0.100595474,0.543352601)
(0.100595474,0.546242775)
(0.109051704,0.546242775)
(0.109051704,0.549132948)
(0.109198570,0.549132948)
(0.109198570,0.552023121)
(0.110455513,0.552023121)
(0.110455513,0.554913295)
(0.113307953,0.554913295)
(0.113307953,0.557803468)
(0.116171837,0.557803468)
(0.116171837,0.560693642)
(0.116653442,0.560693642)
(0.116653442,0.563583815)
(0.117283821,0.563583815)
(0.117283821,0.566473988)
(0.117413521,0.566473988)
(0.117413521,0.569364162)
(0.118077278,0.569364162)
(0.118077278,0.572254335)
(0.124880791,0.572254335)
(0.124880791,0.575144509)
(0.127660751,0.575144509)
(0.127660751,0.578034682)
(0.130516052,0.578034682)
(0.130516052,0.580924855)
(0.130741119,0.580924855)
(0.130741119,0.583815029)
(0.130990028,0.583815029)
(0.130990028,0.586705202)
(0.131756783,0.586705202)
(0.131756783,0.589595376)
(0.136437416,0.589595376)
(0.136437416,0.592485549)
(0.139389038,0.592485549)
(0.139389038,0.595375723)
(0.140806198,0.595375723)
(0.140806198,0.598265896)
(0.143755913,0.598265896)
(0.143755913,0.601156069)
(0.144680023,0.601156069)
(0.144680023,0.604046243)
(0.152574539,0.604046243)
(0.152574539,0.606936416)
(0.153215408,0.606936416)
(0.153215408,0.609826590)
(0.153651237,0.609826590)
(0.153651237,0.612716763)
(0.156265259,0.612716763)
(0.156265259,0.615606936)
(0.159843445,0.615606936)
(0.159843445,0.618497110)
(0.165087700,0.618497110)
(0.165087700,0.621387283)
(0.167132378,0.621387283)
(0.167132378,0.624277457)
(0.167766571,0.624277457)
(0.167766571,0.627167630)
(0.169247627,0.627167630)
(0.169247627,0.630057803)
(0.171792030,0.630057803)
(0.171792030,0.632947977)
(0.174821854,0.632947977)
(0.174821854,0.635838150)
(0.178028107,0.635838150)
(0.178028107,0.638728324)
(0.179435730,0.638728324)
(0.179435730,0.641618497)
(0.180336952,0.641618497)
(0.180336952,0.644508671)
(0.182412148,0.644508671)
(0.182412148,0.647398844)
(0.186467171,0.647398844)
(0.186467171,0.650289017)
(0.189964294,0.650289017)
(0.189964294,0.653179191)
(0.191658974,0.653179191)
(0.191658974,0.656069364)
(0.197575569,0.656069364)
(0.197575569,0.658959538)
(0.203168869,0.658959538)
(0.203168869,0.661849711)
(0.203684807,0.661849711)
(0.203684807,0.664739884)
(0.217490196,0.664739884)
(0.217490196,0.667630058)
(0.219084740,0.667630058)
(0.219084740,0.670520231)
(0.237065315,0.670520231)
(0.237065315,0.673410405)
(0.237843513,0.673410405)
(0.237843513,0.676300578)
(0.239300728,0.676300578)
(0.239300728,0.679190751)
(0.250901222,0.679190751)
(0.250901222,0.682080925)
(0.252483368,0.682080925)
(0.252483368,0.684971098)
(0.262947083,0.684971098)
(0.262947083,0.687861272)
(0.269662857,0.687861272)
(0.269662857,0.690751445)
(0.291261673,0.690751445)
(0.291261673,0.693641618)
(0.295753479,0.693641618)
(0.295753479,0.696531792)
(0.308116913,0.696531792)
(0.308116913,0.699421965)
(0.413013458,0.699421965)
(0.413013458,0.702312139)
(0.530436516,0.702312139)
(0.530436516,0.705202312)
(0.545284271,0.705202312)
(0.545284271,0.708092486)
(0.594315529,0.708092486)
(0.594315529,0.710982659)
(0.609313965,0.710982659)
(0.609313965,0.713872832)
(0.842195511,0.713872832)
(0.842195511,0.716763006)
(0.913593292,0.716763006)
(0.913593292,0.719653179)
(0.932668686,0.719653179)
(0.932668686,0.722543353)
(0.964049339,0.722543353)
(0.964049339,0.725433526)
(0.995770454,0.725433526)
(0.995770454,0.728323699)
(1.020518303,0.728323699)
(1.020518303,0.731213873)
(1.096347809,0.731213873)
(1.096347809,0.734104046)
(1.131927490,0.734104046)
(1.131927490,0.736994220)
(1.151204109,0.736994220)
(1.151204109,0.739884393)
(1.189287186,0.739884393)
(1.189287186,0.742774566)
(1.240646362,0.742774566)
(1.240646362,0.745664740)
(1.253330231,0.745664740)
(1.253330231,0.748554913)
(1.301481247,0.748554913)
(1.301481247,0.751445087)
(1.312383652,0.751445087)
(1.312383652,0.754335260)
(1.312801361,0.754335260)
(1.312801361,0.757225434)
(1.364917755,0.757225434)
(1.364917755,0.760115607)
(1.398091316,0.760115607)
(1.398091316,0.763005780)
(1.428270340,0.763005780)
(1.428270340,0.765895954)
(1.442154884,0.765895954)
(1.442154884,0.768786127)
(1.477525711,0.768786127)
(1.477525711,0.771676301)
(1.504503250,0.771676301)
(1.504503250,0.774566474)
(1.541754723,0.774566474)
(1.541754723,0.777456647)
(1.542875290,0.777456647)
(1.542875290,0.780346821)
(1.548994064,0.780346821)
(1.548994064,0.783236994)
(1.632438660,0.783236994)
(1.632438660,0.786127168)
(1.638628006,0.786127168)
(1.638628006,0.789017341)
(1.665376663,0.789017341)
(1.665376663,0.791907514)
(1.727894783,0.791907514)
(1.727894783,0.794797688)
(1.730357170,0.794797688)
(1.730357170,0.797687861)
(1.911924362,0.797687861)
(1.911924362,0.800578035)
(1.926698685,0.800578035)
(1.926698685,0.803468208)
(2.026398659,0.803468208)
(2.026398659,0.806358382)
(2.067292213,0.806358382)
(2.067292213,0.809248555)
(2.098522186,0.809248555)
(2.098522186,0.812138728)
(2.160779953,0.812138728)
(2.160779953,0.815028902)
(2.171801567,0.815028902)
(2.171801567,0.817919075)
(2.245495796,0.817919075)
(2.245495796,0.820809249)
(2.247835159,0.820809249)
(2.247835159,0.823699422)
(2.249439240,0.823699422)
(2.249439240,0.826589595)
(2.250622749,0.826589595)
(2.250622749,0.829479769)
(2.286360741,0.829479769)
(2.286360741,0.832369942)
(2.306130409,0.832369942)
(2.306130409,0.835260116)
(2.332760811,0.835260116)
(2.332760811,0.838150289)
(2.360136032,0.838150289)
(2.360136032,0.841040462)
(2.376787186,0.841040462)
(2.376787186,0.843930636)
(2.424018860,0.843930636)
(2.424018860,0.846820809)
(2.460692406,0.846820809)
(2.460692406,0.849710983)
(2.465215683,0.849710983)
(2.465215683,0.852601156)
(2.509326935,0.852601156)
(2.509326935,0.855491329)
(2.564399719,0.855491329)
(2.564399719,0.858381503)
(2.572593689,0.858381503)
(2.572593689,0.861271676)
(2.577330589,0.861271676)
(2.577330589,0.864161850)
(2.587465286,0.864161850)
(2.587465286,0.867052023)
(2.667371750,0.867052023)
(2.667371750,0.869942197)
(2.777155876,0.869942197)
(2.777155876,0.872832370)
(2.787052155,0.872832370)
(2.787052155,0.875722543)
(2.849433899,0.875722543)
(2.849433899,0.878612717)
(2.905464172,0.878612717)
(2.905464172,0.881502890)
(2.983240128,0.881502890)
(2.983240128,0.884393064)
(3.057772636,0.884393064)
(3.057772636,0.887283237)
(3.111129761,0.887283237)
(3.111129761,0.890173410)
(3.117304802,0.890173410)
(3.117304802,0.893063584)
(3.144447327,0.893063584)
(3.144447327,0.895953757)
(3.145476341,0.895953757)
(3.145476341,0.898843931)
(3.153233528,0.898843931)
(3.153233528,0.901734104)
(3.224532127,0.901734104)
(3.224532127,0.904624277)
(3.225382805,0.904624277)
(3.225382805,0.907514451)
(3.230832100,0.907514451)
(3.230832100,0.910404624)
(3.236519814,0.910404624)
(3.236519814,0.913294798)
(3.256375313,0.913294798)
(3.256375313,0.916184971)
(3.273511887,0.916184971)
(3.273511887,0.919075145)
(3.274180412,0.919075145)
(3.274180412,0.921965318)
(3.485916138,0.921965318)
(3.485916138,0.924855491)
(3.514197350,0.924855491)
(3.514197350,0.927745665)
(3.520363808,0.927745665)
(3.520363808,0.930635838)
(3.550540924,0.930635838)
(3.550540924,0.933526012)
(3.563360214,0.933526012)
(3.563360214,0.936416185)
(3.667315483,0.936416185)
(3.667315483,0.939306358)
(3.934188843,0.939306358)
(3.934188843,0.942196532)
(4.001835823,0.942196532)
(4.001835823,0.945086705)
(4.016134262,0.945086705)
(4.016134262,0.947976879)
(4.027205467,0.947976879)
(4.027205467,0.950867052)
(4.032059669,0.950867052)
(4.032059669,0.953757225)
(4.038103104,0.953757225)
(4.038103104,0.956647399)
(4.045647621,0.956647399)
(4.045647621,0.959537572)
(4.138804436,0.959537572)
(4.138804436,0.962427746)
(4.163980484,0.962427746)
(4.163980484,0.965317919)
(4.169958115,0.965317919)
(4.169958115,0.968208092)
(4.191331863,0.968208092)
(4.191331863,0.971098266)
(4.300021172,0.971098266)
(4.300021172,0.973988439)
(4.318253517,0.973988439)
(4.318253517,0.976878613)
(4.496239662,0.976878613)
(4.496239662,0.979768786)
(4.510597229,0.979768786)
(4.510597229,0.982658960)
(4.875897408,0.982658960)
(4.875897408,0.985549133)
(5.069027901,0.985549133)
(5.069027901,0.988439306)
(5.078999519,0.988439306)
(5.078999519,0.991329480)
(5.133082390,0.991329480)
(5.133082390,0.994219653)
(5.522347450,0.994219653)
(5.522347450,0.997109827)
(5.824607849,0.997109827)
(5.824607849,1.000000000)
(6,1)
};

%% file: figures/global_ecdf_2b_paths.tex
\draw[globalcurve] plot coordinates {
(-1,0)
(-0.861026764,0.000000000)
(-0.861026764,0.015625000)
(-0.587623596,0.015625000)
(-0.587623596,0.031250000)
(-0.568748474,0.031250000)
(-0.568748474,0.046875000)
(-0.521778107,0.046875000)
(-0.521778107,0.062500000)
(-0.438901901,0.062500000)
(-0.438901901,0.078125000)
(-0.396575928,0.078125000)
(-0.396575928,0.093750000)
(-0.382612228,0.093750000)
(-0.382612228,0.109375000)
(-0.360389709,0.109375000)
(-0.360389709,0.125000000)
(-0.352764130,0.125000000)
(-0.352764130,0.140625000)
(-0.352453232,0.140625000)
(-0.352453232,0.156250000)
(-0.340436935,0.156250000)
(-0.340436935,0.171875000)
(-0.301435471,0.171875000)
(-0.301435471,0.187500000)
(-0.275234222,0.187500000)
(-0.275234222,0.203125000)
(-0.272821426,0.203125000)
(-0.272821426,0.218750000)
(-0.227462769,0.218750000)
(-0.227462769,0.234375000)
(-0.198266983,0.234375000)
(-0.198266983,0.250000000)
(-0.186391830,0.250000000)
(-0.186391830,0.265625000)
(-0.139110565,0.265625000)
(-0.139110565,0.281250000)
(-0.078475952,0.281250000)
(-0.078475952,0.296875000)
(-0.075891495,0.296875000)
(-0.075891495,0.312500000)
(-0.058027267,0.312500000)
(-0.058027267,0.328125000)
(-0.048089981,0.328125000)
(-0.048089981,0.343750000)
(-0.031484604,0.343750000)
(-0.031484604,0.359375000)
(-0.011810303,0.359375000)
(-0.011810303,0.375000000)
(0.026081085,0.375000000)
(0.026081085,0.390625000)
(0.043815613,0.390625000)
(0.043815613,0.406250000)
(0.053903580,0.406250000)
(0.053903580,0.421875000)
(0.073278427,0.421875000)
(0.073278427,0.437500000)
(0.119876862,0.437500000)
(0.119876862,0.453125000)
(0.148891449,0.453125000)
(0.148891449,0.468750000)
(0.152387619,0.468750000)
(0.152387619,0.484375000)
(0.155971527,0.484375000)
(0.155971527,0.500000000)
(0.243648529,0.500000000)
(0.243648529,0.515625000)
(0.297264099,0.515625000)
(0.297264099,0.531250000)
(0.386438370,0.531250000)
(0.386438370,0.546875000)
(0.434816360,0.546875000)
(0.434816360,0.562500000)
(0.492834091,0.562500000)
(0.492834091,0.578125000)
(0.494705200,0.578125000)
(0.494705200,0.593750000)
(0.509937286,0.593750000)
(0.509937286,0.609375000)
(0.520250320,0.609375000)
(0.520250320,0.625000000)
(0.548591614,0.625000000)
(0.548591614,0.640625000)
(0.571584702,0.640625000)
(0.571584702,0.656250000)
(0.598485947,0.656250000)
(0.598485947,0.671875000)
(0.605100632,0.671875000)
(0.605100632,0.687500000)
(0.620965958,0.687500000)
(0.620965958,0.703125000)
(0.644104004,0.703125000)
(0.644104004,0.718750000)
(0.658437729,0.718750000)
(0.658437729,0.734375000)
(0.705905914,0.734375000)
(0.705905914,0.750000000)
(0.711912155,0.750000000)
(0.711912155,0.765625000)
(0.737335205,0.765625000)
(0.737335205,0.781250000)
(0.743114471,0.781250000)
(0.743114471,0.796875000)
(0.747243881,0.796875000)
(0.747243881,0.812500000)
(0.773269653,0.812500000)
(0.773269653,0.828125000)
(0.798223495,0.828125000)
(0.798223495,0.843750000)
(0.847850800,0.843750000)
(0.847850800,0.859375000)
(0.958585739,0.859375000)
(0.958585739,0.875000000)
(0.979230881,0.875000000)
(0.979230881,0.890625000)
(1.008043289,0.890625000)
(1.008043289,0.906250000)
(1.021524429,0.906250000)
(1.021524429,0.921875000)
(1.027803421,0.921875000)
(1.027803421,0.937500000)
(1.062084198,0.937500000)
(1.062084198,0.953125000)
(1.065429688,0.953125000)
(1.065429688,0.968750000)
(1.130714417,0.968750000)
(1.130714417,0.984375000)
(1.248256683,0.984375000)
(1.248256683,1.000000000)
(6,1)
};
\draw[inputcurve] plot coordinates {
(-1,0)
(0.000000000,0.000000000)
(0.000000000,0.015625000)
(0.023693085,0.015625000)
(0.023693085,0.031250000)
(0.025667191,0.031250000)
(0.025667191,0.046875000)
(0.027669907,0.046875000)
(0.027669907,0.062500000)
(0.032953262,0.062500000)
(0.032953262,0.078125000)
(0.034318924,0.078125000)
(0.034318924,0.093750000)
(0.072126389,0.093750000)
(0.072126389,0.109375000)
(0.075050354,0.109375000)
(0.075050354,0.125000000)
(0.080568314,0.125000000)
(0.080568314,0.140625000)
(0.094377518,0.140625000)
(0.094377518,0.156250000)
(0.105283737,0.156250000)
(0.105283737,0.171875000)
(0.111392975,0.171875000)
(0.111392975,0.187500000)
(0.155040741,0.187500000)
(0.155040741,0.203125000)
(0.161880493,0.203125000)
(0.161880493,0.218750000)
(0.163890839,0.218750000)
(0.163890839,0.234375000)
(0.167522430,0.234375000)
(0.167522430,0.250000000)
(0.181665421,0.250000000)
(0.181665421,0.265625000)
(0.183639526,0.265625000)
(0.183639526,0.281250000)
(0.184219360,0.281250000)
(0.184219360,0.296875000)
(0.192697525,0.296875000)
(0.192697525,0.312500000)
(0.201477051,0.312500000)
(0.201477051,0.328125000)
(0.208328247,0.328125000)
(0.208328247,0.343750000)
(0.220357895,0.343750000)
(0.220357895,0.359375000)
(0.227359772,0.359375000)
(0.227359772,0.375000000)
(0.233089447,0.375000000)
(0.233089447,0.390625000)
(0.242906570,0.390625000)
(0.242906570,0.406250000)
(0.243648529,0.406250000)
(0.243648529,0.421875000)
(0.299591064,0.421875000)
(0.299591064,0.437500000)
(0.321306229,0.437500000)
(0.321306229,0.453125000)
(0.338266373,0.453125000)
(0.338266373,0.468750000)
(0.372083664,0.468750000)
(0.372083664,0.484375000)
(0.402788162,0.484375000)
(0.402788162,0.500000000)
(0.402816772,0.500000000)
(0.402816772,0.515625000)
(0.416368484,0.515625000)
(0.416368484,0.531250000)
(0.426460266,0.531250000)
(0.426460266,0.546875000)
(0.516817093,0.546875000)
(0.516817093,0.562500000)
(0.544530869,0.562500000)
(0.544530869,0.578125000)
(0.569732666,0.578125000)
(0.569732666,0.593750000)
(0.628448486,0.593750000)
(0.628448486,0.609375000)
(0.720451355,0.609375000)
(0.720451355,0.625000000)
(0.722028732,0.625000000)
(0.722028732,0.640625000)
(0.756763458,0.640625000)
(0.756763458,0.656250000)
(0.776968002,0.656250000)
(0.776968002,0.671875000)
(0.781808853,0.671875000)
(0.781808853,0.687500000)
(0.782978058,0.687500000)
(0.782978058,0.703125000)
(0.786214828,0.703125000)
(0.786214828,0.718750000)
(0.840858459,0.718750000)
(0.840858459,0.734375000)
(0.841459274,0.734375000)
(0.841459274,0.750000000)
(0.858592987,0.750000000)
(0.858592987,0.765625000)
(0.911346436,0.765625000)
(0.911346436,0.781250000)
(0.918283463,0.781250000)
(0.918283463,0.796875000)
(0.929296494,0.796875000)
(0.929296494,0.812500000)
(0.970483780,0.812500000)
(0.970483780,0.828125000)
(0.991390228,0.828125000)
(0.991390228,0.843750000)
(1.036001205,0.843750000)
(1.036001205,0.859375000)
(1.100914001,0.859375000)
(1.100914001,0.875000000)
(1.155950546,0.875000000)
(1.155950546,0.890625000)
(1.329704285,0.890625000)
(1.329704285,0.906250000)
(1.366168976,0.906250000)
(1.366168976,0.921875000)
(1.391756058,0.921875000)
(1.391756058,0.937500000)
(1.415021896,0.937500000)
(1.415021896,0.953125000)
(1.472618103,0.953125000)
(1.472618103,0.968750000)
(1.480014801,0.968750000)
(1.480014801,0.984375000)
(1.554008484,0.984375000)
(1.554008484,1.000000000)
(6,1)
};

%% file: figures/global_fullrange.tex
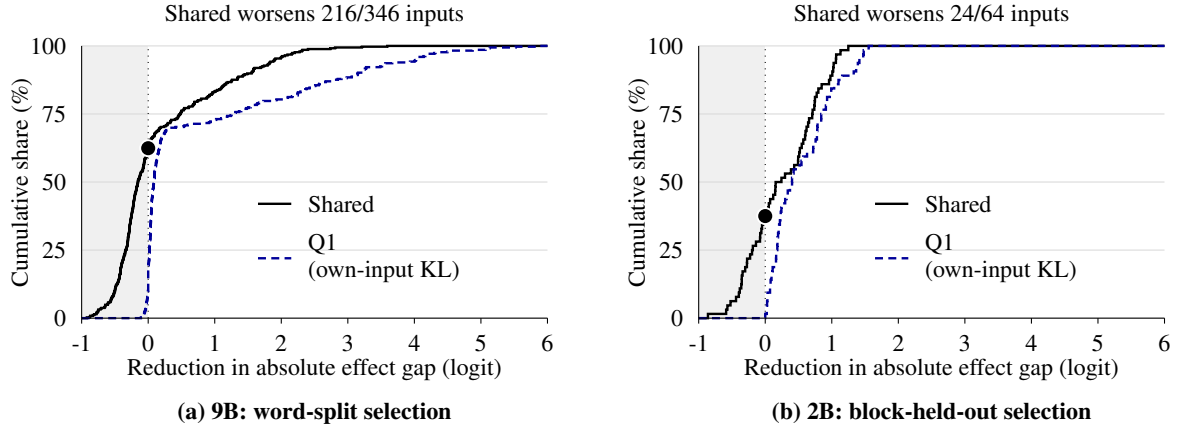
\begin{figure*}[t]
\centering
\newcommand{\fullcalibrationpanel}[4]{%
\begin{tikzpicture}[x=.88cm,y=3.6cm,font=\small,
 globalcurve/.style={black,line width=.9pt},
 inputcurve/.style={blue!60!black,densely dashed,line width=.9pt}]
\fill[black!6] (-1,0) rectangle (0,1);
\foreach \y/\label in {0/0,.25/25,.5/50,.75/75,1/100} {
  \draw[black!15] (-1,\y) -- (6,\y);
  \node[anchor=east] at (-1.08,\y) {\label};
}
\draw[dotted] (0,0) -- (0,1);
\input{#2}
\draw (-1,0) -- (6,0);
\draw (-1,0) -- (-1,1);
\foreach \x in {-1,0,1,2,3,4,5,6} {
  \draw (\x,0) -- (\x,-.02);
  \node[anchor=north] at (\x,-.025) {\x};
}
\node[rotate=90] at (-1.9,.5) {Cumulative share (\%)};
\node at (2.5,-.19) {Reduction in absolute effect gap (logit)};
\node[font=\small\bfseries,text height=1.5ex,text depth=.25ex] at (2.5,-.34) {#1};
\node[anchor=south] at (2.5,1.035) {\commentedit{Shared} worsens #3/#4 inputs};
\draw[globalcurve] (1.65,.42) -- (2.15,.42);
\node[anchor=west] at (2.25,.42) {\commentedit{Shared}};
\draw[inputcurve] (1.65,.22) -- (2.15,.22);
\node[anchor=west,align=left] at (2.25,.22) {Q1\\(own-input KL)};
\draw[white,fill=black,line width=.7pt] (0,{#3/#4}) circle (3pt);
\end{tikzpicture}%
}
\begin{minipage}{.49\textwidth}\centering
\begin{tikzpicture}[x=.88cm,y=3.6cm,font=\small,
 globalcurve/.style={black,line width=.9pt},
 inputcurve/.style={blue!60!black,densely dashed,line width=.9pt}]
\fill[black!6] (-1,0) rectangle (0,1);
\foreach \y/\label in {0/0,.25/25,.5/50,.75/75,1/100} {
  \draw[black!15] (-1,\y) -- (6,\y);
  \node[anchor=east] at (-1.08,\y) {\label};
}
\draw[dotted] (0,0) -- (0,1);
\input{figures/global_ecdf_paths}
\draw (-1,0) -- (6,0);
\draw (-1,0) -- (-1,1);
\foreach \x in {-1,0,1,2,3,4,5,6} {
  \draw (\x,0) -- (\x,-.02);
  \node[anchor=north] at (\x,-.025) {\x};
}
\node[rotate=90] at (-1.9,.5) {Cumulative share (\%)};
\node at (2.5,-.19) {Reduction in absolute effect gap (logit)};
\node[font=\small\bfseries,text height=1.5ex,text depth=.25ex] at (2.5,-.34) {(a) 9B: word-split selection};
\node[anchor=south] at (2.5,1.035) {\commentedit{Shared} worsens \GlobalNineWorsened/\GlobalNineInputs inputs};
\draw[globalcurve] (1.65,.42) -- (2.15,.42);
\node[anchor=west] at (2.25,.42) {\commentedit{Shared}};
\draw[inputcurve] (1.65,.22) -- (2.15,.22);
\node[anchor=west,align=left] at (2.25,.22) {Q1\\(own-input KL)};
\draw[white,fill=black,line width=.7pt] (0,{\GlobalNineWorsened/\GlobalNineInputs}) circle (3pt);
\end{tikzpicture}%

\end{minipage}\hfill
\begin{minipage}{.49\textwidth}\centering
\begin{tikzpicture}[x=.88cm,y=3.6cm,font=\small,
 globalcurve/.style={black,line width=.9pt},
 inputcurve/.style={blue!60!black,densely dashed,line width=.9pt}]
\fill[black!6] (-1,0) rectangle (0,1);
\foreach \y/\label in {0/0,.25/25,.5/50,.75/75,1/100} {
  \draw[black!15] (-1,\y) -- (6,\y);
  \node[anchor=east] at (-1.08,\y) {\label};
}
\draw[dotted] (0,0) -- (0,1);
\input{figures/global_ecdf_2b_paths}
\draw (-1,0) -- (6,0);
\draw (-1,0) -- (-1,1);
\foreach \x in {-1,0,1,2,3,4,5,6} {
  \draw (\x,0) -- (\x,-.02);
  \node[anchor=north] at (\x,-.025) {\x};
}
\node[rotate=90] at (-1.9,.5) {Cumulative share (\%)};
\node at (2.5,-.19) {Reduction in absolute effect gap (logit)};
\node[font=\small\bfseries,text height=1.5ex,text depth=.25ex] at (2.5,-.34) {(b) 2B: block-held-out selection};
\node[anchor=south] at (2.5,1.035) {\commentedit{Shared} worsens \GlobalTwoWorsened/\GlobalTwoInputs inputs};
\draw[globalcurve] (1.65,.42) -- (2.15,.42);
\node[anchor=west] at (2.25,.42) {\commentedit{Shared}};
\draw[inputcurve] (1.65,.22) -- (2.15,.22);
\node[anchor=west,align=left] at (2.25,.22) {Q1\\(own-input KL)};
\draw[white,fill=black,line width=.7pt] (0,{\GlobalTwoWorsened/\GlobalTwoInputs}) circle (3pt);
\end{tikzpicture}%

\end{minipage}
\caption{\textbf{Full-range counterpart to Figure~\ref{fig:global_distribution}.}
The wider axis reveals the positive tail clipped in that figure, with
unchanged cohorts and equal input weights. Reductions are in logits;
negative values denote larger gaps from \commentedit{the encoder-based reference intervention $E$}, and dots at zero mark the \commentedit{shared-coefficient}
worsening fraction. All 11 failed 9B Q1 searches remain included.
The panels retain their distinct selection protocols, and Q1 uses additional
evaluation-input information. Curves are descriptive, without confidence intervals.}
\label{fig:global_fullrange}
\end{figure*}

%% file: sections/h_llama_details.tex
\raggedbottom
\section{\frank{Appendix: Llama Configurations and Prospective Confirmation}}
\label{app:llama}

The two Llama configurations use the same base model and layer but different
SAEs and independently selected groups. They are not separate model-family
replications. This appendix records the assets, selection history and
denominators behind the main-text comparisons.

\subsection{Assets and Inference Coordinates}

\begingroup\raggedright
\frank{The model and tokenizer are} \nolinkurl{meta-llama/Llama-3.1-8B}.
The 128K SAE is \nolinkurl{Llama3_1-8B-Base-L15R-32x/checkpoints/final.safetensors}
from \nolinkurl{OpenMOSS-Team/Llama3_1-8B-Base-LXR-32x}.
The 32K SAE is the corresponding \nolinkurl{Llama3_1-8B-Base-L15R-8x/checkpoints/final.safetensors}
from \nolinkurl{OpenMOSS-Team/Llama3_1-8B-Base-LXR-8x}.
\par\endgroup

\frank{Both SAEs are applied to the residual output of zero-indexed block 15.} The archived
configuration widths are 131,072 and 32,768. Raw checkpoint tensors are
not directly used as inference-coordinate parameters. The adapter casts
stored tensors to FP32, folds dataset activation scaling and then normalizes
decoder columns, applying the compensating scaling to encoder rows and
biases. \frank{This follows the authors' loading procedure at the recorded version;} the conversion
parameters and support calculation were checked against those operations.
The JumpReLU thresholds are 0.330078125 and 0.35546875. Encoding does not
subtract decoder bias. Model forwards use an FP16 body, eager attention
and an FP32 output head. Patch positions, \commentedit{reconstruction residual} preservation and output
metrics follow \frank{Section~\ref{sec:protocol}}.

The 128K group IDs are
3478, 5946, 7489, 9836, 23366, 47994, 59024, 60208, 68163, 85257, 92087
and 108136. The 32K group IDs are
11664, 14236, 16426, 18776, 21286, 21734, 23413, 25319, 25943, 28323,
30921 and 32760.

\subsection{Selection and Eligible Cohorts}

Both dictionaries use the same 128-pair development inventory, built from
64 subject--attractor lemma pairs. One lexical block fixes both lemmas and
\emph{near}, varying subject and attractor number to give two pairs and four
inputs. Lemmas do not repeat across blocks or roles within a panel.
Cases label each recipient's subject and attractor number. Reconstruction
qualification is dictionary-specific, so the discovery rows can differ.

\frank{Discovery uses the normalized-margin utility $U(G)$ in
Appendix~\ref{app:group_utility}.}
It measures all active singleton contrasts, then forms a 64-feature shortlist
from the top 16 singleton utilities in each case, filling remaining places
by overall utility. A group qualifies if it transfers on at least half the
development inputs, transfers in at least 8 blocks, and has a transfer
and positive mean margin decrease in every case. At each greedy step,
proposals are ranked by qualification, number of covered cases, worst-case
utility, overall utility and feature IDs, in that order. Final selection
among qualified proposals maximizes overall utility, breaking ties by smaller
group size and then feature IDs. This rule can select a larger group after
a smaller one has already qualified.

The 128K six-feature search did not meet the development efficacy rule.
A subsequent, explicitly extended search allowed larger groups without
lowering that rule. This was a bounded singleton/greedy procedure, not an
exhaustive search over all subsets. \frank{The extension continued the existing
six-feature path and shortlist through size 12, evaluating 333 additional
joint proposals.} The resulting fixed twelve-feature group underwent separate
lexical validation on 128 pairs with disjoint development/validation word
forms. Validation uses the same 50\% group-transfer threshold and per-case
and eight-block breadth requirements. The native-body and decoded-body donor
controls defined in Appendix~\ref{app:gender_materials} must meet the breadth
requirements as well.

\frank{After validation, an initial candidate-qualification panel failed case
coverage. We record this stop separately from the subsequent 384-pair
expansion. The expansion retained the group, construction, qualification
thresholds and calibration limits, and did not pool old and new counts.}
The expansion's 192 lexical blocks were assigned before tokenization or
new model outputs. Each again contains one subject lemma and one attractor
lemma with both numbers. A fixed deterministic ordering selects 192 authored subject
nouns and 192 attractors after excluding word forms used in the Llama
development, initial validation and backup panels. No forms repeat across
blocks within the expansion. Some nouns originated in earlier project
inventories, so lexical separation here is relative to these Llama panels.

\frank{The expansion yields 600 aligned inputs out of 768 assigned,
and 98 bilateral native-qualified pairs.} Alternative qualification retains
94 NNLS pairs in 72 blocks and 62 candidate pairs in 50 blocks. Their union
has 95 pairs; their common intersection has 61 pairs in 49 blocks.

For 32K, a separate frozen development search selects twelve features under
the same coverage-aware rule, searching from singletons through size 12.
The greedy path first meets efficacy at size 5;
the twelve-feature result is not evidence that fewer features could not work.
Its validation panel assigns a new list of 64 subject--attractor lemma pairs,
excluding word forms from both dictionaries' development and the initial
128K validation panel. It preserves the two-pair, four-input block
construction and the validation efficacy rule above.
Held-out validation retains 58 of 128 source pairs and yields 99 encoder
transfers on 116 inputs. Subsequent qualification retains 56 NNLS pairs
in 39 blocks and 53 candidate pairs in 37 blocks. The candidate cohort is
contained in the NNLS cohort. Group efficacy is already observed when the
alternative-code calibration protocol is fixed; alternative effects are not.

\subsection{Calibration and Inference}
\label{app:llama_inference}

Both protocols specify \commentedit{full-output} and answer-excluded KL selection
before alternative effects, on each code's own eligible cohort. The grid is
$0,0.25,\ldots,2$, with at most 12 refinements and the same tolerance as the
Gemma comparisons. The original protocols add no extra bracket, cap expansion
or effect-based choice after a failed match. The later range continuation is
reported separately in Appendix~\ref{app:range}, preserving these original
results. The candidate's paired reduction on Q1-matched inputs is primary;
NNLS and the second selector retain their
secondary roles. Coverage requires at least 80\% retention and 8 matched
inputs in each original case, and at least 12 matched blocks overall.
Both candidate configurations fail this conjunction under both selectors.

Balanced summaries give equal weight to cases and to represented blocks
within each case. Conditional intervals use 10,000 lexical-block bootstrap
replicates with seed 20260921. Pooled descriptive means in
Table~\ref{tab:llama_calibration} instead weight inputs equally.
No interval assumes independent inputs or pools the two dictionaries.
For Q1-matched candidates, the balanced reductions have paired 95\%
block-bootstrap intervals wholly above zero in both configurations
(Table~\ref{tab:supplementary_intervals}). This
evidence concerns matched subsets because both fail the coverage requirement.

Positionwise norm matching passes on all 188 inputs in 128K and 109 of
112 in 32K. The full 32K positionwise-attempt cohort is retained in the transfer
counts, whereas Table~\ref{tab:positionwise} uses only the 109 inputs matched by both norm controls.
Table~\ref{tab:practical} uses all 112 inputs matched by total-norm matching and Q1. Undefined or out-of-tolerance norm
controls do not become matched observations by inclusion in a total.

\frank{Table~\ref{tab:llama_calibration} retains failed attempts in the
full-cohort columns under both selectors.}

\begin{table*}[t]
\centering
\begin{tabular}{llcrrrrl}
\toprule
\multicolumn{8}{c}{Pooled mean absolute gap (logit)} \\
\midrule
Config. / code & Selector & \frank{\shortstack{Matched /\\attempted}} & \multicolumn{2}{c}{Matched only} & \multicolumn{2}{c}{All attempts} & Coverage \\
 & & & Before & After & Before & Final & \\
\midrule
128K / NNLS & Q1 & 188/188 & 0.166 & 0.016 & 0.166 & 0.016 & Pass \\
 & Q2 & 188/188 & 0.166 & 0.023 & 0.166 & 0.023 & Pass \\
32K / NNLS & Q1 & 112/112 & 0.159 & 0.012 & 0.159 & 0.012 & Pass \\
 & Q2 & 112/112 & 0.159 & 0.014 & 0.159 & 0.014 & Pass \\
\midrule
128K / candidate & Q1 & 92/124 & 1.319 & 0.130 & 1.559 & 0.270 & Fail \\
 & Q2 & 91/124 & 1.318 & 0.174 & 1.559 & 0.302 & Fail \\
32K / candidate & Q1 & 75/106 & 1.234 & 0.041 & 1.494 & 0.197 & Fail \\
 & Q2 & 76/106 & 1.233 & 0.051 & 1.494 & 0.205 & Fail \\
\bottomrule
\end{tabular}
\caption{\frank{Earlier Llama cohorts: pooled mean absolute effect gaps from \commentedit{the encoder-based reference intervention $E$}
in logits. Matching counts use each code's own qualified cohort. Before/after
columns compare identical inputs within a row; matched membership can
differ between selectors. All attempts retain failed searches at their best
measured point, without calling them calibrated. Coverage requires the
prespecified per-case retention and support conditions. These descriptive
means are distinct from the original case/block-balanced estimates.}}
\label{tab:llama_calibration}
\end{table*}

\frank{Among unmatched Llama candidates, initial pooled mean gaps are 2.248 logit
in 128K and 2.124 in 32K. Both exceed the respective matched-subset means
in Table~\ref{tab:llama_calibration}.}

\subsection{Numerical Verification}
\label{app:llama_replay}

Both configurations have complete independent replays, including failed
searches. Final model outputs, applied patches, measured search points and
recomputed analyses agree. The compact records preserve output verification
summaries rather than persistent copies of every full array.

\subsection{Prospective NNLS Confirmation}
\label{app:confirmation}

\begin{table*}[t]
\centering
\input{tables/confirmation_controls}
\caption{\textbf{New-material Llama-128K controls on common inputs.}
Q1 preserves \commentedit{the transfer decisions of the encoder-based reference intervention $E$} and leaves the smallest observed
mean gap among these controls. Every measured matching control matches all
\ConfirmationInputs{} inputs. Gaps are in logits; balanced values equally
weight cases and blocks within case, whereas pooled values weight inputs
equally. Transfers and losses/gains relative to $E$ are unweighted counts.
The separated Q2 row is a follow-up after inspection of the confirmation,
on the same inputs. This cohort remains separate from the earlier 188-input panel.}
\label{tab:confirmation_controls}
\end{table*}

\begin{table}[t]
\centering
\input{tables/confirmation_cases}
\caption{New-material confirmation by recipient subject/attractor number;
$n$ counts inputs, all matched by both primary controls.
The mean total-norm minus Q1 gap is positive in every case. Contrasts are
in logits, with equal block weights within each case.}
\label{tab:confirmation_cases}
\end{table}

This panel tests whether the coefficient-count-matched NNLS comparison
persists on new material within the selected Llama-128K configuration.
We retain its model, SAE, twelve-feature group, inference coordinates and
patch domain. Before tokenizer inspection, we fix a hash-ordered selection
from authored subject and attractor noun lists, yielding
\ConfirmationSourcePairs{} sentence pairs in \ConfirmationSourceBlocks{}
lexical blocks. Each block uses one subject lemma and one attractor lemma,
with both numbers represented; no noun is shared across blocks or roles.
We exclude singular and plural forms from the specified Llama development,
initial validation, 32K validation, 128K expansion and original body-group
inventories, and check exact prefixes against the saved Llama panels.
This separation is relative to those inventories, not all project material
or model pretraining.

Token alignment retains \ConfirmationAlignedPairs{} pairs. Native behavior
and encoder reconstruction retain \ConfirmationNativePairs{} pairs, after
which complete-code NNLS qualification retains \ConfirmationPairs{} pairs
and \ConfirmationBlocks{} blocks. Eligibility uses the bilateral checks in
Appendix~\ref{app:eligibility}, without observing group-patch effects or
selecting on encoder transfer. No rejected pair is replaced. The resulting
\ConfirmationInputs{} inputs have \ConfirmationEffectiveBlocks{} effective
blocks under the final case/block weights. Here the effective count is
$1/\sum_b w_b^2$, where $w_b$ is block $b$'s total weight in the balanced
mean and the weights sum to one. This count measures weight concentration,
without establishing independence between blocks.

\paragraph{Design and uncertainty.}
\frank{The confirmation's prespecified primary contrast is the NNLS effect
gap under total-norm matching minus its gap under full-output KL matching
(Q1), using inputs matched by both controls.} We first average within block and case, then
equally across represented blocks within each case and across the four cases.
Planning uses the earlier Llama cohort solely to estimate variability, with
a smaller planning alternative of 0.07 logit and a 1.5-fold inflation of
the estimated block-level standard deviation. At 64 effective blocks, the
normal approximation meets the planned half-width of at most 0.05 logit and
power of at least 0.80 for a positive lower 95\% bound. These calculations
depend on independent-block variation; 0.07 is a planning alternative,
not an equivalence margin or a minimum meaningful effect.

Both qualification stages and the primary matched subset must retain at
least 64 effective blocks and 32 blocks per case. Each primary control must
match at least 90\% overall and 80\% in every case. Confirmation additionally
requires a positive lower endpoint of the primary 95\% interval and a
half-width no greater than 0.05 logit. We use \ConfirmationDraws{} paired
lexical-block bootstrap draws with seed 20260926, retaining both transfer orientations together,
and report an interval only if at least 9,900 draws retain all cases.
Missing-case draws are not replaced. Total-norm matching retains its 0.001
relative tolerance, and Q1 retains the nine-point $[0,2]$ grid, at most twelve
refinements and the KL tolerance in Appendix~\ref{app:search}.

\paragraph{Outcomes and scope.}
All three controls actually match all qualified inputs, so matched-subset
and all-attempt summaries coincide here. All primary bootstrap draws retain
the required cases, and every prespecified confirmation requirement passes.
Table~\ref{tab:confirmation_controls} separates balanced and pooled gaps;
Table~\ref{tab:confirmation_cases} gives the case means. Deleting any one
block leaves the primary estimate positive, between
\ConfirmationDeletionMin{} and \ConfirmationDeletionMax{} logit.
Positionwise-minus-total-norm gaps and transfer changes are secondary reports.
We do not add a confirmatory test of positionwise matching against Q1.
No candidate is measured on this panel. The later Q2 comparison is described
separately below.

Independent execution repeats native and encoder qualification, NNLS fitting
and qualification, and the full effects/calibration search, including upstream
failures. All arrays and scientific records reproduce exactly.
The new panel provides prospective material
confirmation, conditional on a setting chosen using earlier outcomes; it does
not replace the original prespecified candidate comparison or broaden the model
and task scope.

\paragraph{Illustrative input.}
Figure~\ref{fig:overview} uses the singular-recipient input from the
schoolgirl/iceberg block. We selected it after inspecting outcomes to
illustrate an encoder transfer lost under total-norm matching and retained
under Q1. Its recipient-answer minus donor-answer margin,
$\operatorname{logit}(\textit{is})-\operatorname{logit}(\textit{are})$,
is $\OverviewNativeMargin$ before patching and $\OverviewEncoderMargin$
under the encoder intervention. The margins after scaling the same NNLS
patch are $\OverviewTotalMargin$ under total-norm matching and
$\OverviewQoneMargin$ under Q1, all in logits. These two-token rankings
describe the transfer criterion, not generated completions.
The figure's aggregate counts use the complete qualified panel.

\subsection{Answer-Excluded Follow-Up on the Confirmation Panel}
\label{app:confirmation_q2}

\begin{table}[t]
\centering
\input{tables/confirmation_q2_contrasts}
\caption{\textbf{Q2 follow-up contrasts on common Llama inputs.}
Answer exclusion retains a reduction but leaves a larger \commentedit{residual gap} than Q1.
All \FollowupQtwoInputs{} inputs match the compared controls. Estimates and
pointwise 95\% block-bootstrap intervals are in logits, with equal case
weights and equal block weights within case.}
\label{tab:confirmation_q2}
\end{table}

To test whether the reduction persists after answer exclusion, we specify
a Q2 follow-up after inspecting the original confirmation. It uses the same
\FollowupQtwoInputs{} NNLS-qualified inputs, keeping the model, group,
patches, precision and search bounds fixed. No new material or qualification
is introduced. The primary follow-up contrast is the unscaled-NNLS minus Q2
effect gap; Q2 minus Q1 and total-norm minus Q2 are separate supporting
comparisons. \frank{Each uses inputs meeting its specified matching criteria} and retains an
all-attempt summary. Coverage and block-support thresholds follow the
original confirmation, \frank{but the extension adds no precision requirement.}

Q2 matches every input in every case, so all three contrasts use the same
cohort (Table~\ref{tab:confirmation_q2}). The unscaled-NNLS minus Q2 contrast is
positive in every case and after deletion of any one block. The larger
\commentedit{residual gap} relative to Q1 and the lost transfers in
Table~\ref{tab:confirmation_controls} show that removing direct answer
probabilities retains a reduction without establishing equivalence between
the selectors. Intervals use \ConfirmationDraws{} paired lexical-block
bootstrap draws with seed 20260926 under the original weighting, with the valid-draw rule above.
All draws retain the required cases. These are pointwise conditional intervals,
not simultaneous coverage across the three contrasts.

Independent execution repeats the complete Q2 search and reproduces all
saved arrays and scientific records exactly.


%% file: tables/confirmation_controls.tex
\begin{tabular}{@{}lrrrrrr@{}}
\toprule
Intervention & Inputs & Balanced gap & Pooled gap & Transfers & Lost & Gained \\
\midrule
\commentedit{Reference $E$} & 304 & 0.000 & 0.000 & 175 & 0 & 0 \\
Unscaled NNLS & 304 & 0.213 & 0.207 & 181 & 3 & 9 \\
Total-norm matching & 304 & 0.181 & 0.176 & 165 & 11 & 1 \\
Positionwise norm matching & 304 & 0.134 & 0.130 & 167 & 10 & 2 \\
Full-output KL (Q1) & 304 & 0.021 & 0.020 & 175 & 0 & 0 \\
\midrule
\multicolumn{7}{l}{Follow-up after inspection of the confirmation results} \\
Answer-excluded KL (Q2) & 304 & 0.029 & 0.028 & 173 & 2 & 0 \\
\bottomrule
\end{tabular}

%% file: tables/confirmation_cases.tex
\begin{tabular}{lrr}
\toprule
Case & $n$ & Contrast \\
\midrule
Singular / singular & 65 & 0.195 \\
Singular / plural & 87 & 0.122 \\
Plural / singular & 65 & 0.185 \\
Plural / plural & 87 & 0.138 \\
\bottomrule
\end{tabular}

%% file: tables/confirmation_q2_contrasts.tex
\begin{tabular}{@{}lrl@{}}
\toprule
Effect-gap contrast & \frank{Mean} & 95\% interval \\
\midrule
Unscaled NNLS $-$ Q2 & $0.184$ & $[0.149, 0.223]$ \\
Q2 $-$ Q1 & $0.008$ & $[0.005, 0.011]$ \\
Total norm $-$ Q2 & $0.152$ & $[0.124, 0.183]$ \\
\bottomrule
\end{tabular}

%% file: sections/d_registered_stops.tex
\section{\frank{Appendix: Selection History and Stopped Comparisons}}
\label{app:stops}

\begin{table*}[t]
\centering
\begin{tabular}{p{.23\textwidth}p{.34\textwidth}p{.34\textwidth}}
\toprule
\raggedright Comparison & \raggedright Evidence status & \raggedright Inferential scope\tabularnewline
\midrule
\raggedright 9B lexical candidate & \raggedright Prespecified primary; effects unmeasured when specified &
\raggedright Coverage passes; planned precision unmet\tabularnewline
\addlinespace[2pt]
\raggedright 2B and Llama SVA extensions & \raggedright Specified after group qualification, before
alternative effects & \raggedright 2B matches fully; Llama candidate coverage fails\tabularnewline
\addlinespace[2pt]
\raggedright Gender & \raggedright Selected group validated before code comparison &
\raggedright Partial extension; Q2 reverses the NNLS pattern\tabularnewline
\addlinespace[2pt]
\raggedright Single-scale controls & \raggedright Prespecified follow-ups on exposed panels &
\raggedright Paired comparisons, with unavailable intervals reported\tabularnewline
\addlinespace[2pt]
\raggedright New Llama-128K NNLS panel & \raggedright Materials and analysis specified before tokenization
or group effects & \raggedright Prospective material confirmation; all support, coverage
and precision requirements pass\tabularnewline
\addlinespace[2pt]
\raggedright Q2 on the new Llama panel & \raggedright Specified after inspection of the original
confirmation results & \raggedright Follow-up reduction; larger \commentedit{residual gap} and transfer
losses than Q1 on the same inputs\tabularnewline
\addlinespace[2pt]
\raggedright Candidate search range & \raggedright Specified after inspection of failed search traces &
\raggedright Descriptive continuation to $\alpha=4$; original coverage decisions unchanged\tabularnewline
\addlinespace[2pt]
\raggedright Budgets and coefficient transfer & \raggedright Follow-ups on exposed Gemma SVA panels &
\raggedright Finite budgets and coefficient grid only\tabularnewline
\bottomrule
\end{tabular}
\caption{Evidence status and scope. Exposition starts with practical NNLS
comparisons without replacing the prespecified primary endpoint. The new Llama
panel provides prospective material confirmation separately from exposed-panel
follow-ups.}
\label{tab:evidence_status}
\end{table*}

Table~\ref{tab:evidence_status} distinguishes confirmation from follow-up
evidence. The original lexical confirmation targeted transfer rates, but only
252 of 1,984 source pairs jointly qualified. Under the specified block-variance
scenario, \frank{the retained subset gave a planning half-width of 0.1053} against the required
0.10, so the study stopped before group-effect measurement. This design
calculation limited the attainable precision without providing an observed
effect estimate. The later calibration study used the same 252 pairs to ask
how much Q1 reduces paired absolute effect gaps. Its contrast and bounded
search were specified before group effects were inspected. It supplies
evidence for that separate question, without satisfying the original
transfer-rate criterion. Its own primary estimate also missed its planned
precision, as reported with the main result.

The earlier 83-pair calibration panel failed one case-coverage requirement,
and the low-strength random-group comparison failed full-panel coverage.
Their measurements support only the reported retained-subset comparisons.
\frank{These comparisons are part of a longer adaptive research process,
so the selected settings should not be read as a representative sample of
available interventions. The earlier public-group qualification failure is
detailed in Appendix~\ref{app:secondmodel}.}

The cross-family extension also includes selection and availability failures.
A bounded Llama-128K six-feature search failed, leading to an extended
group-size limit. The initial candidate cohort for the selected group then
failed case coverage. A separately specified expansion supplied the
calibration panel reported here, with the failed initial cohort kept separate.
The 32K dictionary was searched and validated separately. Both final Llama
candidate calibration cohorts still fail coverage, despite fully matched
NNLS controls (Appendix~\ref{app:llama}). Those coverage failures restrict
candidate conclusions to the reported subsets. The bounded searches leave
the availability of compact groups in unselected models unresolved.

\subsection{\frank{Earlier Public-Group Qualification}}
\label{app:secondmodel}

\frank{A separate publicly sourced two-feature candidate in Gemma-2-2B did not
qualify for the same type of comparison. Of 128 native inputs, 125 had
the required correct answer. None passed the reconstruction KL cap of 0.02,
leaving zero eligible pairs out of 64.
The failure concerns reconstruction eligibility, not the causal efficacy of
the feature pair. This stopped candidate is distinct from the subsequently
selected six-feature 2B group in Appendix~\ref{app:2b}; the failed public pair
is not included as a causal replication.}

\subsection{Searches for an Additional Phenomenon}

Table~\ref{tab:phenomenon_stops} separates upstream availability failures
from tests of the calibration claim. A native check establishes answer
behavior; reconstruction qualification checks dictionary fidelity; the donor
control tests whether the patch domain can carry the behavior before selecting
a compact feature group. Failing any of these prerequisites prevents the
downstream code comparison. A bounded group search establishes only what that
search found, not the absence of every effective group.

\noindent\begin{minipage}{\columnwidth}
\centering
\begin{tabular}{@{}p{.46\columnwidth}p{.46\columnwidth}@{}}
\toprule
\raggedright Candidate phenomenon & \raggedright Stopping stage\tabularnewline
\midrule
\raggedright Anaphor agreement & \raggedright Compact-group discovery\tabularnewline
\raggedright Inanimate reflexive agreement & \raggedright Compact-group discovery\tabularnewline
\raggedright Auxiliary-form selection & \raggedright Donor control\tabularnewline
\raggedright Pronoun case & \raggedright Reconstruction qualification\tabularnewline
\raggedright Indirect-object identification & \raggedright Reconstruction qualification\tabularnewline
\raggedright Indirect-object follow-up & \raggedright Native donor control\tabularnewline
\bottomrule
\end{tabular}
\captionof{table}{Stopped additional-phenomenon searches. None reached a test of
calibration, so these failures restrict the available settings without
providing positive or negative calibration results. Reflexive gender agreement did reach calibration; its
adverse Q2 result is reported in Section~\ref{sec:excluded}, separately from
these upstream stops.}
\label{tab:phenomenon_stops}
\end{minipage}


%% file: sections/f_related_tests.tex
\section{\frank{Appendix: Relation to Other Explanation Tests}}
\label{app:related_tests}

\paragraph{Syntactic competence and causal involvement.}
Controlled agreement tests probe whether predictions respect syntactic
structure despite distracting nouns or reduced semantic cues
\citep{gulordava-etal-2018-colorless}. BLiMP extends minimal-pair evaluation
across a wider range of grammatical phenomena
\citep{warstadt-etal-2020-blimp-benchmark}. Our targeted next-token contrasts
do not provide that breadth. Studies of number-tracking units in recurrent
models \citep{lakretz-etal-2019-emergence} and causal mediation in transformers
\citep{finlayson-etal-2021-causal} instead investigate internal contributions
to agreement. Related intervention methods test mediation of gender bias
\citep{vig2020investigating} or remove decodable information to assess its
behavioral role \citep{elazar-etal-2021-amnesic}. These studies motivate
distinguishing information present in a representation from information used
in a prediction. Our experiment asks the narrower question of comparability
between interventions on the same group.

\paragraph{Discovering an explanation versus comparing its effects.}
The indirect-object-identification circuit study evaluates a proposed circuit
using faithfulness, completeness and minimality criteria
\citep{wang2023ioi}. Automated circuit discovery searches for relevant
computational subgraphs \citep{conmy2023automated}, and AtP* addresses errors
in gradient-based approximations used to localize components
\citep{kramar2024atp}. Distributed alignment search learns a correspondence
between proposed causal variables and neural representations
\citep{geiger2024alignments}. We neither rediscover a circuit under each code
nor optimize a new causal alignment. The feature group and decoder stay fixed,
and the reported effects come from executed interventions, not gradient
approximations. Calibration agreement therefore does not establish that this
group is a complete or minimal mechanism.

\paragraph{Robustness is not correctness.}
Sanity checks for saliency maps test whether explanations depend on the model
and data they purport to explain \citep{adebayo2018sanity}.
Explanation-robustness analyses ask how explanations change under small input
perturbations \citep{alvarez2018robustness}. Our varying object is the code
and resulting patch for a fixed input, rather than the input or trained model.
The common lesson is to specify what a stability test holds fixed and what it
can establish. Two interventions agreeing with each other is not an external
criterion of explanatory correctness, even when both change the target answer.

Variable importance can also vary across well-performing prediction models,
as studied through model-class reliance and variable-importance clouds
\citep{fisher2019allmodels,dong2019variable}. Our decoder instead stays fixed
while its codes vary. This connection
motivates specifying the object held fixed in a comparison, without equating
the two sources of variation.

%% file: sections/e_artifact_guide.tex
\section{\frank{Appendix: Reproducibility and Artifact Availability}}
\label{app:artifact}

The numerical checks address three different questions: whether the reported
statistics follow from the saved measurements, whether the saved patches
implement the stated comparison, and whether another execution reproduces the
model outputs. Passing one check does not establish the others.

\subsection{Recomputing the Reported Comparisons}

The result records preserve pair, input, case and block identifiers,
matching decisions, selected coefficients and measured effects. The lexical
and answer-excluded comparisons each account for all 504 attempted inputs;
the global-coefficient evaluation accounts for its 346 evaluation inputs.
Failed matches remain in these records. For the lexical primary estimate,
they are excluded by the stated matching rule, whereas the answer-excluded
primary comparison retains their measured final attempts. The Gemma-2B setting
has 64 candidate and 62 NNLS inputs, with \frank{separate matching decisions} at
each budget (Appendix~\ref{app:response_details}).

Recomputation from these scalar records checks the case-balanced means,
paired transfer counts, conditional intervals and block-deletion summaries.
Recomputing the effects and KL values themselves requires the archived native
and intervened full-vocabulary logits. For Q2, this
check removes the two specified answer-token entries before normalization.
It cannot be reproduced from the answer logits alone or from a truncated list
of the most likely tokens. \frank{The newer 2B response-extension records retain output verification summaries}
instead of full logit arrays; arbitrary new output-statistic checks require model
re-execution there.

For the new Llama confirmation and its Q2 follow-up, recomputation must
preserve the separate evidence status of the two analyses even though they
share inputs. \frank{Each comparison uses inputs meeting its specified matching criteria},
with failed searches retained in the all-attempt summaries.
The bootstrap resamples lexical blocks jointly across the paired controls.
The released records must therefore include block and case identifiers as
well as effects, \frank{matching decisions} and transfer outcomes. Aggregate means alone
cannot reproduce these intervals or the paired gains and losses.

The lexical review also reruns the coefficient-selection logic using the saved
KL search trace. This checks which coefficient the recorded search would
select. It does not independently regenerate the model outputs at intermediate
search points. These output-based checks require no model forward pass, but
remain conditional on the archived outputs.

\subsection{Checking the Code and Physical Patch}

Checking code eligibility additionally requires the decoder, \commentedit{native
activations}, coefficients and support records. With those inputs, linear
algebra suffices to check nonnegativity, support membership and reconstruction
error. Behavioral reconstruction eligibility additionally uses the native and
reconstructed output distributions. The construction and numerical conventions
are specified in Appendix~\ref{app:construction}.

The saved physical patches provide a separate check on what was inserted into
the model. The initial encoder and candidate patches must meet the stated
per-position norm constraint after materialization. For calibration, the
selected coefficient and candidate patch determine the rescaled patch,
including its execution-precision rounding. Verifying that patch does not make
it an admissible sparse code: the calibration controls are physical
interventions, as defined in \frank{Section~\ref{sec:protocol}}.

\subsection{Reproducing Model Outputs}

Model replay requires the pinned model, tokenized inputs,
intervention positions, stored patches and numerical execution configuration.
\commentedit{Archived} and replayed model outputs are checked for equality. The
original lexical replays cover the native, reconstruction and final intervention
outputs described in Appendix~\ref{app:replay}, including unsuccessful matching
attempts. They reproduce final comparisons without rerunning adaptive search.
Later Gemma budget and coefficient-transfer replays cover the complete
measured searches (Appendix~\ref{app:gemma_replay}). Llama replay scope is
reported in Appendix~\ref{app:llama_replay}, with confirmation and Q2
follow-up checks in Appendices~\ref{app:confirmation} and~\ref{app:confirmation_q2}.

Equality of the replayed arrays establishes numerical repeatability for the
tested executions. It does not supply new lexical observations, validate the
block-independence assumption, or guarantee agreement on other hardware.
The scientific checks above additionally establish which interventions and
comparisons the records represent.

Acquiring the model and SAE weights is a separate prerequisite. Model access
must follow the official provider's terms, including gated access where
required. Reproduction requires the recorded repository revisions and SAE
checkpoint files, rather than whichever versions are current at download
time. \frank{The numerical configuration and measured cost of the new Llama
confirmation chain are specified in Appendix~\ref{app:confirmation_compute}.
General calibration-search costs are described in Appendix~\ref{app:compute}.}

\subsection{\frank{Llama Confirmation: Execution Environment and Measured Cost}}
\label{app:confirmation_compute}
\input{tables/reproducibility_metadata}

\frank{The new Llama-128K confirmation used one NVIDIA RTX A6000 GPU per
execution, with four allocated CPU cores and single-threaded numerical
libraries. The three original phases, comprising native qualification,
NNLS qualification and controlled effects, took \ReproConfirmationMinutes{}
minutes of application time in total. The Q2 follow-up took
\ReproQtwoMinutes{} minutes, and replaying all four phases on a different
machine took \ReproReplayMinutes{} minutes. Together these account for
\ReproSubsetHours{} single-GPU application hours. The timers include asset
loading and CPU work inside each phase, not only GPU kernel time.}

\frank{This subtotal excludes earlier cohorts, group discovery, stopped
searches, failed or interrupted attempts, CPU-only reviews and scheduler
overhead. A complete computational budget for the adaptive research process
has not been reconstructed. Historical scheduler accounting is unavailable,
and the retained application timers do not cover every failed or stopped
attempt. We therefore do not extrapolate this reproduction subtotal to a
project-wide estimate.}

\frank{The required package versions were verified before each Llama phase.
All phases checked PyTorch \ReproVersionTorch{}, Transformers
\ReproVersionTransformers{}, NumPy \ReproVersionNumpy{}, Safetensors
\ReproVersionSafetensors{} and Hugging Face Hub \ReproVersionHub{}.
Native qualification additionally checked Tokenizers
\ReproVersionTokenizers{}, and NNLS qualification and effect measurements
checked SciPy \ReproVersionScipy{}. The runs used an FP16 model body, FP32
output head and eager attention, with
deterministic PyTorch operations and TensorFloat-32 (TF32) disabled.
These versions describe this confirmation chain; they do not apply to
every earlier experiment.}

\subsection{Artifact Availability}

Source-to-table checks on the internal archives establish agreement between
manuscript values and reviewed records. They do not independently recompute
every interval or regenerate model outputs.
The accompanying CPU supplement contains the 570 attempted NNLS
strength-control inputs from the five earlier cohorts and the separate
304-input Llama confirmation cohort, including its later Q2 measurements.
It also provides 49 overlapping comparisons covering candidate Q1/Q2
results, shared coefficients, six KL budgets and the seven search-range
continuation cells. The records include sentence prefixes and token IDs,
pair/case/block identifiers, actual matching masks, scalar measurements,
pinned asset metadata and scientific excerpts of the internal plans.
These excerpts are not externally timestamped registrations.
A standalone checker recomputes the specified means, intervals and transfer
counts, reselects shared coefficients from recorded calibration KL values,
and checks retained scalar selections and budget intersections.
It has been tested from a clean extraction in a new Python environment
without the research archive.

This supplement verifies arithmetic on saved scalar measurements. It does
not regenerate logits, recompute \commentedit{full-output} KL or reconstruct every
upstream eligibility decision and stopped search. The package includes
retained scalar traces, including the recovered 9B Q2 records, but does not
reexecute model searches or check every supplemental diagnostic.
Model replay still depends on internal archived arrays and runners.
No model or SAE weights are redistributed. The supplement README and
coverage inventory identify the executable comparisons, checked fields,
dependencies and omissions.
The software and data archives accompany this preprint as arXiv ancillary
files; extract both into the same directory to run the checker.

\input{tables/asset_revisions}

\subsection{AI Assistance}
\label{app:ai_assistance}

AI agents assisted with research brainstorming, code implementation, and manuscript drafting and revision. The authors reviewed the AI-assisted contributions \commentedit{and conducted} numerical and reproducibility checks.

%% file: tables/reproducibility_metadata.tex
\def\ReproConfirmationMinutes{35.5}
\def\ReproQtwoMinutes{14.1}
\def\ReproReplayMinutes{40.4}
\def\ReproSubsetHours{1.50}
\def\ReproVersionTorch{2.9.1}
\def\ReproVersionTransformers{4.57.6}
\def\ReproVersionNumpy{2.4.6}
\def\ReproVersionScipy{1.17.1}
\def\ReproVersionSafetensors{0.8.0}
\def\ReproVersionHub{0.36.2}
\def\ReproVersionTokenizers{0.22.2}

%% file: tables/asset_revisions.tex
\subsection{Pinned Asset Revisions}
\label{app:asset_revisions}
The following revisions identify the model (including its tokenizer) and SAE
used in the five main configurations and the new Llama confirmation.
The CPU supplement also records checkpoint filenames and SHA-256 digests.
Access follows the provider terms described in the Ethics Statement.
Weights are not included.
\paragraph{Gemma-9B.}
\begingroup\raggedright
Model and tokenizer: \nolinkurl{google/gemma-2-9b-it}, revision\\
\texttt{11c9b309abf73637e4b6}\allowbreak\texttt{f9a3fa1e92e615547819}.
SAE: \nolinkurl{google/gemma-scope-9b-it-res}, revision\\
\texttt{e86af97a5b6fbbccca28}\allowbreak\texttt{ab654f2fda1b0768f770}.
\par\endgroup
\paragraph{Gemma-2B and gender.}
\begingroup\raggedright
Model and tokenizer: \nolinkurl{google/gemma-2-2b}, revision\\
\texttt{c5ebcd40d208330abc69}\allowbreak\texttt{7524c919956e692655cf}.
SAE: \nolinkurl{google/gemma-scope-2b-pt-res}, revision\\
\texttt{fd571b47c1c64851e9b1}\allowbreak\texttt{989792367b9babb4af63}.
\par\endgroup
\paragraph{Llama-128K.}
\begingroup\raggedright
Model and tokenizer: \nolinkurl{meta-llama/Llama-3.1-8B}, revision\\
\texttt{d04e592bb4f6aa9cfee9}\allowbreak\texttt{1e2e20afa771667e1d4b}.
SAE: \nolinkurl{OpenMOSS-Team/Llama3_1-8B-Base-LXR-32x}, revision\\
\texttt{336730e758fed3cb2273}\allowbreak\texttt{276703d836aa8659d293}.
\par\endgroup
\paragraph{Llama-32K.}
\begingroup\raggedright
Model and tokenizer: \nolinkurl{meta-llama/Llama-3.1-8B}, revision\\
\texttt{d04e592bb4f6aa9cfee9}\allowbreak\texttt{1e2e20afa771667e1d4b}.
SAE: \nolinkurl{OpenMOSS-Team/Llama3_1-8B-Base-LXR-8x}, revision\\
\texttt{8dbc1d85edfced43081c}\allowbreak\texttt{03c38b05514dbab1368b}.
\par\endgroup

%% file: main.bbl
\begin{thebibliography}{45}
\providecommand{\natexlab}[1]{#1}

\bibitem[{Adebayo et~al.(2018)Adebayo, Gilmer, Muelly, Goodfellow, Hardt, and
  Kim}]{adebayo2018sanity}
Julius Adebayo, Justin Gilmer, Michael Muelly, Ian Goodfellow, Moritz Hardt,
  and Been Kim. 2018.
\newblock \href
  {https://proceedings.neurips.cc/paper/2018/hash/294a8ed24b1ad22ec2e7efea049b8737-Abstract.html}
  {Sanity checks for saliency maps}.
\newblock In \emph{Advances in Neural Information Processing Systems},
  volume~31.

\bibitem[{Alvarez-Melis and Jaakkola(2018)}]{alvarez2018robustness}
David Alvarez-Melis and Tommi~S. Jaakkola. 2018.
\newblock \href {https://arxiv.org/abs/1806.08049} {On the robustness of
  interpretability methods}.
\newblock ICML Workshop on Human Interpretability in Machine Learning.
\newblock ArXiv:1806.08049.

\bibitem[{Barr et~al.(2013)Barr, Levy, Scheepers, and Tily}]{barr2013random}
Dale~J. Barr, Roger Levy, Christoph Scheepers, and Harry~J. Tily. 2013.
\newblock \href {https://doi.org/10.1016/j.jml.2012.11.001} {Random effects
  structure for confirmatory hypothesis testing: Keep it maximal}.
\newblock \emph{Journal of Memory and Language}, 68(3):255--278.

\bibitem[{Baumgartner et~al.(2025)Baumgartner, Spens, Burgess, and
  Manescu}]{baumgartner2025cognitive}
Caroline Baumgartner, Eleanor Spens, Neil Burgess, and Petru Manescu. 2025.
\newblock \href {https://arxiv.org/abs/2511.13371} {Cognitive maps in language
  models: A mechanistic analysis of spatial planning}.
\newblock CogInterp Workshop at NeurIPS.

\bibitem[{Belinkov(2022)}]{belinkov-2022-probing}
Yonatan Belinkov. 2022.
\newblock \href {https://doi.org/10.1162/coli_a_00422} {Probing classifiers:
  Promises, shortcomings, and advances}.
\newblock \emph{Computational Linguistics}, 48(1):207--219.

\bibitem[{Bricken et~al.(2023)Bricken, Templeton, Batson, Chen, Jermyn,
  Conerly, Turner, Anil, Denison, Askell, Lasenby, Wu, Kravec, Schiefer,
  Maxwell, Joseph, Hatfield-Dodds, Tamkin, Nguyen, McLean, Burke, Hume, Carter,
  Henighan, and Olah}]{bricken2023monosemanticity}
Trenton Bricken, Adly Templeton, Joshua Batson, Brian Chen, Adam Jermyn, Tom
  Conerly, Nick Turner, Cem Anil, Carson Denison, Amanda Askell, Robert
  Lasenby, Yifan Wu, Shauna Kravec, Nicholas Schiefer, Tim Maxwell, Nicholas
  Joseph, Zac Hatfield-Dodds, Alex Tamkin, Karina Nguyen, and 6 others. 2023.
\newblock \href
  {https://transformer-circuits.pub/2023/monosemantic-features/index.html}
  {Towards monosemanticity: Decomposing language models with dictionary
  learning}.
\newblock \emph{Transformer Circuits Thread}.

\bibitem[{Carbo and Nalisnick(2026)}]{carbo2026which}
Alessa Carbo and Eric Nalisnick. 2026.
\newblock \href {https://openreview.net/pdf?id=JiyytGKbA9} {Which sparse code?
  identifiability failures in {SAE} inference}.
\newblock Sci4DL Workshop at ICLR.

\bibitem[{Clark(1973)}]{clark1973language}
Herbert~H. Clark. 1973.
\newblock \href {https://doi.org/10.1016/S0022-5371(73)80014-3} {The
  language-as-fixed-effect fallacy: A critique of language statistics in
  psychological research}.
\newblock \emph{Journal of Verbal Learning and Verbal Behavior},
  12(4):335--359.

\bibitem[{Conmy et~al.(2023)Conmy, Mavor-Parker, Lynch, Heimersheim, and
  Garriga-Alonso}]{conmy2023automated}
Arthur Conmy, Augustine~N. Mavor-Parker, Aengus Lynch, Stefan Heimersheim, and
  Adri{\`a} Garriga-Alonso. 2023.
\newblock \href
  {https://proceedings.neurips.cc/paper_files/paper/2023/hash/34e1dbe95d34d7ebaf99b9bcaeb5b2be-Abstract-Conference.html}
  {Towards automated circuit discovery for mechanistic interpretability}.
\newblock In \emph{Advances in Neural Information Processing Systems},
  volume~36.

\bibitem[{Dong and Rudin(2019)}]{dong2019variable}
Jiayun Dong and Cynthia Rudin. 2019.
\newblock \href {https://arxiv.org/abs/1901.03209} {Variable importance clouds:
  A way to explore variable importance for the set of good models}.
\newblock ArXiv preprint.

\bibitem[{Elazar et~al.(2021)Elazar, Ravfogel, Jacovi, and
  Goldberg}]{elazar-etal-2021-amnesic}
Yanai Elazar, Shauli Ravfogel, Alon Jacovi, and Yoav Goldberg. 2021.
\newblock \href {https://doi.org/10.1162/tacl_a_00359} {Amnesic probing:
  Behavioral explanation with amnesic counterfactuals}.
\newblock \emph{Transactions of the Association for Computational Linguistics},
  9:160--175.

\bibitem[{Elhage et~al.(2022)Elhage, Hume, Olsson, Schiefer, Henighan, Kravec,
  Hatfield-Dodds, Lasenby, Drain, Chen, Grosse, McCandlish, Kaplan, Amodei,
  Wattenberg, and Olah}]{elhage2022superposition}
Nelson Elhage, Tristan Hume, Catherine Olsson, Nicholas Schiefer, Tom Henighan,
  Shauna Kravec, Zac Hatfield-Dodds, Robert Lasenby, Dawn Drain, Carol Chen,
  Roger Grosse, Sam McCandlish, Jared Kaplan, Dario Amodei, Martin Wattenberg,
  and Christopher Olah. 2022.
\newblock \href {https://arxiv.org/abs/2209.10652} {Toy models of
  superposition}.
\newblock ArXiv preprint arXiv:2209.10652.

\bibitem[{Finlayson et~al.(2021)Finlayson, Mueller, Gehrmann, Shieber, Linzen,
  and Belinkov}]{finlayson-etal-2021-causal}
Matthew Finlayson, Aaron Mueller, Sebastian Gehrmann, Stuart~M. Shieber, Tal
  Linzen, and Yonatan Belinkov. 2021.
\newblock \href {https://doi.org/10.18653/v1/2021.acl-long.144} {Causal
  analysis of syntactic agreement mechanisms in neural language models}.
\newblock In \emph{Proceedings of the 59th Annual Meeting of the Association
  for Computational Linguistics and the 11th International Joint Conference on
  Natural Language Processing (Volume 1: Long Papers)}, pages 1828--1843.
  Association for Computational Linguistics.

\bibitem[{Fisher et~al.(2019)Fisher, Rudin, and Dominici}]{fisher2019allmodels}
Aaron Fisher, Cynthia Rudin, and Francesca Dominici. 2019.
\newblock \href {https://jmlr.org/papers/v20/18-760.html} {All models are
  wrong, but many are useful: Learning a variable's importance by studying an
  entire class of prediction models simultaneously}.
\newblock \emph{Journal of Machine Learning Research}, 20(177):1--81.

\bibitem[{Gao et~al.(2025)Gao, Dupr{\'e}~la Tour, Tillman, Goh, Troll, Radford,
  Sutskever, Leike, and Wu}]{gao2025scaling}
Leo Gao, Tom Dupr{\'e}~la Tour, Henk Tillman, Gabriel Goh, Rajan Troll, Alec
  Radford, Ilya Sutskever, Jan Leike, and Jeffrey Wu. 2025.
\newblock \href
  {https://proceedings.iclr.cc/paper_files/paper/2025/hash/42ef3308c230942d223c411adf182c88-Abstract-Conference.html}
  {Scaling and evaluating sparse autoencoders}.
\newblock In \emph{The Thirteenth International Conference on Learning
  Representations}.

\bibitem[{Geiger et~al.(2021)Geiger, Lu, Icard, and Potts}]{geiger2021causal}
Atticus Geiger, Hanson Lu, Thomas Icard, and Christopher Potts. 2021.
\newblock \href
  {https://papers.nips.cc/paper/2021/hash/4f5c422f4d49a5a807eda27434231040-Abstract.html}
  {Causal abstractions of neural networks}.
\newblock In \emph{Advances in Neural Information Processing Systems},
  volume~34.

\bibitem[{Geiger et~al.(2024)Geiger, Wu, Potts, Icard, and
  Goodman}]{geiger2024alignments}
Atticus Geiger, Zhengxuan Wu, Christopher Potts, Thomas Icard, and Noah
  Goodman. 2024.
\newblock \href {https://proceedings.mlr.press/v236/geiger24a.html} {Finding
  alignments between interpretable causal variables and distributed neural
  representations}.
\newblock In \emph{Proceedings of the Third Conference on Causal Learning and
  Reasoning}, volume 236 of \emph{Proceedings of Machine Learning Research},
  pages 160--187. PMLR.

\bibitem[{{Gemma Team}(2024)}]{gemmateam2024gemma2}
{Gemma Team}. 2024.
\newblock \href {https://arxiv.org/abs/2408.00118} {{Gemma 2}: Improving open
  language models at a practical size}.
\newblock ArXiv preprint arXiv:2408.00118.

\bibitem[{Grattafiori et~al.(2024)Grattafiori, Dubey, Jauhri, Pandey, Kadian,
  Al-Dahle et~al.}]{grattafiori2024llama3}
Aaron Grattafiori, Abhimanyu Dubey, Abhinav Jauhri, Abhinav Pandey, Abhishek
  Kadian, Ahmad Al-Dahle, et~al. 2024.
\newblock \href {https://arxiv.org/abs/2407.21783v3} {The {Llama 3} herd of
  models}.
\newblock ArXiv preprint arXiv:2407.21783, version 3.

\bibitem[{Gulordava et~al.(2018)Gulordava, Bojanowski, Grave, Linzen, and
  Baroni}]{gulordava-etal-2018-colorless}
Kristina Gulordava, Piotr Bojanowski, {\'E}douard Grave, Tal Linzen, and Marco
  Baroni. 2018.
\newblock \href {https://doi.org/10.18653/v1/N18-1108} {Colorless green
  recurrent networks dream hierarchically}.
\newblock In \emph{Proceedings of the 2018 Conference of the North {A}merican
  Chapter of the Association for Computational Linguistics: Human Language
  Technologies, Volume 1 (Long Papers)}, pages 1195--1205, New Orleans,
  Louisiana. Association for Computational Linguistics.

\bibitem[{He et~al.(2024)He, Shu, Ge, Chen, Wang, Zhou, Liu, Guo, Huang, Wu,
  Jiang, and Qiu}]{he2024llamascope}
Zhengfu He, Wentao Shu, Xuyang Ge, Lingjie Chen, Junxuan Wang, Yunhua Zhou,
  Frances Liu, Qipeng Guo, Xuanjing Huang, Zuxuan Wu, Yu-Gang Jiang, and Xipeng
  Qiu. 2024.
\newblock \href {https://arxiv.org/abs/2410.20526} {{Llama Scope}: Extracting
  millions of features from {Llama-3.1-8B} with sparse autoencoders}.
\newblock ArXiv preprint arXiv:2410.20526.

\bibitem[{Heimersheim and Nanda(2024)}]{heimersheim2024patching}
Stefan Heimersheim and Neel Nanda. 2024.
\newblock \href {https://arxiv.org/abs/2404.15255} {How to use and interpret
  activation patching}.
\newblock ArXiv preprint arXiv:2404.15255.

\bibitem[{Hewitt and Liang(2019)}]{hewitt-liang-2019-designing}
John Hewitt and Percy Liang. 2019.
\newblock \href {https://doi.org/10.18653/v1/D19-1275} {Designing and
  interpreting probes with control tasks}.
\newblock In \emph{Proceedings of the 2019 Conference on Empirical Methods in
  Natural Language Processing and the 9th International Joint Conference on
  Natural Language Processing (EMNLP-IJCNLP)}, pages 2733--2743, Hong Kong,
  China. Association for Computational Linguistics.

\bibitem[{Huang et~al.(2024)Huang, Wu, Potts, Geva, and
  Geiger}]{huang-etal-2024-ravel}
Jing Huang, Zhengxuan Wu, Christopher Potts, Mor Geva, and Atticus Geiger.
  2024.
\newblock \href {https://doi.org/10.18653/v1/2024.acl-long.470} {{RAVEL}:
  Evaluating interpretability methods on disentangling language model
  representations}.
\newblock In \emph{Proceedings of the 62nd Annual Meeting of the Association
  for Computational Linguistics (Volume 1: Long Papers)}, pages 8669--8687,
  Bangkok, Thailand. Association for Computational Linguistics.

\bibitem[{Huben et~al.(2024)Huben, Cunningham, Smith, Ewart, and
  Sharkey}]{huben2024sparse}
Robert Huben, Hoagy Cunningham, Logan Smith, Aidan Ewart, and Lee Sharkey.
  2024.
\newblock \href
  {https://proceedings.iclr.cc/paper_files/paper/2024/hash/1fa1ab11f4bd5f94b2ec20e794dbfa3b-Abstract-Conference.html}
  {Sparse autoencoders find highly interpretable features in language models}.
\newblock In \emph{The Twelfth International Conference on Learning
  Representations}.

\bibitem[{Jacovi and Goldberg(2020)}]{jacovi-goldberg-2020-towards}
Alon Jacovi and Yoav Goldberg. 2020.
\newblock \href {https://doi.org/10.18653/v1/2020.acl-main.386} {Towards
  faithfully interpretable {NLP} systems: How should we define and evaluate
  faithfulness?}
\newblock In \emph{Proceedings of the 58th Annual Meeting of the Association
  for Computational Linguistics}, pages 4198--4205. Association for
  Computational Linguistics.

\bibitem[{Karvonen et~al.(2025)Karvonen, Rager, Lin, Tigges, Bloom, Chanin,
  Lau, Farrell, Mcdougall, Ayonrinde, Till, Wearden, Conmy, Marks, and
  Nanda}]{karvonen2025saebench}
Adam Karvonen, Can Rager, Johnny Lin, Curt Tigges, Joseph~Isaac Bloom, David
  Chanin, Yeu-Tong Lau, Eoin Farrell, Callum~Stuart Mcdougall, Kola Ayonrinde,
  Demian Till, Matthew Wearden, Arthur Conmy, Samuel Marks, and Neel Nanda.
  2025.
\newblock \href {https://proceedings.mlr.press/v267/karvonen25a.html}
  {{SAEB}ench: A comprehensive benchmark for sparse autoencoders in language
  model interpretability}.
\newblock In \emph{Proceedings of the 42nd International Conference on Machine
  Learning}, volume 267 of \emph{Proceedings of Machine Learning Research},
  pages 29223--29264. PMLR.

\bibitem[{Kram{\'a}r et~al.(2024)Kram{\'a}r, Lieberum, Shah, and
  Nanda}]{kramar2024atp}
J{\'a}nos Kram{\'a}r, Tom Lieberum, Rohin Shah, and Neel Nanda. 2024.
\newblock \href {https://arxiv.org/abs/2403.00745} {{AtP*}: An efficient and
  scalable method for localizing {LLM} behaviour to components}.
\newblock ArXiv preprint arXiv:2403.00745.

\bibitem[{Lakretz et~al.(2019)Lakretz, Kruszewski, Desbordes, Hupkes, Dehaene,
  and Baroni}]{lakretz-etal-2019-emergence}
Yair Lakretz, Germ{\'a}n Kruszewski, Th{\'e}o Desbordes, Dieuwke Hupkes,
  Stanislas Dehaene, and Marco Baroni. 2019.
\newblock \href {https://doi.org/10.18653/v1/N19-1002} {The emergence of number
  and syntax units in {LSTM} language models}.
\newblock In \emph{Proceedings of the 2019 Conference of the North {A}merican
  Chapter of the Association for Computational Linguistics: Human Language
  Technologies, Volume 1 (Long and Short Papers)}, pages 11--20, Minneapolis,
  Minnesota. Association for Computational Linguistics.

\bibitem[{Lieberum et~al.(2024)Lieberum, Rajamanoharan, Conmy, Smith, Sonnerat,
  Varma, Kram{\'a}r, Dragan, Shah, and Nanda}]{lieberum2024gemmascope}
Tom Lieberum, Senthooran Rajamanoharan, Arthur Conmy, Lewis Smith, Nicolas
  Sonnerat, Vikrant Varma, J{\'a}nos Kram{\'a}r, Anca Dragan, Rohin Shah, and
  Neel Nanda. 2024.
\newblock \href {https://arxiv.org/abs/2408.05147} {{Gemma Scope}: Open sparse
  autoencoders everywhere all at once on {Gemma 2}}.
\newblock ArXiv preprint arXiv:2408.05147.

\bibitem[{Linzen et~al.(2016)Linzen, Dupoux, and
  Goldberg}]{linzen-etal-2016-assessing}
Tal Linzen, Emmanuel Dupoux, and Yoav Goldberg. 2016.
\newblock \href {https://doi.org/10.1162/tacl_a_00115} {Assessing the ability
  of {LSTM}s to learn syntax-sensitive dependencies}.
\newblock \emph{Transactions of the Association for Computational Linguistics},
  4:521--535.

\bibitem[{Luo et~al.(2026)Luo, Liang, and Xuan}]{luo2026steercheck}
Daming Luo, Christy Liang, and Junyu Xuan. 2026.
\newblock \href {https://arxiv.org/abs/2608.24335v1} {{SteerCheck}: Attribution
  specificity and alignment leakage in activation-steering audits}.
\newblock ArXiv preprint arXiv:2608.24335, version 1.

\bibitem[{Makelov et~al.(2024)Makelov, Lange, Geiger, and
  Nanda}]{makelov2024subspace}
Aleksandar Makelov, Georg Lange, Atticus Geiger, and Neel Nanda. 2024.
\newblock \href
  {https://proceedings.iclr.cc/paper_files/paper/2024/hash/70b8505ac79e3e131756f793cd80eb8d-Abstract-Conference.html}
  {Is this the subspace you are looking for? an interpretability illusion for
  subspace activation patching}.
\newblock In \emph{The Twelfth International Conference on Learning
  Representations}.

\bibitem[{Marks et~al.(2025)Marks, Rager, Michaud, Belinkov, Bau, and
  Mueller}]{marks2025sparse}
Samuel Marks, Can Rager, Eric~J. Michaud, Yonatan Belinkov, David Bau, and
  Aaron Mueller. 2025.
\newblock \href {https://features.baulab.info/} {Sparse feature circuits:
  Discovering and editing interpretable causal graphs in language models}.
\newblock In \emph{International Conference on Learning Representations}.

\bibitem[{Marvin and Linzen(2018)}]{marvin-linzen-2018-targeted}
Rebecca Marvin and Tal Linzen. 2018.
\newblock \href {https://doi.org/10.18653/v1/D18-1151} {Targeted syntactic
  evaluation of language models}.
\newblock In \emph{Proceedings of the 2018 Conference on Empirical Methods in
  Natural Language Processing}, pages 1192--1202, Brussels, Belgium.
  Association for Computational Linguistics.

\bibitem[{O'Neill et~al.(2025)O'Neill, Gumran, and Klindt}]{oneill2025compute}
Charles O'Neill, Alim Gumran, and David Klindt. 2025.
\newblock \href {https://proceedings.mlr.press/v267/o-neill25a.html} {Compute
  optimal inference and provable amortisation gap in sparse autoencoders}.
\newblock In \emph{Proceedings of the 42nd International Conference on Machine
  Learning}, volume 267 of \emph{Proceedings of Machine Learning Research},
  pages 46877--46896. PMLR.

\bibitem[{Rajamanoharan et~al.(2024{\natexlab{a}})Rajamanoharan, Conmy, Smith,
  Lieberum, Varma, Kram{\'a}r, Shah, and Nanda}]{rajamanoharan2024gated}
Senthooran Rajamanoharan, Arthur Conmy, Lewis Smith, Tom Lieberum, Vikrant
  Varma, J{\'a}nos Kram{\'a}r, Rohin Shah, and Neel Nanda. 2024{\natexlab{a}}.
\newblock \href {https://doi.org/10.52202/079017-0024} {Improving sparse
  decomposition of language model activations with gated sparse autoencoders}.
\newblock In \emph{Advances in Neural Information Processing Systems},
  volume~37, pages 775--818. Curran Associates, Inc.

\bibitem[{Rajamanoharan et~al.(2024{\natexlab{b}})Rajamanoharan, Lieberum,
  Sonnerat, Conmy, Varma, Kram{\'a}r, and Nanda}]{rajamanoharan2024jumping}
Senthooran Rajamanoharan, Tom Lieberum, Nicolas Sonnerat, Arthur Conmy, Vikrant
  Varma, J{\'a}nos Kram{\'a}r, and Neel Nanda. 2024{\natexlab{b}}.
\newblock \href {https://arxiv.org/abs/2407.14435} {Jumping ahead: Improving
  reconstruction fidelity with {JumpReLU} sparse autoencoders}.
\newblock ArXiv preprint arXiv:2407.14435.

\bibitem[{Rimsky et~al.(2024)Rimsky, Gabrieli, Schulz, Tong, Hubinger, and
  Turner}]{rimsky2024steering}
Nina Rimsky, Nick Gabrieli, Julian Schulz, Meg Tong, Evan Hubinger, and
  Alexander Turner. 2024.
\newblock \href {https://doi.org/10.18653/v1/2024.acl-long.828} {Steering
  {Llama 2} via contrastive activation addition}.
\newblock In \emph{Proceedings of the 62nd Annual Meeting of the Association
  for Computational Linguistics (Volume 1: Long Papers)}, pages 15504--15522.

\bibitem[{Sun et~al.(2026)Sun, Wang, and Hu}]{sun2026price}
Wenjie Sun, Di~Wang, and Lijie Hu. 2026.
\newblock \href
  {https://proceedings.iclr.cc/paper_files/paper/2026/hash/76358e88c3ebd38ba58af9d0d84b71e4-Abstract-Conference.html}
  {The price of amortized inference in sparse autoencoders}.
\newblock In \emph{International Conference on Learning Representations}.

\bibitem[{Turner et~al.(2024)Turner, Thiergart, Leech, Udell, Vazquez, Mini,
  and MacDiarmid}]{turner2024actadd}
Alexander~Matt Turner, Lisa Thiergart, Gavin Leech, David Udell, Juan~J.
  Vazquez, Ulisse Mini, and Monte MacDiarmid. 2024.
\newblock \href {https://arxiv.org/abs/2308.10248v4} {Activation addition:
  Steering language models without optimization}.
\newblock ArXiv preprint arXiv:2308.10248, version 4.

\bibitem[{Vig et~al.(2020)Vig, Gehrmann, Belinkov, Qian, Nevo, Singer, and
  Shieber}]{vig2020investigating}
Jesse Vig, Sebastian Gehrmann, Yonatan Belinkov, Sharon Qian, Daniel Nevo,
  Yaron Singer, and Stuart Shieber. 2020.
\newblock \href
  {https://proceedings.neurips.cc/paper/2020/hash/92650b2e92217715fe312e6fa7b90d82-Abstract.html}
  {Investigating gender bias in language models using causal mediation
  analysis}.
\newblock In \emph{Advances in Neural Information Processing Systems},
  volume~33.

\bibitem[{Wang et~al.(2023)Wang, Variengien, Conmy, Shlegeris, and
  Steinhardt}]{wang2023ioi}
Kevin Wang, Alexandre Variengien, Arthur Conmy, Buck Shlegeris, and Jacob
  Steinhardt. 2023.
\newblock \href {https://arxiv.org/abs/2211.00593} {Interpretability in the
  wild: A circuit for indirect object identification in {GPT-2} small}.
\newblock In \emph{The Eleventh International Conference on Learning
  Representations}.

\bibitem[{Warstadt et~al.(2020)Warstadt, Parrish, Liu, Mohananey, Peng, Wang,
  and Bowman}]{warstadt-etal-2020-blimp-benchmark}
Alex Warstadt, Alicia Parrish, Haokun Liu, Anhad Mohananey, Wei Peng, Sheng-Fu
  Wang, and Samuel~R. Bowman. 2020.
\newblock \href {https://doi.org/10.1162/tacl_a_00321} {{BL}i{MP}: The
  benchmark of linguistic minimal pairs for {E}nglish}.
\newblock \emph{Transactions of the Association for Computational Linguistics},
  8:377--392.

\bibitem[{Zhang and Nanda(2024)}]{zhang2024bestpractices}
Fred Zhang and Neel Nanda. 2024.
\newblock \href
  {https://proceedings.iclr.cc/paper_files/paper/2024/hash/06a52a54c8ee03cd86771136bc91eb1f-Abstract-Conference.html}
  {Towards best practices of activation patching in language models: Metrics
  and methods}.
\newblock In \emph{The Twelfth International Conference on Learning
  Representations}.

\end{thebibliography}
